\documentclass[10pt]{article}
\usepackage[preprint]{tmlr}

\usepackage[T1]{fontenc}
\usepackage[utf8]{inputenc}
\usepackage{lmodern}
\usepackage{textcomp}
\usepackage{amsmath,amssymb,bm}
\usepackage{algorithm}
\usepackage{algpseudocode}
\usepackage{booktabs}
\usepackage{multirow}
\usepackage{makecell}
\usepackage{xcolor}
\usepackage{graphicx}
\usepackage{tikz}
\usetikzlibrary{fit,positioning}
\graphicspath{{figures/}}
\usepackage{silence}
\usepackage{microtype}
\usepackage{enumitem}
\usepackage{url}
\usepackage{hyperref}
\hypersetup{hidelinks}
\usepackage{array}
\usepackage{tabularx}  %
\usepackage{caption}
\usepackage{placeins}
\usepackage{adjustbox}
\usepackage{rotating}
\usepackage{titletoc}

\makeatletter
\g@addto@macro\appendix{%
  \@addtoreset{table}{section}\@addtoreset{figure}{section}\@addtoreset{equation}{section}%
  \setcounter{table}{0}\setcounter{figure}{0}\setcounter{equation}{0}%
  \gdef\theHtable{\@Alph\c@section.\arabic{table}}%
  \gdef\theHfigure{\@Alph\c@section.\arabic{figure}}%
  \gdef\theHequation{\@Alph\c@section.\arabic{equation}}}
\makeatother

\newcommand{\appref}[1]{Appendix~\ref{#1}}

\newcommand{\ms}[2]{#1{\scriptsize$\pm$#2}}
\newcommand{\msp}[3]{#2{\scriptsize#1$\pm$#3}}
\newcommand{\sr}[2]{#1~{\scriptsize(#2)}}
\newcommand{\colcite}[2][\footnotesize]{{#1 (#2)}}
\newlength{\colcellwidth}
\newlength{\colcelltmp}
\newcommand{\colcellreset}{\setlength{\colcellwidth}{0pt}}
\newcommand{\colcellfit}[1]{%
  \settowidth{\colcelltmp}{#1}%
  \ifdim\colcelltmp>\colcellwidth\setlength{\colcellwidth}{\colcelltmp}\fi}

\newcommand{\papertitle}{vMF Sentence LDA: A Spherical Topic Model over Sentence Embeddings}

\newcommand{\paperabstract}{%
Latent Dirichlet Allocation (LDA) and models derived from it remain widely used topic models. LDA observes each document as a bag-of-words and models each topic by a categorical distribution over the vocabulary, so that it uses neither the internal structure of the document nor the similarity in meaning between words.
Earlier work has responded to this limitation in two ways: many models have introduced embeddings, and some have assigned topics to sentences rather than to words. Their combination, a topic model that observes sentence embeddings, remains little explored.
We propose \textit{vMF Sentence LDA} (vSLDA), which keeps the admixture structure of LDA, observes each sentence as its L2-normalized embedding and models each topic by a von Mises--Fisher (vMF) distribution, which matches the cosine geometry of sentence embeddings.
Its per-topic parameter count and per-iteration cost are linear in the embedding dimension, versus quadratic for the full-covariance Gaussian distribution in the existing model over sentence embeddings.
We evaluate vSLDA where the limitation is expected to matter most, among topics that share much of their vocabulary: the topics that subdivide the one subject of a collection, and the narrow topics that result when a corpus is divided into a large number of topics.
On two corpora, vSLDA attains the best mean rank against eight baselines when the fine categories within each coarse category are classified from the document-topic distributions. On the whole corpus, its advantage appears or widens as the number of topics grows.
Weighting the word frequencies of each sentence by its topic posterior yields expected topic-word counts of the same form as those of LDA, so that the standard topic coherence and diversity measures apply to models that assign topics to sentences. On their product, topic quality, vSLDA leads in most within-category conditions of both corpora.
}

\begin{document}

\title{\papertitle\thanks{Code is available at \url{https://github.com/izumi-lab/vSLDA}.}}
\author{\name Ryotaro Kobayashi \email kobayashi-ryotaro0212@g.ecc.u-tokyo.ac.jp \\
      \name Yuri Murayama \email murayama.yuri@sys.t.u-tokyo.ac.jp \\
      \name Kiyoshi Izumi \email izumi@sys.t.u-tokyo.ac.jp \\
      \addr Graduate School of Engineering, The University of Tokyo, Tokyo, Japan}
\maketitle
\begin{abstract}
\paperabstract
\end{abstract}

\section{Introduction}
\label{sec:intro}
Latent topic models represent each document as a mixture of latent topics, and their per-document topic distributions are used across a wide range of fields as interpretable low-dimensional document representations~\citep{griffiths-2004-finding,roberts-2014-stm,tirunillai-2014-chatter,boyd-graber-2017-applications,hansen-2018-fomc,wu-2024-survey}. Latent Dirichlet Allocation (LDA)~\citep{blei-2003-lda} and the models derived from it remain among the most widely used of these models~\citep{jelodar-2019-lda,chauhan-2022-lda,hoyle-2022-neural}. LDA observes each document as a bag-of-words (BoW) and models each topic by a categorical distribution over the vocabulary.
Because a BoW-based model relies only on the co-occurrence of words within documents, however, it uses neither the similarity in meaning between different words~\citep{das-2015-gaussian,dieng-2020-topic} nor the context that a sentence gives to its words~\citep{gruber-2007-htmm,cha-2024-sentencelda}. This paper concerns the condition under which this limitation is expected to matter most: the topics to be distinguished share much of their vocabulary. Such topics differ little in the word frequencies on which a BoW-based model relies, so that separating them is expected to depend on the meaning and the context that such a model does not use.

The condition arises in two situations. In the first, the documents of a collection address one subject and the topics are its subdivisions. Applications of topic models often have this form: the collection is the reviews of one kind of product~\citep{titov-2008-mglda} or the risk disclosures of firms~\citep{bao-2014-risk}, and the topics of interest are not global properties of the subject, such as the brand of a product, but its aspects, such as the features of a product that reviewers rate or the types of risk that firms disclose. Earlier studies report that word-level topic models have difficulty in this situation~\citep{titov-2008-mglda,cha-2024-sentencelda}. In the second, a corpus is divided into a large number of topics. Topic models with large numbers of topics have been trained~\citep{yuan-2015-lightlda} and reported to improve applications as the number grows~\citep{wang-2015-peacock}. A larger number gives topics that address more specific areas~\citep{griffiths-2004-finding}, so that each topic covers a narrower range of content and is set against topics that use largely the same words. The experiments of this paper indicate this difficulty of word-level models in both situations.

One response to this limitation changes the observation unit, the piece of text that receives one topic assignment. In topic models, topics are typically assigned to words, but some models over word frequencies have assigned them to sentences, a unit intermediate between the word and the document~\citep{bao-2014-risk,balikas-2016-slda}. By giving all words of a sentence the same topic, these models reflect the division of a document into sentences, a structure within the document that a BoW representation discards: a topic is assigned to the combination of words in the sentence rather than to each word separately. The underlying assumption is that a sentence mostly addresses a single subject and is therefore a topically more homogeneous unit than the word~\citep{balikas-2016-slda}. A document is then a set of sentences treated as exchangeable. Throughout the paper, a topic model is called word-level, sentence-level or document-level according to its observation unit. The other response changes the representation of the observation: embeddings have been introduced into topic models in several ways~\citep{nguyen-2015-lftm,dieng-2020-topic,bianchi-2021-ctm,cha-2024-sentencelda}. Word embeddings learned from large external corpora place words with similar meanings close to each other, and so bring semantic knowledge from outside the corpus into the topic model~\citep{das-2015-gaussian,dieng-2020-topic}. Gaussian LDA~\citep{das-2015-gaussian} is an early probabilistic topic model that takes embeddings from a pretrained model as its observations, in its case word embeddings. Sentence encoders map a sentence to a single vector such that semantically similar sentences are close to each other, and their embeddings are used for semantic similarity comparison, clustering and retrieval~\citep{reimers-2019-sbert,gao-2021-simcse,muennighoff-2023-mteb}. Observing sentence embeddings therefore combines the two responses. SentenceLDA~\citep{cha-2024-sentencelda} does so by extending Gaussian LDA directly to the sentence unit, and we call it Gaussian Sentence LDA (GSLDA) to make its observation distribution explicit. Gaussian LDA and GSLDA both model each topic by a multivariate Gaussian distribution in Euclidean space.

Semantic similarity between words or sentences, however, is commonly measured by the cosine similarity of their embeddings~\citep{meng-2019-spherical}, and sentence encoders are typically trained with objectives defined on cosine similarity and evaluated by it~\citep{cer-2017-sts,reimers-2019-sbert,gao-2021-simcse}. For an encoder trained in this way, the direction of an embedding is the component that training constrains to reflect meaning, whereas its norm is not.
GSLDA already observes sentence embeddings, but its likelihood is not matched to its observations. In addition, the embeddings of widely used sentence encoders have several hundred dimensions or more ($M=384$ and $768$ for the encoders used in this paper; Section~\ref{subsec:implementation}), and at this dimensionality a full covariance matrix per topic is demanding both in computing time and in the number of observations that its estimation requires. Computing the likelihood of one sentence under one topic takes $O(M^2)$ operations, where $M$ is the embedding dimension, and the $M(M+1)/2$ covariance parameters of each topic must be estimated from the sentences assigned to that topic, far fewer than the word tokens that Gaussian LDA pools, so the estimate is poorly determined (Section~\ref{subsec:why_gslda}). Representations compared by direction are naturally modeled by probability distributions on the sphere, and spherical topic models have applied this idea to normalized word embeddings through von Mises--Fisher (vMF) distributions~\citep{batmanghelich-2016-nonparametric,li-2016-integrating}. The vMF distribution has only a mean direction and a concentration, $M$ parameters per topic, and its density depends on an embedding only through its cosine similarity to the topic direction. Yet, to our knowledge, no spherical topic model has taken the sentence as its observation unit, and this combination, although promising for making topic models over sentence embeddings both accurate and practical to train, has not been examined.

We propose \textit{vMF Sentence LDA} (vSLDA), a latent topic model that observes each sentence as its L2-normalized embedding, a point on the unit hypersphere, and models each topic by a vMF distribution on that hypersphere. It keeps the admixture structure of LDA (per-document topic proportions and a latent topic assigned to each observation) and differs from LDA in two respects: the observation unit is the sentence rather than the word, and the observation distribution is a vMF distribution over embeddings rather than a categorical distribution over the vocabulary. With respect to GSLDA, vSLDA observes the same sentence embeddings and replaces the full-covariance Gaussian distribution of each topic with a vMF distribution. The baselines include models that differ from vSLDA in only one of the two respects, so that the contribution of each can be examined separately (Section~\ref{subsec:baselines}).
We examine this combination in both of the situations above. For the first, each model is fitted only to the documents of one coarse category of a labeled corpus, and its topics are evaluated against the fine categories within that coarse category. For the second, each model is fitted to the whole corpus at an increasing number of topics. All models are evaluated under a shared protocol by downstream classification accuracy, topic coherence and topic diversity. Sentence-level topic models lack an established word-based evaluation protocol, so the topic-word counts that coherence and diversity require are obtained from the topic posteriors of the sentences (Section~\ref{subsec:repwords}). The experiments compare vSLDA with word-level, sentence-level, document-level and neural topic models, and examine how the differences depend on document length, the number of topics and the amount of labeled data available to the downstream classifier.

The contributions of this work are the following.
\begin{itemize}[leftmargin=*]
\item We propose vSLDA, a spherical topic model that keeps the admixture structure of LDA, assigns topics to sentences rather than words and models each topic by a vMF distribution over L2-normalized sentence embeddings. Its parameters are estimated by a stochastic approximation EM (SAEM) with a Markov chain E-step, whose Monte Carlo sufficient statistics are averaged across iterations. Its per-iteration cost is linear in the embedding dimension, and in a single-threaded comparison of our implementations, training vSLDA was about two orders of magnitude faster than training the GSLDA of \citet{cha-2024-sentencelda} (Section~\ref{sec:method}).
\item On two corpora with coarse and fine category labels, we compare vSLDA with baselines that span word-level, sentence-level, document-level and neural topic models, in the two situations in which topics share much of their vocabulary. In classification within coarse categories, vSLDA attains the best mean rank against eight baselines. The advantage largely persists under limited labels and, on the whole corpus, appears or widens as the number of topics grows. On coherence, vSLDA leads on the corpus with relatively longer documents but can fall behind a neural baseline on the corpus with short documents. Its topic diversity stays near its maximum, however, so on topic quality, the product of the two, vSLDA is close to the neural baseline on the corpus with short documents and ahead on the other (Section~\ref{sec:results}). Because the baselines vary the observation unit and the likelihood separately (Table~\ref{tab:model_map}), the comparison also indicates that neither choice alone reaches the result of vSLDA. A further analysis attributes the deficit of GSLDA to the covariance structure of its topics (Section~\ref{sec:discussion}). The comparison is complemented by robustness experiments across sentence encoders, downstream classifiers, coherence measures, a wide range of topic counts and the hyperparameters of vSLDA.
\item We show that weighting within-sentence word frequencies by sentence-topic posterior probabilities yields expected topic-word counts of the same form as the topic-word counts of word-level LDA, so that the topics estimated by sentence-level and word-level models can be evaluated consistently by standard word-based coherence and diversity without modifying the metrics, which the previous study left open~\citep{cha-2024-sentencelda} (Section~\ref{subsec:repwords}). The implementation and the evaluation workflow are made available.
\end{itemize}
\section{Related Work}
\label{sec:related}

Section~\ref{sec:intro} described two responses to the BoW limitation and the view of embeddings as directions. This section reviews the corresponding three lines of prior work: topic models in embedding spaces, spherical topic models and sentence-level topic models, the last of which contains the closest prior work. The baselines among the models discussed here are collected in Table~\ref{tab:model_map} (Section~\ref{subsec:baselines}).

\subsection{Topic Models in Embedding Spaces}
Gaussian LDA (GLDA) replaces the categorical topic distributions of LDA by multivariate Gaussian distributions over word embeddings while retaining the document-level mixture~\citep{das-2015-gaussian}. Word embeddings have also been used to inform the word generation process in other ways~\citep{nguyen-2015-lftm,xun-2017-correlated,zhao-2018-wedtm}.
Neural topic models integrate embeddings without observing them: the Embedded Topic Model (ETM) parameterizes the topic-word distribution by inner products between topic and word embeddings~\citep{dieng-2020-topic}, and the Contextualized Topic Model (ConTM here, to avoid confusion with the Correlated Topic Model) feeds a pretrained document embedding to its inference network alongside the BoW input~\citep{bianchi-2021-ctm}.
A separate line replaces the generative model by clustering of word, contextual token, sentence or document embeddings~\citep{sia-2020-tired,fujita-2027-topiclear}, as in Top2Vec and BERTopic~\citep{angelov-2020-top2vec,grootendorst-2022-bertopic,angelov-2024-tokens}, or elicits topics directly from large language models~\citep{pham-2024-topicgpt}.
These pipelines do not infer per-document topic proportions as latent variables under a likelihood.

\subsection{Spherical Topic Models}
\label{subsec:spherical}
Treating text representations as directions on the unit sphere is a view shared beyond topic modeling, by spherical text embeddings~\citep{meng-2019-spherical} and vMF output layers~\citep{kumar-2019-vmf}. The spherical topic models reviewed below build on the same view. Mixtures of vMF distributions were applied to the normalized term-frequency vectors of documents~\citep{banerjee-2005-clustering,zhong-2005-generative,gopal-2014-clustering}, where each document belongs to a single component and no per-document topic proportions are inferred. The Spherical Admixture Model (SAM) models normalized document vectors with vMF likelihoods while retaining an admixture structure similar to LDA~\citep{reisinger-2010-spherical}. Because each document is a single observation, the admixture is formed over the topic directions rather than over per-unit assignments, so SAM has no sentence- or word-level topic assignments. We include SAM as the document-level vMF baseline, with normalized term-frequency vectors as observations.
Spherical topic models over word embeddings place vMF distributions on normalized word embeddings, within a hierarchical Dirichlet process~\citep{batmanghelich-2016-nonparametric}, with several vMF components per topic~\citep{li-2016-integrating} or alongside knowledge graph embeddings~\citep{ennajari-2023-spherical,ennajari-2025-correlated}. They are the most direct precedents for vSLDA, which differs from them in observing sentence rather than word embeddings and, from \citet{li-2016-integrating}, further in estimating its parameters by a stochastic approximation EM rather than by their Monte Carlo EM (Section~\ref{subsec:inference}), and we include their model, run with a single component per topic, as the word-level vMF baseline (vLDA).
Hyperspherical geometry has also entered neural topic models in two ways. One line, building on the hyperspherical variational auto-encoder of \citet{davidson-2018-hyperspherical}, places the vMF distribution on the latent document representation~\citep{xu-2018-spherical,xu-2023-vontss,adhya-2025-s2wtm}. The other parameterizes the topic-word distribution of the decoder by vMF distributions over pretrained word embeddings~\citep{wang-2024-set,guo-2026-vmnvtm}. In both, the observation remains the BoW, so a concentration parameter, where present, acts as a temperature of the topic-word distribution over fixed embeddings rather than measuring how tightly the observed units cluster around a topic direction, as the concentration parameter of vSLDA does (Section~\ref{subsec:relation}), and neither line assigns topics to sentences. The neural line is therefore represented in this paper by ETM and ConTM, two widely used neural topic models that integrate embeddings through the decoder and the inference network, respectively.

\subsection{Sentence-Level Topic Models}
Sentence-level topic models change the unit of topic assignment from the word to the sentence~\citep{gruber-2007-htmm,jo-2011-asum,bao-2014-risk,balikas-2016-slda}. The sentence-level LDA of \citet{balikas-2016-slda}, which we denote SentLDA to avoid confusion with supervised LDA, is the direct form of this change, and the observations of all these models remain word frequencies.
GSLDA~\citep{cha-2024-sentencelda} combines Sentence-BERT with Gaussian LDA by treating each sentence embedding as an observation from a Gaussian topic distribution. SenClu~\citep{schneider-2024-senclu} also takes sentence embeddings, of groups of a few consecutive sentences (three by default), but, as its author presents it, performs similarity-based clustering: within an expectation--maximization-style loop, each group is hard-assigned to a single topic by the cosine similarity to the topic centroids weighted by the document's topic proportions, and those proportions are in turn the smoothed fractions of groups so assigned, and we leave it outside Table~\ref{tab:model_map}.
GSLDA is the closest prior work: it keeps the admixture structure of LDA and makes the sentence embedding its observation. It inherits the full-covariance likelihood of Gaussian LDA, however, and \citet{cha-2024-sentencelda} report that the default prior of the reference implementation fails on their data. vSLDA keeps the observation and the generative process of GSLDA and replaces this likelihood by a vMF distribution (Section~\ref{sec:method}).
\section{vMF Sentence LDA}
\label{sec:method}

\subsection{Problem Setting}
\label{subsec:setup}
Let $\mathcal{D} = \{D_d\}_{d=1}^D$ be a collection of $D$ documents, where document $D_d$ consists of $N_d$ sentences $\{s_{di}\}_{i=1}^{N_d}$, and let $N_s=\sum_{d=1}^{D}N_d$ be the total number of sentences.
A pretrained sentence encoder $f_\phi$ maps each sentence to $\bm h_{di} = f_\phi(s_{di}) \in \mathbb{R}^M$, and L2 normalization $\bm w_{di} = \bm h_{di}/\|\bm h_{di}\|_2$ places it on the unit hypersphere $\mathbb{S}^{M-1} = \{\bm x \in \mathbb{R}^M : \|\bm x\|_2 = 1\}$.
Let $K$ be the number of topics, $\bm\theta_d$ the topic distribution of document $d$, $z_{di} \in \{1,\dots,K\}$ the topic assignment of sentence $i$ of document $d$, and $\Delta_k=(\bm\mu_k,\kappa_k)$ the parameters of topic $k$, where $\bm\mu_k\in\mathbb{S}^{M-1}$ is its mean direction and $\kappa_k\ge 0$ its concentration parameter. We write $\bm\Delta=\{\Delta_k\}_{k=1}^{K}$, and $\bm\alpha\in\mathbb{R}_{>0}^{K}$ for the Dirichlet parameter.
Given the normalized sentence embeddings $\{\bm w_{di}\}$, the goal is to estimate the document-topic distributions $\bm\theta = \{\bm\theta_d\}_{d=1}^{D}$, the sentence-topic assignments $\bm z = \{z_{di}\}$ and the topic parameters $\bm\Delta$.
The encoder $f_\phi$ is fixed throughout and is not part of the model. Any encoder whose outputs are meant to be compared by cosine similarity can be used after normalization.

\subsection{Generative Process}
\label{subsec:relation}
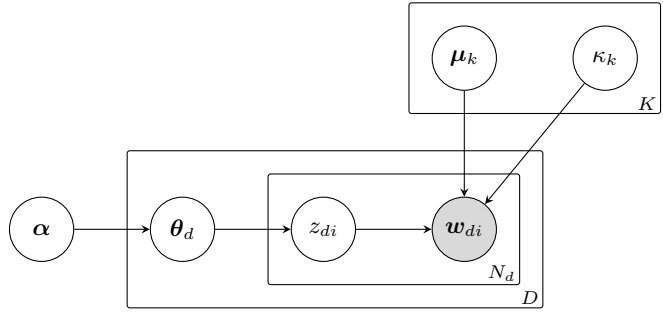
\begin{figure}[htbp]
    \centering
    \begin{tikzpicture}[
      node distance=8mm and 10mm,
      latent/.style={circle,draw,minimum size=8.5mm,inner sep=0pt,font=\small},
      obs/.style={latent,fill=black!15},
      plate/.style={draw,inner sep=3mm,rounded corners=1pt},
      plate label/.style={font=\scriptsize,anchor=south east,inner sep=1.5pt},
      >=stealth]
      \node[latent] (alpha) {$\bm\alpha$};
      \node[latent,right=of alpha] (theta) {$\bm\theta_d$};
      \node[latent,right=of theta] (z) {$z_{di}$};
      \node[obs,right=of z] (w) {$\bm w_{di}$};
      \node[latent,above=14mm of w] (mu) {$\bm\mu_k$};
      \node[latent,right=of mu] (kappa) {$\kappa_k$};
      \draw[->] (alpha)--(theta);
      \draw[->] (theta)--(z);
      \draw[->] (z)--(w);
      \draw[->] (mu)--(w);
      \draw[->] (kappa)--(w);
      \node[plate,fit=(z)(w)] (sent) {};
      \node[plate label] at (sent.south east) {$N_d$};
      \node[plate,fit=(theta)(sent)] (doc) {};
      \node[plate label] at (doc.south east) {$D$};
      \node[plate,fit=(mu)(kappa)] (top) {};
      \node[plate label] at (top.south east) {$K$};
    \end{tikzpicture}
    \caption{\textbf{Graphical model of vMF Sentence LDA (vSLDA).} The plates repeat over the $N_d$ sentences of document $d$, the $D$ documents and the $K$ topics. $\bm\alpha$ is the Dirichlet parameter, $\bm\theta_d$ the document-topic distribution of document $d$, $z_{di}$ the topic assignment of its sentence $i$, $\bm w_{di}$ the normalized sentence embedding of that sentence, and $(\bm\mu_k,\kappa_k)$ the mean direction and concentration parameter of the vMF distribution of topic $k$.}
    \label{fig:gm}
\end{figure}

The von Mises--Fisher (vMF) distribution on $\mathbb{S}^{M-1}$ with mean direction $\bm\mu$ and concentration parameter $\kappa$ has density
\begin{equation}
    f( \bm w \mid \bm\mu, \kappa )
    = C_M(\kappa) \exp( \kappa\,\bm\mu^\top \bm w),
    \qquad
    C_M(\kappa)=\frac{\kappa^{M/2-1}}{(2\pi)^{M/2}\, I_{M/2-1}(\kappa)},
    \label{eq:vmf}
\end{equation}
where $I_\nu$ denotes the modified Bessel function of the first kind of order $\nu$.
The generative process of vSLDA is as follows:
\begin{enumerate}
  \item For each document $d$ in the corpus $\mathcal{D}$:
  \begin{enumerate}
    \item Draw topic proportions $\bm\theta_d \sim \mathrm{Dir}(\bm\alpha)$.
    \item For each sentence index $i=1,\dots,N_d$:
    \begin{enumerate}
      \item Draw a topic assignment $z_{di} \sim \mathrm{Cat}(\bm\theta_d)$.
      \item Draw a sentence embedding $\bm w_{di} \sim f(\bm w \mid \Delta_{z_{di}})$.
    \end{enumerate}
  \end{enumerate}
\end{enumerate}
A document is thus a set of directions on the unit hypersphere and a topic a concentration of directions around $\bm\mu_k$ that tightens as $\kappa_k$ grows. Figure~\ref{fig:gm} shows the graphical model, which retains the document-level mixture structure of LDA~\citep{blei-2003-lda} and changes the observation in both unit and representation. LDA observes a word, a discrete index into a fixed vocabulary drawn from a categorical topic-word distribution, whereas vSLDA observes a sentence as its normalized embedding, a continuous point on $\mathbb{S}^{M-1}$ drawn from a vMF distribution.

The vMF likelihood depends on a sentence embedding only through its cosine similarity to the mean direction $\bm\mu_k$, which agrees with the common practice of measuring semantic similarity by the cosine similarity between embeddings. For unit vectors, $\bm\mu_k^\top \bm w = \cos\angle(\bm\mu_k,\bm w)$, so for a fixed topic $k$ the log-likelihood of sentence $\bm w_{di}$, $\ell_{dik} = \log f(\bm w_{di}\mid \Delta_k) = \log C_M(\kappa_k) + \kappa_k \cos\angle(\bm\mu_k,\bm w_{di})$, is affine and increasing in that similarity. The assignment of a sentence therefore compares topics through their cosine similarities to it, weighted by their concentrations, and the M-step of Section~\ref{subsec:inference} sets each mean direction to the normalized resultant of the assigned sentences (Eq.~\eqref{eq:mstep}), the direction that maximizes their summed cosine similarity~\citep{banerjee-2005-clustering}.
Across topics, the normalizing constant acts as a per-topic bias. With $A_M(\kappa)=I_{M/2}(\kappa)/I_{M/2-1}(\kappa)$, the mean resultant length of the vMF distribution, $\partial_\kappa \log C_M(\kappa) = -A_M(\kappa) < 0$, so a topic with a large concentration attains a high likelihood only for sentences close to its mean direction.

\subsection{Posterior Inference}
\label{subsec:inference}
We estimate the Dirichlet parameter $\bm\alpha$ and the topic parameters $\bm\Delta=\{(\bm\mu_k,\kappa_k)\}_{k=1}^{K}$ by maximizing the marginal likelihood in which the document-topic distributions $\bm\theta$ and the sentence-topic assignments $\bm z$ are marginalized out.
Integrating out $\bm\theta_d$ by Dirichlet--multinomial conjugacy gives the collapsed joint distribution of the data and the sentence-topic assignments
\begin{equation}
  p(\mathcal{D},\bm z\mid \bm\alpha,\bm\Delta)
  =
  \prod_{d=1}^{D}
  \frac{\Gamma(\sum_{k}\alpha_k)}{\Gamma(N_d+\sum_{k}\alpha_k)}
  \prod_{k=1}^{K}\frac{\Gamma(n_{dk}+\alpha_k)}{\Gamma(\alpha_k)}
  \prod_{i=1}^{N_d}
  f(\bm w_{di}\mid \Delta_{z_{di}}),
  \label{eq:collapsed}
\end{equation}
where $n_{dk}$ is the number of sentences of document $d$ assigned to topic $k$ (so that $\sum_k n_{dk}=N_d$) and $\Gamma(\cdot)$ is the gamma function.
The marginal likelihood $p(\mathcal{D}\mid\bm\alpha,\bm\Delta)$ is the sum of Eq.~\eqref{eq:collapsed} over all $K^{N_s}$ assignments $\bm z$, a sum that is intractable, so we resort to a stochastic approximation EM (SAEM)~\citep{delyon-1999-saem,kuhn-2004-saem} with a Markov chain E-step. The procedure differs from the Monte Carlo EM that \citet{li-2016-integrating} use for their vMF topic model in the two respects on which the convergence statement below rests: the Monte Carlo estimates of the sufficient statistics are averaged across outer iterations with decreasing step sizes, and the concentration equation of the M-step is solved exactly rather than by a closed-form approximation (Algorithm~\ref{alg:mcem}, \appref{app:estimation}). Each outer iteration $t=1,\dots,T$ consists of three steps: the E-step applies, to the assignments carried over from the previous iteration, a Gibbs kernel that leaves $p(\bm z\mid\mathcal{D},\bm\alpha^{(t-1)},\bm\Delta^{(t-1)})$ invariant, a stochastic approximation step averages the resulting sufficient statistics with those of the previous iterations, and the M-step maximizes the resulting estimate of the expected complete-data log-likelihood with respect to $\bm\Delta$, giving $\bm\Delta^{(t)}$, and, during a burn-in period, updates $\bm\alpha^{(t)}$.
\paragraph{E-step}
With $\bm\alpha^{(t-1)}$ and $\bm\Delta^{(t-1)}$ held fixed, the assignments are updated by collapsed Gibbs sampling on Eq.~\eqref{eq:collapsed}~\citep{griffiths-2004-finding}, with the conditional
\begin{equation}
  p(z_{di}=k\mid \bm z_{d,-i},\mathcal{D},\bm\alpha^{(t-1)},\bm\Delta^{(t-1)})
  \propto
  \left(n_{dk}^{-i}+\alpha^{(t-1)}_{k}\right)
  \exp(\ell_{dik}),
  \label{eq:gibbs}
\end{equation}
where $n_{dk}^{-i}$ is the number of sentences of document $d$ other than $i$ that are assigned to topic $k$, and $\ell_{dik}$ is evaluated at $\bm\Delta^{(t-1)}$. This is the conditional of LDA with its posterior predictive topic-word term, which depends on the other assignments, replaced by the vMF density at the current point estimate, which depends on none. Each E-step runs $\zeta$ sweeps over all sentences and stores the last $B$ of them, $\{\bm z^{(b)}\}_{b=1}^{B}$.
\paragraph{Stochastic approximation step}
Let $\tilde{\bm S}_t=\{(\tilde n_k,\tilde{\bm r}_k)\}_{k=1}^{K}$ collect the count and the resultant vector of the sentences assigned to each topic, averaged over the $B$ stored sweeps (Eq.~\eqref{eq:suffstats}). The statistics passed to the M-step are given by the stochastic approximation update
\begin{equation}
  \bm s_t = \bm s_{t-1} + \gamma_t\,\bigl(\tilde{\bm S}_t - \bm s_{t-1}\bigr),
  \qquad
  \gamma_t =
  \begin{cases}
    1 & t \le T_0,\\
    (t-T_0)^{-a} & t > T_0,
  \end{cases}
  \label{eq:saem}
\end{equation}
with a burn-in length $T_0$ and an exponent $a\in(\tfrac12,1]$, so that $\sum_t\gamma_t=\infty$ and $\sum_t\gamma_t^2<\infty$~\citep{delyon-1999-saem}. During the burn-in, $\bm s_t=\tilde{\bm S}_t$ and, as in the Monte Carlo EM~\citep{wei-1990-mcem} of \citet{li-2016-integrating}, every M-step reads the samples of the current iteration alone. Afterwards, the Monte Carlo error of the statistics is reduced by averaging across iterations.
\paragraph{M-step}
Let $n_k$ and $\bm r_k$ be the count and the resultant vector of topic $k$ in $\bm s_t$, and $\bar r_k=\|\bm r_k\|/n_k$.
The estimate of the expected complete-data log-likelihood then separates over topics into $\sum_{k}\{n_k\log C_M(\kappa_k)+\kappa_k\bm\mu_k^\top\bm r_k\}$, and maximizing over $\bm\mu_k$ subject to $\|\bm\mu_k\|=1$ and then over $\kappa_k$ using $\partial_\kappa\log C_M=-A_M$ gives an update of the form of the maximum likelihood estimate of \citet{banerjee-2005-clustering},
\begin{equation}
  \bm\mu_k = \frac{\bm r_k}{\|\bm r_k\|},
  \qquad
  A_M(\kappa_k) = \bar r_k .
  \label{eq:mstep}
\end{equation}
The second equation has no closed-form solution, so we solve it by Newton's method~\citep{sra-2012-note}, initialized at the approximation of \citet{banerjee-2005-clustering},
\begin{equation}
  \hat\kappa_k = \frac{\bar r_k\,(M - \bar r_k^2)}{1-\bar r_k^2},
  \label{eq:kappa-approx}
\end{equation}
which lies within a relative distance $2/(M-2)$ of the root by the bounds of \citet{tanabe-2007-parameter}, below one percent at the embedding dimensions used here ($M=384$ and $M=768$). From this start, Newton's method reaches machine precision in at most two iterations. A topic holding at most one sentence receives a fixed fallback concentration $\kappa_0$, and a topic left without sentences after an E-step is reseeded with one sentence. Neither occurred in the runs of the main experiments.

$\bm\alpha$ is updated by the fixed-point iteration of \citet{minka-2000-dirichlet} on the document-topic counts of the final sweep, since $\bm s_t$ carries no per-document counts, applied per component, so that the estimated $\bm\alpha$ is asymmetric, a prior structure known to be beneficial for the document-topic distributions~\citep{wallach-2009-rethinking}. The update is applied during the burn-in iterations $t\le T_0$, and $\bm\alpha$ is held fixed afterwards, because the convergence statement below assumes a fixed $\bm\alpha$. Estimating $\bm\alpha$ by marginal likelihood is the empirical Bayes treatment inherited from LDA~\citep{minka-2000-dirichlet,blei-2003-lda}, whereas $\bm\Delta$ parameterizes the likelihood and is point-estimated as in the EM algorithms of \citet{blei-2003-lda} and \citet{banerjee-2005-clustering} rather than integrated out as $\bm\theta$ is.

\paragraph{Convergence}
The averaging of Eq.~\eqref{eq:saem} makes a convergence statement available. For fixed $\bm\alpha$, the complete-data model is a curved exponential family in $\bm\Delta$ with the bounded sufficient statistics $(n_k,\bm r_k)_k$, a Gibbs sweep is a uniformly ergodic kernel on the finite space of assignments, and the M-step maximizes the estimated expected complete-data log-likelihood exactly. Hence, as long as, during the iterations, the averaged count $n_k$ of every topic stays above some constant larger than one, the concentration estimates stay below some constant smaller than the cap $\kappa_{\max}$ of Algorithm~\ref{alg:mcem}, and no topic is reseeded, the argument by which \citet{kuhn-2004-saem} prove the convergence of SAEM coupled with a Markov chain applies (\appref{app:estimation}), and the natural parameters $\kappa_k^{(t)}\bm\mu_k^{(t)}$ converge almost surely to the set of stationary points of the marginal likelihood at the fixed $\bm\alpha$. The statement concerns $t\to\infty$. Whether the training length of the main experiments suffices is assessed separately, by the plateau of the training log-likelihood (\appref{app:wallclock}). The argument covers neither the burn-in, during which $\bm\alpha$ is estimated, nor the Monte Carlo EM without averaging ($T_0=T$), which apart from the exact solution for $\kappa_k$ and the estimation of $\bm\alpha$ is the algorithm of \citet{li-2016-integrating}, whose Monte Carlo error is not reduced across iterations and for which we know of no comparable statement.

\subsection{Time Complexity}
\label{subsec:complexity}
Let $T_\alpha$ denote the number of fixed-point iterations of the $\bm\alpha$ update. The one-time cost of sentence encoding is excluded.
Because $\bm\Delta^{(t-1)}$ is held fixed throughout the E-step of iteration $t$, evaluating $\ell_{dik}$ for all sentence--topic pairs amounts to inner products between all sentence embeddings and all topic mean directions, $O(N_sKM)$ time per sweep in our implementation, and each Gibbs sweep adds $O(N_sK)$ time for combining them with the counts $n_{dk}^{-i}$ and sampling.
The E-step therefore costs $O(\zeta N_sKM)$, the M-step $O(BN_sM + KM)$ for the sufficient statistics, their averaging in Eq.~\eqref{eq:saem} and the updates, and the $\bm\alpha$ update $O(T_\alpha DK)$, so that one outer iteration costs $O(\zeta N_sKM + BN_sM + KM + T_\alpha DK)$. The E-step likelihood term dominates whenever $\zeta K \gg B$ and $\zeta N_sM \gg T_\alpha D$. Our settings satisfy both ($\zeta = 20$, $K \geq 10$, $B = 8$, $T_\alpha \leq 100$, $M \geq 384$, $N_s \geq D$), and the $\bm\alpha$ update runs only during the burn-in.
This linear dependence on $M$ contrasts, as \citet{li-2016-integrating} note for word embeddings, with full-covariance Gaussian formulations, whose density involves a quadratic form with the inverse covariance matrix and takes $O(M^2)$ per sentence--topic pair. One sweep of the collapsed sampler of Gaussian LDA and GSLDA~\citep{das-2015-gaussian,cha-2024-sentencelda} therefore costs $O(N_sKM^2)$ for the likelihood alone, even with the Cholesky-based updates of their samplers, against $O(N_sKM)$ for a sweep of vSLDA. The two per-pair likelihood costs thus differ by a factor of the order of $M$, an inner product of length $M$ against a triangular solve of $M(M+1)/2$ terms, which is several hundred at the dimensions of the sentence encoders used here ($M=384$ and $768$). Measured on one experimental unit with our own implementations of both models, built to the same structure and with the thread pools pinned to a single thread, doubling $M$ from 384 to 768 by exchanging MiniLM for MPNet multiplied the time of one Gibbs sweep by 1.4 for vSLDA and by 4.8 for GSLDA, where the time of a sweep includes the work each model does around its sweeps, and training GSLDA took 80 and 262 times as long as training vSLDA at the two dimensions (\appref{app:wallclock}).

\subsection{Expected Topic-Word Counts for Word-Based Evaluation}
\label{subsec:repwords}
Topic coherence and diversity require a ranked list of representative words per topic, usually read off a native topic-word distribution, which sentence-level models over embeddings do not possess. The GSLDA study notes that the conventional coherence and diversity measures are fitted to word-level models and evaluates by classification as a surrogate~\citep{cha-2024-sentencelda}.
We instead obtain expected topic-word counts from unit-level topic posteriors, by a construction that applies uniformly to every model whose generative process assigns each unit (a word token or a sentence) to a latent topic (LDA, GLDA, vLDA, SentLDA, GSLDA and vSLDA).
Write $u_{di}$ for unit $i$ of document $d$, a sentence for the sentence-level models and a word token for the word-level models, and $n(w,u_{di})$ for the number of occurrences of word type $w$ in that unit, so that $n(w,u_{di})=\mathbb{I}[x_{di}=w]$ for a word token $x_{di}$. These counts are determined by the text alone.
Let $c_w=\sum_{d,i} n(w,u_{di})$ be the corpus frequency of $w$ over the evaluation vocabulary, and $r_{dik}=p(z_{di}=k \mid \mathcal{D})$ the posterior probability, under the fitted parameters of the model held fixed, that unit $u_{di}$ is assigned to topic $k$, with $\sum_{k} r_{dik}=1$.
The expected counts weight the word frequencies of each unit by its topic posterior:
\begin{equation}
  q_{kw} \;=\; \sum_{d,i} r_{dik}\, n(w, u_{di}),
  \qquad
  q_k \;=\; \sum_{w} q_{kw}.
  \label{eq:induced-counts}
\end{equation}
Three properties follow directly: by linearity of expectation, $q_{kw}$ is the posterior expected number of occurrences of $w$ generated under topic $k$ when every token inherits the topic of its unit; because $\sum_k r_{dik}=1$ for every unit, the counts conserve token mass, $\sum_k q_{kw}=c_w$; and when every unit is a single token, $q_{kw}=\sum_{d,i} r_{dik}\,\mathbb{I}[x_{di}=w]$, the topic-word count of word-level LDA.
Representative words are the top-ranked words by the normalized pointwise mutual information (NPMI) between word and topic under the plug-in estimates $p_{\mathrm{ind}}(w,k)=q_{kw}/N_{\mathrm{tok}}$, $p_{\mathrm{ind}}(w)=c_w/N_{\mathrm{tok}}$ and $p_{\mathrm{ind}}(k)=q_k/N_{\mathrm{tok}}$, with $N_{\mathrm{tok}}=\sum_w c_w$.
The posteriors $r_{dik}$ are obtained by collapsed fold-in sampling with the fitted topic parameters frozen for every model that assigns a topic to an observed unit, and from the model's own posterior for ETM, ConTM and SAM, which do not.
\section{Experiments}
\label{sec:exp}
\subsection{Datasets}
\label{subsec:datasets}
We evaluate on two English corpora with coarse and fine category labels: 20 Newsgroups~\citep{lang-1995-newsweeder}, whose 19 topical newsgroups are the fine categories and are grouped into 6 coarse categories following \citet{cha-2024-sentencelda}, and the New York Times corpus (NYT) released by \citet{mekala-2021-coarse2fine}, collected through the New York Times API by \citet{meng-2018-westclass} and annotated with coarse and fine categories. After the exclusions described in \appref{app:category_mapping}, NYT keeps 4 coarse and 24 fine categories, and the corpora hold 17{,}153 and 11{,}642 documents. For 20 Newsgroups, headers, footers and quoted replies are removed to reduce the dependence of classification on metadata that leaks the newsgroup label. The removal is heuristic and not complete: names from the signatures of posters remain in the text.
Each corpus is split into training and test sets with a stratified 6:4 ratio. Only the training data are used for learning the topic models, and the test data only for evaluating the classifiers.
Each coarse category is one independent experimental unit. Its documents form the corpus on which every model is trained, and its fine categories are the classification targets, so that, as in the coarse-to-fine setting of \citet{mekala-2021-coarse2fine}, classification takes place within one coarse category, among fine categories whose surface vocabulary distributions are close. The ten units (six from 20 Newsgroups, four from NYT) hold between 983 and 8{,}629 documents and between 2 and 9 fine categories each (Table~\ref{tab:dataset_stats}).
Documents in NYT hold 23 to 26 sentences on average, whereas those in 20 Newsgroups are much shorter, at 6 to 13, a difference to which Section~\ref{subsec:coherence_doclen} returns.
The topic-count sweep of Section~\ref{subsec:topic_sweep} uses one further unit, the corpus as a whole, in which the fine categories of all coarse categories are the classification targets at once, 19 on 20 Newsgroups and 24 on NYT. On this unit the models are compared as the number of topics grows.

\subsection{Baselines and the Factorial Design}
\label{subsec:baselines}
We compare vSLDA with eight topic models, the six baselines of Table~\ref{tab:model_map} and two widely used neural topic models, ETM and ConTM, all of which infer the document-topic proportions $\bm\theta_d$ as latent variables of a generative process.
The comparison is restricted to models that infer document-level topic proportions under a probabilistic admixture, the class within which the observation unit and the likelihood can be varied separately. The clustering pipelines over embeddings of Section~\ref{sec:related} (BERTopic, Top2Vec, SenClu) are related alternatives outside this class about which the experiments make no claim.
\begin{table}[!t]
\centering
\footnotesize
\caption{Compared models by observation unit (rows) and observation distribution (columns). Each cell names the model of Section~\ref{sec:related} that we take as the representative of that combination. Other models discussed there fall in the same cells. \textsc{vSLDA} is the proposed model, and ``--'' marks a combination for which no model was discussed. The $O(\cdot)$ entries in the column headers give the cost of evaluating one observation's likelihood under one topic as a function of the embedding dimension $M$ (Section~\ref{subsec:complexity}).}
\label{tab:model_map}
\renewcommand{\arraystretch}{1.1}
\begin{tabular}{@{}l ccc@{}}
\toprule
 & \multicolumn{3}{c}{Observation distribution} \\
\cmidrule(l){2-4}
Observation unit & \makecell{Categorical\\{\scriptsize over the vocabulary}} & \makecell{Gaussian\\{\scriptsize over embeddings, $O(M^2)$}} & \makecell{vMF\\{\scriptsize over embeddings, $O(M)$}} \\
\midrule
Word & \makecell{LDA\\\colcite{\citealp{blei-2003-lda}}} & \makecell{GLDA\\\colcite{\citealp{das-2015-gaussian}}} & \makecell{vLDA\\\colcite{\citealp{li-2016-integrating}}} \\
\addlinespace
Sentence & \makecell{SentLDA\\\colcite{\citealp{balikas-2016-slda}}} & \makecell{GSLDA\\\colcite{\citealp{cha-2024-sentencelda}}} & \makecell{\textsc{vSLDA}\\\colcite{proposed}} \\
\addlinespace
Document & -- & -- & \makecell{SAM\\\colcite{\citealp{reisinger-2010-spherical}}} \\
\bottomrule
\end{tabular}
\end{table}
The six models in the word and sentence rows of Table~\ref{tab:model_map} form a factorial design: each column contrasts the observation unit under a fixed observation distribution (LDA $\rightarrow$ SentLDA, GLDA $\rightarrow$ GSLDA, vLDA $\rightarrow$ vSLDA), and each row contrasts the observation distribution on the same observations (GLDA $\rightarrow$ vLDA on word embeddings, GSLDA $\rightarrow$ vSLDA on sentence embeddings).
The three remaining models lie outside the factorial design: ETM~\citep{dieng-2020-topic} and ConTM~\citep{bianchi-2021-ctm} are included as widely used neural topic models, and SAM~\citep{reisinger-2010-spherical}, which observes the L2-normalized term-frequency vector of each document, is the document-level vMF admixture.
In what follows, LDA, SentLDA and SAM, whose observations are counts or term frequencies over the vocabulary, are referred to as vocabulary-based, and the models observing word or sentence embeddings as embedding-based.

\subsection{Evaluation Protocol and Metrics}
\label{subsec:metrics}
The main experiments cover the ten coarse-category units of Section~\ref{subsec:datasets} at $K\in\{10,20,30\}$. Calling one unit at one number of topics a \emph{condition}, there are 30 conditions, each reported as the mean and standard deviation over 5 independent runs that differ only in the training seed.
Since the three topic counts and the five runs are repeated measurements within a unit, inference across $K$ forms one effect per unit, the difference between two models averaged over $K$ and the paired runs, and tests it by exact enumeration of all $2^{10}$ within-unit sign flips of the ten effects. Formal tests are confined to five comparisons: the classification accuracy of vSLDA against SentLDA, the primary comparison of the design (the model of Table~\ref{tab:model_map} that differs from vSLDA in the observation distribution alone), and the four coherence comparisons of Section~\ref{subsec:coh_results} (vSLDA against SentLDA and GSLDA, and against ETM on each corpus), with a Holm correction over this family.
To summarize several conditions, we rank the models within each condition (1 = best, ties averaged) and report the mean rank over those conditions.
\paragraph{Classification performance}
The document-topic distributions $\bm\theta_d$ estimated by each model are used as features for a logistic regression classifier trained on the training data to predict the fine-category labels, and its accuracy on the test data is reported.
Document classification with the document-topic distributions as features is a standard evaluation of unsupervised topic models~\citep{balikas-2016-slda,li-2016-integrating}.
\paragraph{Sample efficiency}
The classification above is repeated with fewer labeled documents. Each topic model is trained once, without supervision, on the entire training corpus of a unit, so that $\bm\theta_d$ is identical across label fractions. Only the fraction of labeled documents used to train the classifier is reduced, over 5, 10, 20, 40, 60, 80 and 100\%, the subsets at each fraction below 100\% being drawn by stratified sampling over the fine categories five times per run.
\paragraph{Topic coherence}
Whether the top-$N$ representative words of a topic center on a single theme is measured by $C_V$~\citep{roder-2015-coherence}, which scores the word set by the similarity of the NPMI profiles of its words in a reference corpus. $C_V$ is the primary measure because it attains the highest correlation with human ratings among the measures evaluated in that work, and because it remains positive in our setting, which its product with topic diversity, defined below, requires. NPMI and $C_V$ are defined in \appref{app:coherence_npmi}, which also repeats the comparison with the NPMI-based measure $C_{\mathrm{NPMI}}$.
\paragraph{Topic diversity}
A model can attain high coherence while estimating several nearly identical topics, so we also report topic diversity (TD), the proportion of unique words among the top-$N'$ representative words of all $K$ topics~\citep{dieng-2020-topic}. The product of $C_V$ and TD is reported as \emph{topic quality} following \citet{dieng-2020-topic}.
\subsection{Implementation}
\label{subsec:implementation}
The implementation used in the experiments is publicly available, with command-line workflows for dataset preparation, experiment execution and evaluation. The source code, command-line workflows and experiment configuration files used in this work are publicly available at \url{https://github.com/izumi-lab/vSLDA}.
The primary sentence encoder is \path{sentence-transformers/all-MiniLM-L6-v2}\footnote{\url{https://huggingface.co/sentence-transformers/all-MiniLM-L6-v2}, revision \texttt{1110a24}.} (MiniLM, $M=384$), a Sentence-BERT-type bi-encoder~\citep{reimers-2019-sbert}. All tables and figures in the main text use MiniLM embeddings. \appref{app:encoders} repeats the main experiments with \texttt{all-mpnet-base-v2} and \texttt{bge-base-en-v1.5} (both $M=768$).
vSLDA is run with $T=10$ outer iterations of the SAEM of Section~\ref{subsec:inference}, a burn-in of $T_0=5$ iterations followed by step sizes $\gamma_t=(t-T_0)^{-1}$, under which $\bm s_T$ is the equal-weight mean of the statistics of the last $T-T_0$ iterations, $\zeta = 20$ Gibbs sweeps per E-step, $B = 8$ stored sweeps and $\alpha_0=50/K$ following \citet{griffiths-2004-finding}. The number of topics is $K\in\{10,20,30\}$ in the main experiments and is varied over $\{10,20,30,50,100,200,300\}$ on the corpus as a whole and up to $K=100$ on the coarse-category units in Section~\ref{subsec:topic_sweep}.
The sensitivity of vSLDA to these settings is examined in \appref{app:hyper_sensitivity}: varied one at a time, no setting changes the accuracy by more than $0.8$ points except the number of outer iterations and the burn-in of the stochastic approximation, which matter only when they are small.
Training longer than these settings changes neither vSLDA nor GSLDA: the accuracy and coherence of vSLDA are flat from $T=10$ to $T=30$, raising the GSLDA training sweeps from 10 to 20 changes its accuracy by at most $0.3$ points, and the training log-likelihood of both models reaches its plateau within the training length of the main experiments.
The number of representative words is $N=10$ for $C_V$ and $N'=25$ for topic diversity. The reference corpus is a random sample of $10^{6}$ articles of English Wikipedia.
The baseline settings, the common preprocessing, the fold-in sampler of Section~\ref{subsec:repwords} and the coherence implementation are given in \appref{app:hyperparams}, and the computing environment in \appref{app:wallclock}.
\section{Results}
\label{sec:results}

Table~\ref{tab:results_summary} summarizes the per-condition results by dataset and number of topics, each cell giving the mean of a model's scores over the categories of that dataset and its mean rank among the nine models. The per-condition tables are given in \appref{app:per_condition}.

\begin{table}[!t]
\centering
\footnotesize
\setlength{\tabcolsep}{4pt}
\caption{Overall comparison by dataset and number of topics: (a) classification accuracy (\%) of logistic regression with document-topic distributions as features, (b) topic coherence $C_V$, and (c) topic quality, the product of $C_V$ coherence and topic diversity. Each cell is the mean over the coarse categories of the dataset (6 for 20 Newsgroups, 4 for NYT) of the model's per-condition score, followed in parentheses by its mean rank among the nine compared models over the same conditions. The per-category values are those of Tables~\ref{tab:cls_logreg_minilm}, \ref{tab:topic_coherence_c_v} and \ref{tab:topic_quality} in \appref{app:per_condition}. Values are means over 5 independent runs. The sentence encoder is MiniLM. \textbf{Bold}/\underline{underline} indicate the best/second-best score within each column.}
\label{tab:results_summary}
\adjustbox{max width=\textwidth}{%
\begin{tabular}{@{}l *{3}{c}@{\hspace{7pt}} *{3}{c}@{}}
\toprule
\multirow[b]{3}{*}{Model} & \multicolumn{3}{c}{20news} & \multicolumn{3}{c}{NYT} \\
\cmidrule(lr){2-4} \cmidrule(lr){5-7}
 & 10 & 20 & 30 & 10 & 20 & 30 \\
 & \multicolumn{3}{c}{\scriptsize score (mean rank)} & \multicolumn{3}{c}{\scriptsize score (mean rank)} \\
\midrule
\multicolumn{7}{@{}l}{\textbf{(a) Classification accuracy}} \\
\addlinespace[1pt]
LDA {\scriptsize(Blei, 2003)} & \sr{60.08}{4.00} & \sr{62.10}{4.17} & \sr{63.25}{4.17} & \sr{65.91}{4.25} & \sr{69.57}{3.75} & \sr{70.93}{3.25} \\
SentLDA {\scriptsize(Balikas, 2016)} & \sr{\underline{66.14}}{2.17} & \sr{\underline{67.79}}{2.17} & \sr{\underline{67.45}}{2.33} & \sr{\underline{75.80}}{2.00} & \sr{\underline{83.56}}{2.00} & \sr{\underline{82.94}}{2.00} \\
SAM {\scriptsize(Reisinger, 2010)} & \sr{45.18}{7.17} & \sr{51.98}{6.17} & \sr{53.51}{6.50} & \sr{64.77}{3.75} & \sr{70.99}{3.50} & \sr{68.19}{4.50} \\
GLDA {\scriptsize(Das, 2015)} & \sr{44.74}{7.83} & \sr{47.92}{8.17} & \sr{48.33}{8.33} & \sr{43.01}{7.88} & \sr{41.75}{8.38} & \sr{41.51}{8.62} \\
vLDA {\scriptsize(Li, 2016)} & \sr{46.01}{7.50} & \sr{50.75}{7.17} & \sr{54.08}{6.50} & \sr{41.64}{8.00} & \sr{45.35}{7.75} & \sr{46.01}{7.00} \\
ETM {\scriptsize(Dieng, 2020)} & \sr{63.50}{3.33} & \sr{62.26}{4.83} & \sr{63.42}{4.67} & \sr{58.42}{5.88} & \sr{60.67}{5.88} & \sr{59.32}{6.62} \\
ConTM {\scriptsize(Bianchi, 2021)} & \sr{53.20}{5.83} & \sr{58.15}{4.83} & \sr{61.92}{4.67} & \sr{61.69}{5.00} & \sr{66.24}{5.25} & \sr{68.37}{4.25} \\
GSLDA {\scriptsize(Cha, 2024)} & \sr{55.15}{6.17} & \sr{54.84}{6.50} & \sr{53.99}{6.83} & \sr{55.35}{6.75} & \sr{57.71}{7.00} & \sr{54.73}{7.25} \\
\textsc{vSLDA} {\scriptsize(proposed)} & \sr{\textbf{71.25}}{1.00} & \sr{\textbf{72.96}}{1.00} & \sr{\textbf{73.62}}{1.00} & \sr{\textbf{80.95}}{1.50} & \sr{\textbf{86.42}}{1.50} & \sr{\textbf{87.46}}{1.50} \\
\midrule
\multicolumn{7}{@{}l}{\textbf{(b) Topic coherence}} \\
\addlinespace[1pt]
LDA {\scriptsize(Blei, 2003)} & \sr{0.351}{5.67} & \sr{0.306}{6.67} & \sr{0.300}{6.33} & \sr{0.456}{5.00} & \sr{0.404}{5.75} & \sr{0.384}{4.75} \\
SentLDA {\scriptsize(Balikas, 2016)} & \sr{0.377}{5.00} & \sr{0.323}{5.67} & \sr{0.294}{6.33} & \sr{\underline{0.529}}{2.25} & \sr{0.481}{2.50} & \sr{0.427}{3.50} \\
SAM {\scriptsize(Reisinger, 2010)} & \sr{0.351}{6.50} & \sr{0.340}{5.33} & \sr{0.337}{5.00} & \sr{0.493}{3.50} & \sr{0.440}{4.75} & \sr{0.368}{6.50} \\
GLDA {\scriptsize(Das, 2015)} & \sr{0.410}{3.67} & \sr{0.390}{3.67} & \sr{0.380}{3.50} & \sr{0.410}{6.00} & \sr{0.389}{5.50} & \sr{0.379}{5.00} \\
vLDA {\scriptsize(Li, 2016)} & \sr{0.399}{4.00} & \sr{0.403}{3.17} & \sr{0.400}{3.33} & \sr{0.379}{7.00} & \sr{0.359}{7.25} & \sr{0.353}{7.00} \\
ETM {\scriptsize(Dieng, 2020)} & \sr{\textbf{0.530}}{1.67} & \sr{\textbf{0.511}}{1.33} & \sr{\textbf{0.512}}{1.00} & \sr{0.485}{3.75} & \sr{\underline{0.486}}{2.50} & \sr{\underline{0.476}}{1.75} \\
ConTM {\scriptsize(Bianchi, 2021)} & \sr{0.221}{9.00} & \sr{0.230}{8.83} & \sr{0.231}{8.83} & \sr{0.318}{8.75} & \sr{0.331}{8.00} & \sr{0.326}{7.50} \\
GSLDA {\scriptsize(Cha, 2024)} & \sr{0.289}{7.67} & \sr{0.246}{8.17} & \sr{0.242}{8.17} & \sr{0.380}{7.50} & \sr{0.344}{7.75} & \sr{0.320}{7.75} \\
\textsc{vSLDA} {\scriptsize(proposed)} & \sr{\underline{0.520}}{1.83} & \sr{\underline{0.460}}{2.17} & \sr{\underline{0.434}}{2.50} & \sr{\textbf{0.608}}{1.25} & \sr{\textbf{0.556}}{1.00} & \sr{\textbf{0.516}}{1.25} \\
\midrule
\multicolumn{7}{@{}l}{\textbf{(c) Topic quality}} \\
\addlinespace[1pt]
LDA {\scriptsize(Blei, 2003)} & \sr{0.350}{5.50} & \sr{0.300}{5.67} & \sr{0.290}{5.50} & \sr{0.454}{4.25} & \sr{0.384}{5.25} & \sr{0.347}{4.25} \\
SentLDA {\scriptsize(Balikas, 2016)} & \sr{0.377}{4.33} & \sr{0.322}{4.67} & \sr{0.292}{5.17} & \sr{\underline{0.528}}{2.25} & \sr{\underline{0.480}}{2.00} & \sr{\underline{0.422}}{2.50} \\
SAM {\scriptsize(Reisinger, 2010)} & \sr{0.349}{5.67} & \sr{0.334}{4.50} & \sr{0.327}{4.33} & \sr{0.491}{3.50} & \sr{0.431}{3.75} & \sr{0.273}{7.00} \\
GLDA {\scriptsize(Das, 2015)} & \sr{0.410}{3.33} & \sr{0.390}{3.00} & \sr{0.380}{2.83} & \sr{0.410}{5.50} & \sr{0.389}{4.50} & \sr{0.379}{3.50} \\
vLDA {\scriptsize(Li, 2016)} & \sr{0.295}{6.50} & \sr{0.242}{7.50} & \sr{0.218}{7.67} & \sr{0.273}{8.25} & \sr{0.219}{8.50} & \sr{0.199}{8.25} \\
ETM {\scriptsize(Dieng, 2020)} & \sr{\textbf{0.524}}{1.67} & \sr{\textbf{0.479}}{1.67} & \sr{\textbf{0.452}}{1.67} & \sr{0.456}{4.50} & \sr{0.412}{4.75} & \sr{0.369}{3.75} \\
ConTM {\scriptsize(Bianchi, 2021)} & \sr{0.167}{9.00} & \sr{0.155}{9.00} & \sr{0.142}{9.00} & \sr{0.295}{8.25} & \sr{0.273}{7.75} & \sr{0.246}{7.25} \\
GSLDA {\scriptsize(Cha, 2024)} & \sr{0.288}{7.50} & \sr{0.238}{7.50} & \sr{0.226}{7.33} & \sr{0.353}{7.25} & \sr{0.286}{7.50} & \sr{0.240}{7.25} \\
\textsc{vSLDA} {\scriptsize(proposed)} & \sr{\underline{0.520}}{1.50} & \sr{\underline{0.458}}{1.50} & \sr{\underline{0.430}}{1.50} & \sr{\textbf{0.608}}{1.25} & \sr{\textbf{0.553}}{1.00} & \sr{\textbf{0.509}}{1.25} \\
\bottomrule
\end{tabular}}
\end{table}

\subsection{Classification Performance}
\label{subsec:cls_results}
\emph{vSLDA attains the best mean rank across the 30 conditions and the best accuracy in 27 of them. LDA is the best in the three remaining conditions, all on NYT \textit{sports}.}
There, LDA is ahead by $7.3$, $1.7$ and $0.5$ points at $K=10$, $20$ and $30$ (Table~\ref{tab:cls_logreg_minilm}; Section~\ref{subsec:when_lda}).
The mean rank of vSLDA over the 30 conditions is $1.20$, followed by SentLDA ($2.13$) and LDA ($3.97$). These values pool the conditions behind the six columns of Table~\ref{tab:results_summary}(a), so each column enters with the weight of its number of categories (6 for 20 Newsgroups, 4 for NYT) rather than equally.
vSLDA is ahead of the best baseline of each condition in all five paired runs in 20 of the 30 conditions. At the level of the ten experimental units, with the difference to SentLDA averaged over the three topic counts and the paired runs (Section~\ref{subsec:metrics}), vSLDA is ahead on 9 of the 10 units, by $4.96$ points on average (median $4.23$), the exception being NYT \textit{sports} ($-1.29$ points); the exact within-unit sign-flip test gives $p=0.004$ ($p=0.012$ after the Holm correction). The lead is the same at each topic count taken separately, vSLDA being ahead on at least 9 of the 10 units at every $K$ ($p\le0.006$). Against each of the other seven baselines it is ahead on at least 9 of the 10 units.
The factorial design of the baselines separates the two design choices of vSLDA.
Averaged over the 30 conditions, replacing words with sentences as the observation unit improves accuracy by $8.0$ points under the categorical likelihood (LDA $\rightarrow$ SentLDA, positive in 27 of 30 conditions) and by $29.6$ points under the vMF likelihood (vLDA $\rightarrow$ vSLDA, positive in all 30), while replacing the full-covariance Gaussian likelihood with the vMF likelihood improves accuracy by $2.9$ points for word embeddings (GLDA $\rightarrow$ vLDA, positive in 22 of 30) and by $22.4$ points for sentence embeddings (GSLDA $\rightarrow$ vSLDA, positive in all 30). Under the Gaussian likelihood, the unit effect (GLDA $\rightarrow$ GSLDA) is $10.1$ points, positive in 21 of 30.
Each design choice thus has a larger effect in the presence of the other, which indicates that both choices are needed. What each contributes on its own is examined in Section~\ref{subsec:why_vmf}. The GSLDA column is the full-covariance model of Section~\ref{subsec:baselines}. Its low accuracy is analyzed in Section~\ref{subsec:why_gslda}, which also refits it with a diagonal and an isotropic covariance.
SAM, the document-level vMF admixture outside the factorial design, is behind vSLDA in all 30 conditions and ahead of LDA in 9.
With a nonlinear classifier the ordering by mean rank is unchanged except for two nearly tied pairs of baselines (vLDA and GSLDA, ConTM and ETM) (\appref{app:svm}). It is confirmed under two further sentence encoders (\appref{app:encoders}).

\paragraph{Sample efficiency}
\label{subsec:sample_eff}
\emph{The classification advantage largely persists as the labeled fraction decreases, except on NYT \textit{sports} and at the smallest fractions of NYT \textit{business} and \textit{politics}, where LDA or SentLDA leads.}
Over the 210 combinations of dataset, category, topic count and labeled fraction, vSLDA is the most accurate model in 175.
The per-category accuracy is plotted against the labeled fraction at $K=10$, $20$ and $30$, with MiniLM embeddings and logistic regression only (\appref{app:sample_eff}).
With 5\% of the labels vSLDA retains at least $88$\% of its full-supervision accuracy on every 20 Newsgroups unit, whereas on NYT the figure falls as low as $54$\%. On 20 Newsgroups, ETM loses more accuracy than the sentence-level models as labels are removed.
On NYT \textit{business} and \textit{politics}, SentLDA is the most accurate model at the smallest fractions at all three topic counts (at 5\% on both units and, on \textit{business}, up to 40\% of the labels at $K=30$) and vSLDA takes the lead only as more labels are added. On NYT \textit{sports}, LDA or SentLDA is ahead of vSLDA at every fraction and topic count, by $5.8$ to $7.3$ points at $K=10$, at most $3.0$ at $K=20$ and at most $1.4$ at $K=30$.

\subsection{Topic Coherence and Diversity}
\label{subsec:coh_results}
\emph{vSLDA and ETM are the two leading models on coherence, vSLDA ahead on NYT and ETM on 20 Newsgroups. The diversity of vSLDA stays near its maximum while that of ETM falls as $K$ grows, so vSLDA leads on the product in most conditions of both corpora.}
Panels~b and~c of Table~\ref{tab:results_summary} summarize $C_V$ coherence and topic quality under the extraction protocol of Section~\ref{subsec:repwords}.
On coherence, vSLDA and ETM attain the highest score in 16 and 13 of the 30 conditions, respectively, and the split follows the dataset. vSLDA is the highest in 10 of the 12 NYT conditions, whereas ETM is the highest in 12 of the 18 20 Newsgroups conditions (Section~\ref{subsec:coherence_doclen}). SentLDA takes the one remaining condition.
Within the sentence-level models, vSLDA is higher than SentLDA in 29 of the 30 conditions and than GSLDA in all of them, and on every one of the ten experimental units, by $0.116$ and $0.212$ on average (medians $0.117$ and $0.206$; exact $p=0.002$ against both, $p=0.010$ after the Holm correction; Section~\ref{subsec:metrics}). Against ETM, vSLDA is higher on all four NYT units, by $0.078$ on average (exact $p=0.125$, the smallest value four units can attain), and on two of the six 20 Newsgroups units, where ETM is ahead by $0.046$ on average (exact $p=0.219$); after the Holm correction both comparisons give $p=0.250$. ConTM and GSLDA are below all three vocabulary-based models in 28 of the 30 conditions.
On topic diversity, vSLDA and SentLDA each reach $0.98$ or above in 29 of the 30 conditions, whereas the diversity of ETM on 20 Newsgroups declines from at least $0.970$ at $K=10$ to between $0.804$ and $0.964$ at $K=30$. vLDA never exceeds $0.801$.
Taking the product therefore changes the 20 Newsgroups comparison. vSLDA attains the highest topic quality in 12 of the 18 conditions, against 6 under coherence alone. ETM remains ahead only on \textit{computer} and \textit{sports}, where its coherence advantage outweighs its lower diversity. On NYT, every row is led by the same model as under coherence alone.
The ordering within the sentence-level models and the direction of the split with ETM are the same under the NPMI-based measure (\appref{app:coherence_npmi}) and under the two further sentence encoders.

\subsection{Whole-Corpus Results as the Number of Topics Grows}
\label{subsec:topic_sweep}
\begin{figure}[!t]
\centering
\includegraphics{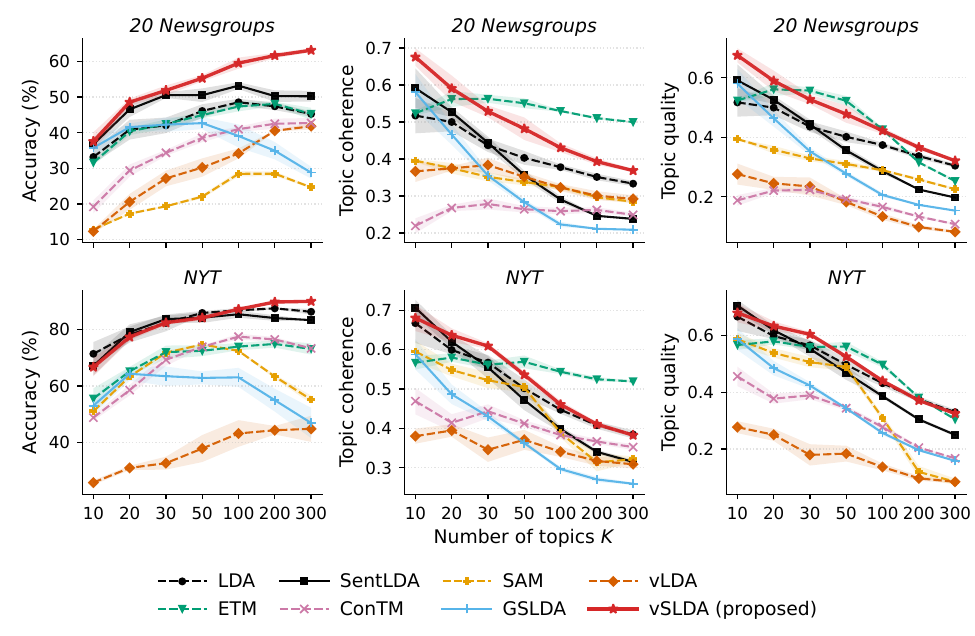}
\caption{\textbf{Topic-count sweep on the whole corpus (MiniLM).} Rows: 20 Newsgroups (top) and NYT (bottom); columns: the accuracy (\%) of fine-category classification by logistic regression on the document-topic distributions, $C_V$ topic coherence and topic quality, the product of $C_V$ and topic diversity (the diversity panels are in \appref{app:topic_sweep_categories}). The horizontal axis is the number of topics $K$. Each point is the mean over 5 runs ($n=5$), shaded bands are $\pm1$ standard deviation, and the eight models with runs across the whole grid are compared. Solid lines are the sentence-level models (SentLDA, GSLDA, vSLDA) and dashed lines the word- and document-level models.}
\label{fig:topic_sweep}
\end{figure}
\emph{On the corpus as a whole, the classification advantage of vSLDA grows with the number of topics: on 20 Newsgroups its margin widens at every step, and on NYT, where LDA or SentLDA leads up to $K=50$, vSLDA is ahead of every baseline from $K=100$ onward. On coherence, ETM overtakes vSLDA beyond the smallest counts, whereas on topic quality vSLDA comes back at large topic counts, leading at $K\ge200$ on 20 Newsgroups and drawing level with the leaders at $K=300$ on NYT.}
Figure~\ref{fig:topic_sweep} sweeps $K$ over $\{10,20,30,50,100,200,300\}$ on each corpus as a whole. GLDA is not part of this analysis: its full-covariance sampler costs $O(M^2)$ per word--topic pair and grows linearly with $K$ and with the number of word tokens, so we did not run it over the whole-corpus grid. The remaining eight models have runs across the grid.
On 20 Newsgroups, vSLDA attains the highest accuracy at every topic count, its margin over the best baseline (SentLDA throughout) widening from $0.8$ points at $K=10$ to $12.9$ at $K=300$. LDA, SentLDA, SAM and ETM peak at $K=100$ or $K=200$ and then decline, whereas vSLDA improves at every step of the grid.
On NYT, LDA or SentLDA is the most accurate model up to $K=50$, vSLDA being behind by $4.7$ points at $K=10$ and by $1.3$ to $1.7$ at $K=20$, $30$ and $50$, less than the run-to-run standard deviation of SentLDA at $K=20$ and $K=30$. From $K=100$ onward vSLDA is ahead of every baseline, by $0.4$ points over LDA at $K=100$, $2.2$ at $K=200$ and $3.7$ at $K=300$.
$C_V$ coherence falls with $K$ for LDA, SentLDA, SAM, GSLDA and vSLDA (vSLDA from $0.676$ to $0.369$ on 20 Newsgroups and from $0.680$ to $0.383$ on NYT) but is nearly flat for ETM, so the split of Section~\ref{subsec:coh_results} does not persist across the grid. vSLDA is the most coherent model at $K=10$ and $20$ on 20 Newsgroups and at $K=20$ and $30$ on NYT, SentLDA at $K=10$ on NYT and ETM at every larger count.
Topic diversity falls far more steeply for vLDA and the two neural models (on 20 Newsgroups and NYT, ETM to $0.506$ and $0.586$ at $K=300$, against $0.871$ and $0.848$ for vSLDA). On topic quality vSLDA therefore leads at $K\le20$ and $K\ge200$ on 20 Newsgroups and at $K=20$ and $30$ on NYT, with SentLDA ahead at $K=10$, LDA at $K=300$ and ETM in between. The coherence advantage of ETM thus carries over to topic quality only over a band of intermediate topic counts.

\FloatBarrier
\section{Discussion}
\label{sec:discussion}
This section interprets the results of Section~\ref{sec:results}: what each of the sentence unit and the vMF likelihood contributes (Section~\ref{subsec:why_vmf}); why the coherence comparison with ETM depends on document length and on the number of topics (Section~\ref{subsec:coherence_doclen}); why the full-covariance Gaussian formulations underperform (Section~\ref{subsec:why_gslda}); how vSLDA and SentLDA subdivide a subject into topics and what the concentration parameter indicates (Section~\ref{subsec:kappa_hierarchy}); and when bag-of-words LDA remains sufficient (Section~\ref{subsec:when_lda}).

\subsection{Contributions of the Sentence Unit and the vMF Likelihood}
\label{subsec:why_vmf}
The two evaluation settings of Section~\ref{sec:results}, classification within the coarse categories and modeling of the whole corpus as the number of topics grows, are the settings in which the topics to be distinguished are expected to share much of their vocabulary (Section~\ref{sec:intro}). In both, vSLDA leads. On classification accuracy, it attains the best mean rank and the best accuracy in 27 of the 30 within-category conditions. On the whole corpus it is ahead of every baseline at every topic count on 20 Newsgroups and from $K=100$ onward on NYT, with a margin that grows with $K$ on both corpora. It also leads on topic quality in most within-category conditions and, on coherence alone, is above the sentence-level baselines on every unit (Section~\ref{subsec:coh_results}). The results thus indicate that vSLDA estimates such topics, and the factorial design of Table~\ref{tab:model_map} indicates that both of its design choices are needed. With the vMF likelihood but words as the observation unit, vLDA has a mean rank between $6.5$ and $8.0$ in the six columns of Table~\ref{tab:results_summary}(a) and is among the three least accurate models at every topic count of the sweep (Figure~\ref{fig:topic_sweep}). With sentences as the unit but the categorical likelihood, SentLDA is the strongest baseline (mean rank $2.0$ to $2.3$), yet it is behind vSLDA on 9 of the 10 coarse-category units and, on 20 Newsgroups, falls further behind at every step of the sweep, from $0.8$ points at $K=10$ to $12.9$ at $K=300$.

The two choices contribute in different ways. On its own, the sentence unit brings SentLDA to second place, for the reason that motivated sentence-level LDA~\citep{balikas-2016-slda}: a sentence is assigned one topic as a whole, so the document-topic distribution is estimated from units that are topically more homogeneous than words. The vMF likelihood on its own does not, and its effect is large only on sentence embeddings, $22.4$ points against $2.9$ on word embeddings (Section~\ref{subsec:cls_results}). For sentence embeddings the direction is the component that training constrains to reflect meaning (Section~\ref{sec:intro}). The sentence-level Gaussian model does not realize the same gain, which Section~\ref{subsec:why_gslda} relates to the covariance matrix that it estimates per topic.

\subsection{Document Length, the Number of Topics and the Coherence Comparison with ETM}
\label{subsec:coherence_doclen}
Of the compared models, vSLDA and ETM are the two most coherent, and which of the two is ahead depends on the dataset and on the number of topics (Sections~\ref{subsec:coh_results} and~\ref{subsec:topic_sweep}). We restate the two dependences and then propose an account from the side of each model.
The dataset dependence coincides with document length. Of the ten experimental units (the coarse categories of Table~\ref{tab:dataset_stats}), vSLDA is behind ETM (averaged over the three topic counts) on the three with the fewest sentences per document (20 Newsgroups \textit{ride}, \textit{sports} and \textit{computer}, at $6.4$ to $8.4$), within $0.005$ of ETM on the other three 20 Newsgroups units and ahead on all four NYT units, which have $23$ to $26$ sentences per document.
The dependence on the number of topics is the same on both corpora. The coherence of vSLDA falls with $K$ whereas that of ETM is nearly flat, so ETM overtakes vSLDA beyond the smallest topic counts on the whole corpora and on a growing number of the coarse-category units (Section~\ref{subsec:topic_sweep}).

On the side of vSLDA, both dependences can be read from the extraction of Section~\ref{subsec:repwords}. Because the expected counts of Eq.~\eqref{eq:induced-counts} are sums over sentence assignments, all words of a sentence enter one topic together, whatever their subject, and each assignment is informed by the rest of the document only through the document-topic counts of Eq.~\eqref{eq:gibbs}.
When documents have many sentences, the assignments are made under informative document-topic counts and the likelihood pools lexically distinct sentences with similar directions, so the sums concentrate the expected counts on semantically related words with similar co-occurrence profiles. This is the property that $C_V$ rewards, and it is consistent with vSLDA being more coherent in nearly all conditions than SentLDA, which pools sentences only through shared words.
When documents have few sentences, the assignment rests on the sentence alone, and a sentence of mixed subject carries all of its words into one topic. A single-sentence document has no document context at all. Such documents account for $8$ to $12$\% of the documents in the 20 Newsgroups units but for under $1$\% in the NYT units.
Across the ten experimental units, the coherence difference between vSLDA and ETM correlates with sentences per document at $r \approx 0.8$ (a correlation carried largely by the contrast between the two corpora), in the direction this reading predicts.
As $K$ grows, each topic rests on fewer assignments and its representative words become less stable, as they do for LDA, SentLDA and SAM (Section~\ref{subsec:topic_sweep}).

On the side of ETM, the topic-word distribution is parameterized by inner products between topic and word embeddings~\citep{dieng-2020-topic}, so it is unaffected by how words are grouped into sentences, and the representative words of a topic are the neighbors of its embedding in the word-embedding space, which \citet{dieng-2020-topic} report to remain coherent as the vocabulary grows and which remain cohesive as $K$ grows in our sweep.
Under the same parameterization, however, the topic embeddings increasingly share neighborhoods as $K$ grows, which would explain why the diversity of ETM falls with $K$, already at $K=30$ on 20 Newsgroups and steeply at the upper end of the sweep (Sections~\ref{subsec:coh_results} and~\ref{subsec:topic_sweep}).

\subsection{Collapse of the Full-Covariance Gaussian Formulations}
\label{subsec:why_gslda}

This subsection examines why GSLDA falls far behind vSLDA in Section~\ref{sec:results}. Each topic of GSLDA is a full-covariance Gaussian with $M(M+1)/2$ free covariance parameters ($7.4\times10^{4}$ at $M=384$), estimated from the sentences assigned to the topic. GSLDA and vSLDA observe the same normalized embeddings, and the comparisons below attribute the gap between them to that covariance structure. The ratio of observations to dimensions, introduced first, sets how large the gap is and whether the topics collapse, but the gap remains where the ratio is large.
Writing $n$ for the median number of observations per non-empty topic of a run, the ratio $n/M$ states how many observations are available per dimension (\appref{app:n_over_m}). For example, on 20 Newsgroups/\textit{computer} at $K=20$, GSLDA has $n=267$ sentences against $M=384$ ($n/M=0.69$).
With $n<M$ observations, the posterior predictive of a topic is set by the prior in at least $M-n$ directions, at a scale that depends on $n$ alone and that no single scale matrix of the covariance prior matches at every topic size. A sentence is therefore less likely under a topic of a few hundred sentences than under a much larger one with the same distribution, and the assignments drift toward the topics that are already large.

Across the 30 conditions, $n/M$ stays below $1$ for GSLDA in 24 of them, that is, the sample covariance of a typical topic is singular, and the accuracy gap to vSLDA widens as the ratio falls. Over these 24 conditions the gap averages $24.3$ points and the largest topic holds $58.5\%$ of the sentences, against $14.5$ points and $21.4\%$ over the six conditions with $n/M\ge1$. Over the $450$ runs obtained by also varying the sentence encoder, $n/M$ correlates with the gap of the same run at Spearman $\rho=-0.55$ (Figure~\ref{fig:n_over_m}).
The ratio thus sets the regime of the collapse and the size of the gap, but does not account for its presence. On NYT/\textit{sports} at $K=10$, the condition with the most observations per dimension among the 30 ($n/M=30.8$), the full-covariance runs show none of the concentration of the collapse, yet GSLDA trails vSLDA by $17.5$ points. The same holds when GSLDA is fitted to each corpus as a whole at $K\in\{10,20,30\}$, where $n/M$ lies between $3.3$ and $41.1$: no topic is empty and the largest holds at most $16.9\%$ of the sentences, yet GSLDA trails vSLDA by $1.8$ to $18.9$ points, and on both corpora the gap is largest at $K=30$ (Table~\ref{tab:whole_corpus_gslda}). \appref{app:size_bias} suggests that the predictive of a full-covariance topic remains short of its large-sample limit at topic sizes comparable to the largest in our data. Whether this remaining gap is a matter of the prior scale or a property of the covariance structure is what the prior sweep and the refits below test.

The deficit is not removed by rescaling the covariance prior. Sweeping the prior scale $\lambda$, which sets the scale matrix of the inverse-Wishart prior on the topic covariances to $\lambda\bm I$, over $\{0.01, 0.1, 1, 3, 10, 30, 3M\}$ at $K=20$ (Table~\ref{tab:prior_scale}) places the accuracy optimum on both datasets at the value of the main experiments, $\lambda=0.1$, where GSLDA remains $18.6$ and $28.3$ points below vSLDA on 20 Newsgroups and NYT.
A smaller prior scale lowers the accuracy, and a larger one lowers it more sharply: at $\lambda=1$ most of the topics are already empty and the accuracy falls by $15.1$ and $16.7$ points. From $\lambda=10$ upwards, including the default $\lambda=3M$ of the reference implementation, $19$ of the $20$ topics are empty in every run, consistent with the report by \citet{cha-2024-sentencelda} that the default prior of the implementation does not work.

To test the attribution to the covariance structure directly, we refit GSLDA with a diagonal and with an isotropic topic covariance, leaving the prior mean, the predictive width of an empty topic and every other setting unchanged (Table~\ref{tab:covariance}).
Over the 30 conditions, the mean accuracy gap to vSLDA falls from $22.4$ points under the full covariance to $0.8$ under the diagonal and $0.0$ under the isotropic structure, which is ahead of vSLDA in $10$ of the $30$. On NYT/\textit{sports} at $K=10$, the condition without the collapse, the isotropic refit brings the gap to $0.2$ points. The concentration of the collapse disappears as well, the largest topic holding $51.0\%$ of the sentences on average under the full covariance against $11.1\%$, with no empty topics, under the isotropic one.
The diagonal refit, close to vSLDA in accuracy, still leaves $30.2\%$ of the $K$ topics empty on average, against $4.0\%$ under the full covariance, so only the isotropic structure removes the deficit and the empty topics together.
The number of variance parameters estimated per topic predicts this ordering, full $\ll$ diagonal $<$ isotropic $\approx$ vMF. All three structures and vSLDA observe the L2-normalized embeddings. On unit vectors an isotropic Gaussian with fixed parameters is a vMF density up to renormalization (\appref{app:size_bias}). It is therefore reasonable that the isotropic refit, the Gaussian form closest to the vMF likelihood, reaches the accuracy of vSLDA, although the two models differ in their priors and estimators.
The directional restriction itself is not what the comparison finds dispensable. Repeated on the encoder output taken before the normalization, the isotropic refit falls $3.7$ points behind vSLDA and is ahead of it in none of the 30 conditions (Table~\ref{tab:covariance}). On the same input the full covariance still collapses, falling $30.1$ points behind vSLDA with $63.3\%$ of the sentences in its largest topic on average. The parity thus appears to rest on the directional restriction that the vMF likelihood builds in, whereas the collapse of the full model is attributable to its covariance structure rather than to the normalization.

Our GSLDA implements the model as \citet{cha-2024-sentencelda} specify it, a Gaussian with a full covariance matrix per topic under an inverse-Wishart prior. The accuracy it reaches is well below the values they report for that model (a mean of $71.9$ over the six coarse categories of 20 Newsgroups at $K=20$). The implementation released with that study, in the version we retrieved, evaluates the quadratic form of the predictive with a diagonal scale matrix in place of the full covariance estimate (Eq.~\eqref{eq:ref_diag}, \appref{app:reference_code}). Run as released on 20 Newsgroups/\textit{sports} at $K=20$, it reaches $89.1 \pm 0.9$ over five runs, in agreement with the value published for that unit ($88.4 \pm 0.6$). With the full covariance estimate used in that quadratic form and nothing else changed, it reaches $56.1 \pm 1.2$, the accuracy level of our GSLDA on that unit ($53.1 \pm 0.8$ on our split; Table~\ref{tab:reference_code}).
We therefore read the published values as those of a sampler whose predictive uses a diagonal scale matrix rather than the full covariance estimate, a reading consistent with the diagonal and isotropic refits of Table~\ref{tab:covariance}, but only as an inference, since we cannot verify that the code we retrieved is the code behind the published values. The reading does not touch the central claim of the study, that sentence-level topic models yield more discriminative document representations than word-level ones, which our own results support under a different likelihood. We take the deficit of GSLDA to be a property of the full-covariance model rather than of our implementation.

\subsection{Estimated Topics and the Concentration Parameter}
\label{subsec:kappa_hierarchy}
This subsection examines the topics that vSLDA and SentLDA estimate on one run per dataset (20 Newsgroups/\textit{computer} and NYT/\textit{arts}, $K=20$, run 0), and asks what the fitted concentration parameter $\kappa_k$ indicates. The all-pair statistics of the three sentence-level models in the embedding space are reported in \appref{app:topic_pairs}. Because vSLDA is fitted in that space, the comparison below does not rest on them.
Table~\ref{tab:family_topics} shows, for each run, one subject that SentLDA represents by one topic and vSLDA divides into several. The topics of the two models are matched by the overlap of the sentences assigned to them. The rules of the matching and of the selection are given in \appref{app:topic_pairs}. Each column gives, for one vSLDA topic, the fitted $\kappa$, $H$ and its top five representative words. $H$ is the normalized entropy of the distribution of the sentences of the topic over the fine categories: it is $0$ when all of them belong to one fine category and $1$ when they spread evenly over the fine categories.
On 20 Newsgroups/\textit{computer} (panel~a), one SentLDA topic of user-interface events (\textit{cursor}, \textit{icon}, \textit{event}) receives two vSLDA topics, \textit{X11/Unix windowing} and \textit{UI events/widgets}.
On NYT/\textit{arts} (panel~b), SentLDA estimates one orchestral-music topic, and the three vSLDA topics matched to it separate \textit{opera}, \textit{vocal repertoire} and \textit{music festivals}.

The two models read different information about a sentence. SentLDA reads its words, and vSLDA reads its embedding. Because the embedding encodes the meaning of a sentence, sentences that use different words can have similar directions. The estimated topics contain examples consistent with this difference. In the 20 Newsgroups/\textit{computer} run, SentLDA has one topic that collects the words of file paths (\textit{usr}, \textit{lib}, \textit{bin}) and one of the names of Unix software and vendors (\textit{sun}, \textit{osf}, \textit{motif}). Both are matched to a single vSLDA topic, \textit{X11/Unix windowing} of panel~a (\textit{xterm}, \textit{xdm}). The representative words of the former are the strings that make up a path rather than words that name a subject. Similarly, in the NYT/\textit{arts} run, the SentLDA topic that collects the words of venue listings (\textit{org}, \textit{avenue}, \textit{com}) is matched to the vSLDA topic of \textit{music festivals}. These readings are qualitative and rest on the representative words of one run per dataset, and the matching rests on the overlap of a part of the assigned sentences. The examples are therefore consistent with the view that vSLDA can hold in one topic, by subject, sentences that SentLDA separates along the use of words.
\begin{table}[!t]
\centering
\caption{Groups of \textsc{vSLDA} topics matched to one SentLDA topic ($K=20$, MiniLM, run 0; every topic of both runs is listed in \appref{app:topic_pairs}). In each of the two runs, every \textsc{vSLDA} topic is matched to the SentLDA topic that receives the largest expected co-assignment of its sentences (not necessarily a majority of them; \appref{app:topic_pairs}); each panel shows the largest group of \textsc{vSLDA} topics matched to one SentLDA topic whose dominant fine category holds at least $50$\% of its posterior mass. (a)~20 Newsgroups/\textit{computer}: two topics matched to the SentLDA topic with the representative words \textit{cursor}, \textit{icon}, \textit{event}, \textit{ctrl}, \textit{expose}; (b)~NYT/\textit{arts}: three topics matched to the SentLDA topic with the representative words \textit{orchestra}, \textit{philharmonic}, \textit{carnegie}, \textit{symphony}, \textit{gilbert}. Each column gives the fitted concentration parameter $\kappa$, the normalized fine-category entropy $H$ of the topic's posterior sentence mass and the top 5 representative words; within a panel the topics are ordered by ascending $H$. The italic label above each topic is a descriptive label assigned by the authors from its representative words and is not an output of the model.}
\label{tab:family_topics}
\small
\setlength{\tabcolsep}{8pt}
\begin{tabular}[t]{@{}cc@{}}
\multicolumn{2}{c}{\textbf{(a)}} \\[2pt]
\toprule
\textit{X11/Unix windowing} & \textit{UI events/widgets} \\
($\kappa=267.3$) & ($\kappa=140.2$) \\
($H=0.17$) & ($H=0.91$) \\
\cmidrule(lr){1-2}
xterm & button \\
xdm & expose \\
xmu & ctrl \\
argv & clockwise \\
libxmu & roller \\
\bottomrule
\end{tabular}%
\hspace{2.5em}
\begin{tabular}[t]{@{}ccc@{}}
\multicolumn{3}{c}{\textbf{(b)}} \\[2pt]
\toprule
\textit{opera} & \textit{vocal repertoire} & \textit{music festivals} \\
($\kappa=289.7$) & ($\kappa=251.5$) & ($\kappa=243.9$) \\
($H=0.13$) & ($H=0.51$) & ($H=0.68$) \\
\cmidrule(lr){1-3}
opera & libretto & festival \\
orchestra & aria & coachella \\
verdi & penseroso & northside \\
philharmonic & coloratura & bonnaroo \\
britten & freshness & randalls \\
\bottomrule
\end{tabular}%
\end{table}

The fitted $\kappa_k$ tends to be larger for topics whose sentences concentrate on fewer fine categories. In both panels of Table~\ref{tab:family_topics}, the topic with the larger $\kappa_k$ has the smaller $H$. The relation is not specific to these runs. Over the $150$ vSLDA runs of the main experiments, the Spearman correlation between $\kappa_k$ and $H$ within a run is negative in every run, with a mean between $-0.69$ and $-0.60$ per dataset and topic count, and with the topic size partialled out it is negative in $146$ of the $150$ runs (Table~\ref{tab:kappa_entropy}). Because $H$ is computed from labels that the model does not use, $\kappa_k$, which is estimated from embedding directions alone, carries information about the breadth of a topic. There are exceptions. Topic 1 of the 20 Newsgroups/\textit{computer} run (\textit{appreciate}, \textit{thank}, \textit{help}) has the fourth largest $\kappa_k$ of the run and $H=0.99$. $\kappa_k$ thus measures directional tightness, which usually but not always coincides with a narrow subject.

\subsection{Conditions under Which Bag-of-Words LDA Remains Sufficient}
\label{subsec:when_lda}
NYT \textit{sports} is the only unit on which LDA is the best model. There, LDA achieves the best accuracy at $K=10$, $K=20$ and $K=30$, by less than the run-to-run standard deviation at $K=30$. It is again the best at $K=50$ and $K=100$ in the per-category sweep (\appref{app:topic_sweep_categories}) and stays ahead of vSLDA at every labeled fraction at $K=10$ and $K=20$.
The lead is bounded by the topic count. On NYT as a whole, LDA or SentLDA leads only up to $K=50$ and vSLDA from $K=100$ on, so LDA is sufficient at the smaller topic counts rather than for the corpus.
On NYT \textit{sports}, the lead also belongs to the admixture rather than to the vocabulary itself: SAM, which shares the vocabulary of LDA but models the document vector directionally, is behind LDA at $K=10$, $20$ and $30$, its admixture being formed over topic directions rather than over per-unit assignments (Section~\ref{subsec:spherical}).
What sets the unit apart is its size rather than the lexical separability of its fine categories. The fine categories of \textit{sports} are individual sports named by distinct vocabulary, but \appref{app:fine_similarity} does not establish that they are lexically separable in a way those of the other NYT units are not. The unit is, however, the largest in this study, a size that favors stable estimation of BoW co-occurrence statistics.
BoW-based LDA thus remains an adequate and inexpensive choice on a large unit at a small number of topics. On the smaller units, and on the whole corpus at the larger topic counts, vSLDA is the more accurate model.

\FloatBarrier

\section{Limitations}
\label{sec:limitations}
vSLDA takes the embedding space of a pretrained sentence encoder as given, keeps it fixed and assumes that cosine similarity in that space reflects semantic similarity. When it does not, the quality of the estimated topics may degrade. Sentence encoders are typically trained with objectives defined on cosine similarity~\citep{gao-2021-simcse} and evaluated by it~\citep{reimers-2019-sbert}, so this premise is expected to hold for general-domain English text such as the two corpora used here.
The premise is less certain for text in a specialized domain or in a low-resource language, for which an encoder of comparable quality may not be available. An adequate sentence embedding, in this sense, is a necessary condition for the model to work, and whether the encoder provides one has to be examined before vSLDA is applied to such text.

The sentences of a document are treated as exchangeable, and the observation unit depends on the sentence segmentation. Each sentence is assigned to exactly one topic, a strong assumption, as a sentence may contain more than one topic (Section~\ref{subsec:coherence_doclen}). Modeling sentence order, discourse relations or multi-topic sentences is a possible extension, at the cost of a more complex model.

The experiments are confined to two English corpora and three encoder-only English sentence encoders. The number of topics goes up to 300, and whether the advantage of vSLDA extends to the far larger numbers of topics used on web-scale corpora has not been examined. The evaluation rests on the classification of fine categories and on automated coherence and diversity measures. Human judgments of the topics would allow a more detailed assessment of their interpretability.
\FloatBarrier

\section{Conclusion}
\label{sec:conclusion}
This paper proposed and evaluated a spherical topic model whose observation unit is the sentence, a combination that, to our knowledge, had not been examined. vMF Sentence LDA (vSLDA) keeps the admixture structure of LDA, assigns one topic to each sentence and models each topic by a von Mises--Fisher (vMF) distribution over L2-normalized sentence embeddings, so that the number of parameters per topic and the per-iteration cost are linear in the embedding dimension.

Within the coarse categories of 20 Newsgroups and NYT, vSLDA attained the best classification accuracy in 27 of 30 conditions, and the advantage largely persisted under limited labels. On each corpus as a whole, vSLDA was ahead of every baseline at every topic count on 20 Newsgroups and from $K=100$ onward on NYT, with a margin that grew with the number of topics. Neither the sentence unit nor the vMF likelihood accounted for this advantage alone, and each had a larger effect in the presence of the other. The sentence-level Gaussian model reached the accuracy of vSLDA when refitted with an isotropic covariance on the normalized embeddings, the Gaussian form closest to the vMF likelihood.

Expected topic-word counts, obtained by weighting the word frequencies of each sentence by its topic posterior, allowed sentence-level and word-level models to be evaluated by the same coherence and diversity measures. vSLDA led on topic quality in most within-category conditions, whereas on coherence alone ETM was ahead in most conditions of 20 Newsgroups, the corpus with the shorter documents, and on both whole corpora beyond the smallest topic counts. LDA was the most accurate model on NYT \textit{sports}, the largest coarse category of the study, and LDA or SentLDA led on NYT as a whole up to $K=50$, so BoW-based LDA remains adequate on a large unit at a small number of topics.

Taken together, the results suggest that vSLDA is a practical topic model over sentence embeddings. It takes the embeddings of a pretrained sentence encoder and returns the per-document topic proportions that have made LDA useful across many fields. Its advantage appeared in the settings where topics are expected to share much of their vocabulary. Such topics arise, for example, when the documents of a collection broadly address one subject and its aspects are the topics of interest. vSLDA is expected to be useful in applications of this kind, and applying it to such collections is one direction for future work.
\FloatBarrier

\subsubsection*{Acknowledgments}
The authors thank the members of the Credit Policy \& Planning Division of MUFG Bank, Ltd. for their valuable advice and discussions.

The authors also thank Daichi Mochihashi (The Institute of Statistical Mathematics) for helpful discussions, and in particular for suggesting the experiments with a large number of topics.

This work was supported by a collaborative research from MUFG Bank, Ltd.; beyond the advice and discussions acknowledged above, the funder had no role in the collection and analysis of the data or in the decision to submit the article for publication.
The views and conclusions expressed in this paper are solely those of the authors and do not represent the official views of the individuals or organizations mentioned above.
\begingroup
\hbadness=10000
\bibliographystyle{tmlr}
\bibliography{references}
\endgroup

\appendix

\clearpage
\startcontents[appendix]
\printcontents[appendix]{l}{1}{\section*{Contents of the Appendix}}

\section{Coarse-to-Fine Category Mapping and Corpus Statistics}
\label{app:category_mapping}
Table~\ref{tab:category_mapping} lists the coarse categories used as experimental units and the fine categories that constitute them, and Table~\ref{tab:dataset_stats} summarizes the resulting units.
The grouping of fine categories into coarse categories follows \citet{cha-2024-sentencelda}.
For 20 Newsgroups, the fine categories are the original newsgroup labels. \textit{misc.forsale} is excluded because it does not belong to any of the six coarse categories.
For NYT, the fine categories are the fine-grained descriptor labels within each coarse section. The \textit{science} section is excluded because it contains only 90 documents, spread over its two fine categories (\textit{cosmos} and \textit{environment}), which is too few to support the within-category classification protocol after stratified splitting.
Documents are longer and more regular in NYT than in 20 Newsgroups, in sentences per document as well as in tokens per sentence.
\begin{table}[!t]
\centering
\footnotesize
\caption{Coarse categories (experimental units) of 20 Newsgroups (20news) and NYT, and their fine categories. For 20 Newsgroups, braces abbreviate a shared prefix (e.g., rec.\{autos, motorcycles\} denotes rec.autos and rec.motorcycles).}
\label{tab:category_mapping}
\begin{tabular}{@{}l l >{\raggedright\arraybackslash}p{0.56\linewidth}@{}}
\toprule
Dataset & Coarse category & Fine categories \\
\midrule
\multirow{6}{*}{20news}
 & computer & comp.\{graphics, os.ms-windows.misc, sys.ibm.pc.hardware, sys.mac.hardware, windows.x\} \\
 & ride & rec.\{autos, motorcycles\} \\
 & sports & rec.sport.\{baseball, hockey\} \\
 & science & sci.\{crypt, electronics, med, space\} \\
 & religion & alt.atheism, soc.religion.christian, talk.religion.misc \\
 & politics & talk.politics.\{guns, mideast, misc\} \\
\midrule
\multirow{4}{*}{NYT}
 & arts & dance, music, movies, television \\
 & business & economy, energy companies, international business, stocks and bonds \\
 & politics & abortion, federal budget, gay rights, gun control, immigration, law enforcement, military, surveillance, the affordable care act \\
 & sports & baseball, basketball, football, golf, hockey, soccer, tennis \\
\bottomrule
\end{tabular}
\end{table}
\begin{table}[!t]
\centering
\footnotesize
\setlength{\tabcolsep}{3pt}
\caption{Statistics of the experimental units (coarse categories) of 20 Newsgroups (20news) and NYT. ``\# Docs'' combines the training and test sets, ``Sent./doc'' is the mean number of sentences per document ($\pm$ standard deviation), and ``Tokens/sent'' is the mean number of tokens per sentence ($\pm$ standard deviation), counted over the evaluation vocabulary.}
\label{tab:dataset_stats}
\begin{tabular}{l l r r r r r}
\toprule
Dataset & Coarse cat. & \# Fine cat. & \# Docs & \# Sentences & Sent./doc & Tokens/sent \\
\midrule
\multirow{6}{*}{20news} & computer & 5 & 4{,}727 & 39{,}881 & \ms{8.44}{27.61} & \ms{15.88}{11.35} \\
 & ride & 2 & 1{,}875 & 12{,}039 & \ms{6.42}{9.78} & \ms{15.84}{10.87} \\
 & sports & 2 & 1{,}868 & 14{,}300 & \ms{7.66}{14.41} & \ms{15.85}{11.36} \\
 & science & 4 & 3{,}807 & 37{,}181 & \ms{9.77}{23.94} & \ms{17.44}{11.89} \\
 & religion & 3 & 2{,}344 & 27{,}963 & \ms{11.93}{18.46} & \ms{17.76}{12.46} \\
 & politics & 3 & 2{,}532 & 33{,}089 & \ms{13.07}{27.68} & \ms{17.91}{13.52} \\
\midrule
\multirow{4}{*}{NYT} & arts & 4 & 1{,}041 & 24{,}338 & \ms{23.38}{11.92} & \ms{26.24}{16.87} \\
 & business & 4 & 983 & 24{,}648 & \ms{25.07}{9.81} & \ms{24.40}{12.76} \\
 & politics & 9 & 989 & 25{,}441 & \ms{25.72}{9.60} & \ms{25.82}{14.59} \\
 & sports & 7 & 8{,}629 & 210{,}151 & \ms{24.35}{12.67} & \ms{23.22}{13.85} \\
\bottomrule
\end{tabular}
\end{table}
\FloatBarrier

\section{Distance between Fine Categories in the Lexical and Embedding Spaces}
\label{app:fine_similarity}
Section~\ref{subsec:when_lda} asks whether the fine categories of NYT \textit{sports} lie farther apart than those of the other units. This appendix measures that distance on the corpora themselves, independently of any fitted model. For each pair of fine categories of a unit, the distance in the lexical space is the Jensen--Shannon divergence (base 2) between the unigram distributions obtained by pooling the documents of each fine category under the preprocessing of Section~\ref{subsec:datasets}, and the distance in the embedding space is the cosine similarity between the centroid directions of their sentence embeddings, the statistic vSLDA estimates as $\bm\mu_k$ for a topic. Table~\ref{tab:fine_similarity} averages both over the pairs of each unit.
Both statistics depend on the amount of text behind each fine category. The Jensen--Shannon divergence between unsmoothed empirical distributions is biased upwards at small sample sizes, and the expected cosine similarity between two independently computed centroid directions falls as the number of sentences behind each decreases. The fine categories of NYT differ in size by a factor of $68$ in tokens, so both statistics are computed at equal fine-category size: for each corpus the size of its smallest fine category (in tokens for the lexical distance, in sentences for the embedding distance) serves as a common budget, that many tokens or sentences are drawn without replacement from every fine category, the statistic is computed on the draw, and the value reported in Table~\ref{tab:fine_similarity} is the average over $200$ such draws.

In both corpora and both spaces, the fine categories of one unit lie closer to one another than to those of other units. The mean divergence is $0.163$ within against $0.235$ across on 20 Newsgroups and $0.228$ against $0.310$ on NYT, and the mean cosine similarity $0.733$ against $0.354$ and $0.652$ against $0.278$. The classification targets defined in Section~\ref{subsec:datasets} are thus drawn from a narrower region of both spaces than the fine categories of the corpus as a whole.
Among the units, 20 Newsgroups \textit{religion} holds the closest fine categories and \textit{science} the most distant, in both spaces and under all three vocabulary treatments tested, namely the tokenization of Section~\ref{subsec:datasets}, a document-frequency floor of five and stopword removal. The four NYT units differ less from one another, spanning $0.018$ in mean divergence against $0.098$ on 20 Newsgroups, and their ranking is not stable across those treatments. \textit{sports} has the largest mean divergence under the first two and \textit{arts} under stopword removal.
\begin{table}[!t]
\centering
\footnotesize
\setlength{\tabcolsep}{6pt}
\caption{Distance between the fine categories inside each experimental unit. $|F|$ is the number of fine categories of the unit. JSD is the mean Jensen--Shannon divergence (base 2) between their unigram distributions and $\cos$ the mean cosine similarity between their centroid directions, both averaged over the pairs of the unit; a larger JSD and a smaller $\cos$ indicate fine categories that lie further apart. JSD and $\cos$ are averaged over the 200 equal-sized draws described in the text; their standard deviation over the draws is at most 0.001 and 0.007. The \textit{Different units} row gives the same pair averages over fine-category pairs drawn from different units. The sentence encoder is MiniLM.}
\label{tab:fine_similarity}
\begin{tabular}{@{}lccc@{\hspace{14pt}}lccc@{}}
\toprule
\multicolumn{4}{@{}l}{\textbf{20 Newsgroups}} & \multicolumn{4}{@{\hspace{14pt}}l}{\textbf{NYT}} \\
\cmidrule(r{7pt}){1-4} \cmidrule(l{7pt}){5-8}
Unit & $|F|$ & JSD & $\cos$ & Unit & $|F|$ & JSD & $\cos$ \\
\midrule
Computer & 5 & 0.164 & 0.781 & Arts & 4 & 0.224 & 0.760 \\
Ride & 2 & 0.129 & 0.838 & Business & 4 & 0.213 & 0.693 \\
Sports & 2 & 0.169 & 0.821 & Politics & 9 & 0.230 & 0.602 \\
Science & 4 & 0.203 & 0.512 & Sports & 7 & 0.231 & 0.695 \\
Religion & 3 & 0.105 & 0.935 &  &  &  &  \\
Politics & 3 & 0.148 & 0.751 &  &  &  &  \\
\addlinespace[2pt]
\textit{Different units} & -- & 0.235 & 0.354 & \textit{Different units} & -- & 0.310 & 0.278 \\
\bottomrule
\end{tabular}
\end{table}

\FloatBarrier

\section{Parameter Estimation and the SAEM Algorithm}
\label{app:estimation}
\paragraph{Point estimation of the topic parameters}
Section~\ref{subsec:inference} integrates out the document-topic proportions $\bm\theta$ but point-estimates the topic parameters $\bm\Delta$ and the Dirichlet parameter $\bm\alpha$. For $\bm\alpha$, this is the empirical Bayes (ML-II) treatment inherited from LDA~\citep{minka-2000-dirichlet,blei-2003-lda}, whereas $\bm\Delta$ parameterizes the likelihood rather than a prior, so estimating it is maximum likelihood rather than empirical Bayes.
For $\bm\Delta$, the role of the sample size is played by the number of sentences $n_k$ assigned to topic $k$. A topic carries only $M$ parameters, and the curvature of its complete-data log-likelihood $n_k\log C_M(\kappa_k)+\kappa_k\bm\mu_k^\top\bm r_k$ grows with $n_k$ in both: in $\bm\mu_k$ the likelihood is proportional to a vMF density with concentration $\kappa_k\|\bm r_k\|=\kappa_kn_k\bar r_k$, and the second derivative in $\kappa_k$ is $-n_kA_M'(\kappa_k)$.
Consistently with this, \citet{gopal-2014-clustering}, who implement fully Bayesian vMF mixtures, observe that the posterior of each concentration parameter was highly concentrated around its mode, which suggests that point estimates of $\bm\Delta$ lose little relative to fully Bayesian inference.

That $\bm\Delta$ is not integrated out in the same way as the document-topic proportions $\bm\theta$ is due to a technical reason. Given $\kappa_k$, the vMF family is conjugate for its mean direction, and $\bm\mu_k$ can be integrated out in closed form, as in the collapsed sampler of \citet{gopal-2014-clustering}. The likelihood of each sentence would then, however, depend on the current assignments, as in the collapsed Gaussian samplers discussed at the end of this appendix. The concentration $\kappa_k$ cannot be integrated out in the same way. Unlike the Dirichlet--multinomial case, the normalizing constant of the conjugate family of the vMF distribution in $(\bm\mu_k,\kappa_k)$ is an integral over $\kappa$ of terms involving the modified Bessel function of the first kind and has no closed form. The propriety of the priors and posteriors themselves also requires care~\citep{hornik-2013-conjugate}. A fully Bayesian treatment would therefore require either explicit sampling of each $\kappa_k$, as in \citet{gopal-2014-clustering}, who place a log-normal prior on $\kappa$ and sample it by MCMC with a log-normal proposal, or a one-dimensional numerical integration over each $\kappa_k$ at every evaluation of the likelihood.
\paragraph{Statement of the algorithm}
From the $B$ stored sweeps of an E-step, the Monte Carlo averaged count and resultant vector of topic $k$ are
\begin{equation}
  \tilde n_k = \frac{1}{B}\sum_{b=1}^{B}\sum_{d,i} \mathbb{I}[z_{di}^{(b)}=k],
  \qquad
  \tilde{\bm r}_k = \frac{1}{B}\sum_{b=1}^{B}\sum_{d,i} \mathbb{I}[z_{di}^{(b)}=k]\,\bm w_{di},
  \label{eq:suffstats}
\end{equation}
and Eq.~\eqref{eq:saem} averages them across outer iterations into the statistics $(n_k,\bm r_k)$ of the M-step, with $\bar r_k=\|\bm r_k\|/n_k$.
Algorithm~\ref{alg:mcem} states the procedure of Section~\ref{subsec:inference}; the handling of a topic left without sentences is described in \appref{app:hyperparams}.
\begin{algorithm}[!t]
\caption{Stochastic approximation EM for vSLDA.}
\label{alg:mcem}
\renewcommand{\algorithmicrequire}{\textbf{Input:}}
\renewcommand{\algorithmicensure}{\textbf{Output:}}
\begin{algorithmic}[1]
\Require normalized sentence embeddings $\{\bm w_{di}\}$; number of topics $K$; outer iterations $T$; burn-in length $T_0$; step-size exponent $a$; Gibbs sweeps per E-step $\zeta$; stored sweeps $B$; initial Dirichlet parameter $\alpha_0$; concentration $\kappa_0$ of a topic holding at most one sentence; concentration cap $\kappa_{\max}$; fixed-point budget $T_\alpha$
\Ensure $\bm\Delta=\{(\bm\mu_k,\kappa_k)\}_{k=1}^{K}$, $\bm\alpha$ and the final assignments $\bm z$
\State assign every sentence to a topic uniformly at random and set $\alpha_k \leftarrow \alpha_0$
\State let $\bm s_0$ be the statistics of these assignments, with components $(n_k,\bm r_k)$; set $\bm\mu_k \leftarrow \bm r_k/\|\bm r_k\|$ and $\kappa_k$ by Eq.~\eqref{eq:kappa-approx}
\For{$t = 1,\dots,T$}
  \For{$j = 1,\dots,\zeta$} \Comment{E-step; the chain continues from the assignments of iteration $t-1$}
    \ForAll{sentences $(d,i)$}
      \State sample $z_{di}$ from Eq.~\eqref{eq:gibbs}
    \EndFor
    \State \textbf{if} $j > \zeta - B$ \textbf{then} store the current $\bm z$ as $\bm z^{(j-\zeta+B)}$
  \EndFor
  \State compute $\tilde{\bm S}_t=\{(\tilde n_k,\tilde{\bm r}_k)\}_{k=1}^{K}$ from the stored sweeps by Eq.~\eqref{eq:suffstats}
  \State $\bm s_t \leftarrow \bm s_{t-1} + \gamma_t(\tilde{\bm S}_t - \bm s_{t-1})$, with components $(n_k,\bm r_k)$ \Comment{Eq.~\eqref{eq:saem}}
  \For{$k = 1,\dots,K$} \Comment{M-step for $\bm\Delta$}
    \State $\bm\mu_k \leftarrow \bm r_k/\|\bm r_k\|$
    \If{$n_k \le 1$}
      \State $\kappa_k \leftarrow \kappa_0$
    \Else
      \State $\kappa_k \leftarrow \min(\kappa_k^\star,\kappa_{\max})$, where $\kappa_k^\star$ solves $A_M(\kappa)=\bar r_k$ by Newton's method started from Eq.~\eqref{eq:kappa-approx}
    \EndIf
  \EndFor
  \State \textbf{if} $t \le T_0$ \textbf{then} update $\bm\alpha$ by Eq.~\eqref{eq:alpha}, to convergence or for at most $T_\alpha$ iterations \Comment{M-step for $\bm\alpha$}
\EndFor
\end{algorithmic}
\end{algorithm}

The fixed-point iteration of \citet{minka-2000-dirichlet} used in the M-step of Algorithm~\ref{alg:mcem} is
\begin{equation}
  \alpha_k \leftarrow \alpha_k\,
  \frac{\sum_{d=1}^{D}\bigl[\Psi(n_{dk}+\alpha_k)-\Psi(\alpha_k)\bigr]}
       {\sum_{d=1}^{D}\bigl[\Psi(N_d+\sum_{j}\alpha_j)-\Psi(\sum_{j}\alpha_j)\bigr]},
  \label{eq:alpha}
\end{equation}
where $\Psi$ is the digamma function and $n_{dk}$ are the document-topic counts of the final sweep.
\paragraph{Solution of the concentration equation}
The approximation of Eq.~\eqref{eq:kappa-approx} lies inside the interval $\bigl[\,\bar r_k (M-2),\; \bar r_k M\,\bigr]/(1-\bar r_k^2)$ that \citet{tanabe-2007-parameter} show to contain the exact solution of $A_M(\kappa_k)=\bar r_k$, so its relative departure from the exact maximum likelihood estimate is at most $2/(M-2)$.
The M-step refines it by Newton's method on $A_M(\kappa)-\bar r_k$, whose derivative is $1-A_M(\kappa)^2-(M-1)A_M(\kappa)/\kappa$~\citep{sra-2012-note}. At most two iterations reach machine precision. At $M=384$ the refined value differs from Eq.~\eqref{eq:kappa-approx} by at most $0.05$\% at any value of $\bar r_k$, and by less at larger $M$, so the refinement is numerically minor, but it brings the M-step to within machine precision of the exact maximizer that the convergence argument below assumes.
Throughout, $\log C_M(\kappa)$ is evaluated through a numerically stable log modified Bessel function, so that the likelihood remains finite for the large values of $\kappa$ that arise in high dimensions.
\paragraph{Finite-sample bias of the concentration estimate}
One consequence of taking sentences rather than words as the observation unit is that $\bar r_k$ is the mean resultant length of finitely many sentences, whose number is $n_k$ up to the averaging of Eq.~\eqref{eq:saem}, and carries a finite-sample upward bias, since the sampling variability of the individual directions inflates the length of their resultant, so that even a topic whose true concentration is zero receives a strictly positive estimate. The bias shrinks as $n_k$ grows and is more prominent for vSLDA than for word-level vMF models, because sentence counts are smaller than word counts by roughly an order of magnitude. Its size follows from the moments of the resultant. For $n$ independent draws from $\mathrm{vMF}(\bm\mu,\kappa)$ with $\bar r=\|\sum_i\bm w_i\|/n$ and $A=A_M(\kappa)$, $\mathbb{E}[\bar r^2]=A^2+(1-A^2)/n$ exactly, so $(n\bar r^2-1)/(n-1)$ is an unbiased estimate of $A^2$, and a second-order expansion of $\bar r$ about $A$ gives $\mathbb{E}[\bar r]-A\approx\mathbb{E}\|(\bm I-\bm\mu\bm\mu^\top)\bm w\|^2/(2An)=(M-1)/(2\kappa n)$. The chain rule through the estimator $g(\bar r)=\bar r(M-\bar r^2)/(1-\bar r^2)$ of Eq.~\eqref{eq:kappa-approx} then gives $(\mathbb{E}[g(\bar r)]-g(A))/g(A)\approx(1+A^2)/(2A^2n)$ to leading order in $1/n$ and $1/M$, about $2.6/n$ at $A=0.49$ ($M=384$, $\kappa=250$). This is the increment over the large-sample value $g(A)$, whose own distance from $\kappa$ is the approximation error of Eq.~\eqref{eq:kappa-approx} and does not depend on $n$. The exact root $A_M^{-1}(\bar r)$ that the M-step uses has the same leading term, and its large-sample value is $\kappa$ itself. It is a leading-order approximation to an expectation, not a bound, and says nothing about the sampling variability of the concentration estimate.
\paragraph{Convergence of the stochastic approximation}
The averaging of Eq.~\eqref{eq:saem} places Algorithm~\ref{alg:mcem} in the class of stochastic approximation EM algorithms with a Markov chain E-step. The convergence theorem of \citet{delyon-1999-saem} requires the missing data of each iteration to be drawn exactly from their conditional distribution, and \citet{kuhn-2004-saem} extend it to a draw from a Markov kernel that starts at the draw of the previous iteration. The argument of \citet{kuhn-2004-saem} carries over to Algorithm~\ref{alg:mcem}. Under the assumptions (i)--(iii) listed below, almost surely on the event that (iv) holds, the distance from the natural parameters $\bm\eta^{(t)}=(\kappa_k^{(t)}\bm\mu_k^{(t)})_k$ to the set of stationary points of the marginal log-likelihood $\ell=\log\sum_{\bm z}p(\mathcal{D},\bm z\mid\bm\alpha,\bm\Delta)$, taken as a function of $\bm\eta$, tends to zero. The statement is made for $\bm\eta$ rather than for $\bm\Delta$ because the direction $\bm\mu_k$ is not identified at $\kappa_k=0$.

The four assumptions are: (i) $\bm\alpha$ is fixed, with every $\alpha_k>0$; (ii) $\gamma_t=(t-T_0)^{-a}$ with $a\in(\tfrac12,1]$; (iii) the M-step returns the exact maximizer of the estimate of the expected complete-data log-likelihood, that is, the exact solution of Eq.~\eqref{eq:mstep}; and (iv) there are constants $c>1$ and $\kappa'<\kappa_{\max}$ such that along the iterations every component $n_k$ of $\bm s_t$ is at least $c$, no component $\bm r_k$ is the zero vector, every concentration estimate stays below $\kappa'$ and no topic is reseeded (\appref{app:hyperparams}). The statement concerns the iterations after the burn-in and is conditional on their starting point, namely $(\bm\Delta^{(T_0)},\bm s_{T_0},\bm z^{(T_0)})$ and the value of $\bm\alpha$ reached at iteration $T_0$, which is random but held fixed thereafter. Assumptions (i) and (ii) are settings of Algorithm~\ref{alg:mcem}, of which (ii) is the step-size condition of the theorem. Assumption (iii) idealizes the M-step, whose Newton iteration reaches the root of $A_M(\kappa)=\bar r_k$ only to machine precision. Assumption (iv) is a condition on the realized run, namely the compactness condition of the theorem written for this model. Under (iv) the direction $\bm r_k/\|\bm r_k\|$ is always defined, and the fallback $\kappa_0$ for $n_k\le1$, the cap $\kappa_{\max}$ and the reseeding step are never used.

Assumption (iv) concerns the whole sequence of iterations and cannot be verified on a finite run. We checked a finite-run proxy for it: in every run of the main experiments ($K\le 30$) and at $K=50$ in the topic-count sweep of Section~\ref{subsec:topic_sweep}, at each of the $T$ iterations, every topic held more than one sentence in the assignments of the final sweep of the E-step and no concentration estimate reached the cap. From $K=100$ upwards this failed in some runs, and the statement does not cover those runs. The check reads the counts of the final sweep, not the averaged counts $n_k$ of $\bm s_t$ that (iv) constrains, which were not recorded.

The verification of the statement follows the proof of Theorem~1 of \citet{kuhn-2004-saem} and has two parts, the model with its M-step and the Markov chain, and in each part one assumption of the theorem does not hold as stated. In the natural parameters $\bm\eta_k=\kappa_k\bm\mu_k$, with the parameter space taken as the open set $\{\|\bm\eta_k\|<\kappa_{\max}\}$, the complete-data log-likelihood is $\sum_k\{n_k(\bm z)\log C_M(\|\bm\eta_k\|)+\bm\eta_k^\top\bm r_k(\bm z)\}$ up to the term $\log p(\bm z\mid\bm\alpha)$, which is constant in $\bm\Delta$ and is absorbed into the reference measure. The complete-data model is a curved exponential family whose sufficient statistic $(n_k,\bm r_k)_k$ is bounded, since $n_k\le N_s$ and $\|\bm r_k\|\le n_k$. Because the assignments range over the finite set $\{1,\dots,K\}^{N_s}$, the conditional expectation of the statistic is a ratio of finite sums of smooth functions of $\bm\eta$ whose denominator is positive, and $\ell$ is the logarithm of that denominator, so both satisfy the regularity conditions of the theorem. For each topic, the direction that maximizes $\bm\eta_k^\top\bm r_k$ is $\bm r_k/\|\bm r_k\|$, and in $\kappa$ the function $n_k\log C_M(\kappa)+\kappa\|\bm r_k\|$ is strictly concave because $A_M'(\kappa)$ is the variance of $\bm\mu^\top\bm w$ under $\mathrm{vMF}(\bm\mu,\kappa)$ and is therefore positive, so the M-step map is $\hat{\bm\eta}_k=A_M^{-1}(\bar r_k)\,\bm r_k/\|\bm r_k\|=h(\bar r_k)\,\bm r_k/n_k$ with $h(q)=A_M^{-1}(q)/q$. Since $A_M$ is an odd analytic function with $A_M'(0)=1/M$, so is its inverse, and $h$ extends to an even analytic function with $h(0)=M$, that is, to an analytic function of $q^2=\|\bm r_k\|^2/n_k^2$. The map is therefore smooth in $(n_k,\bm r_k)$ for $n_k>0$, including at $\bm r_k=\bm 0$ where it equals $\bm 0$.

The theorem asks that the M-step map be defined on an open set containing the convex hull of the range of the statistic, which does not hold here: the hull contains points with $n_k=0$ or $\bar r_k=1$, where the map is not defined. In the proof of \citet{delyon-1999-saem}, to whose stochastic approximation theorem \citet{kuhn-2004-saem} reduce theirs, the hull serves only to keep every $\bm s_t$ inside that open set, and the compactness condition then asks that the closure of $\{\bm s_t\}$ be a compact subset of it. Under (iv) both hold directly, with the open set $\{n_k>0,\ \|\bm r_k\|<n_kA_M(\kappa_{\max})\}$ on which the M-step map is defined and smooth: the closure of $\{\bm s_t\}$ lies in its compact subset $\{c\le n_k\le N_s,\ \|\bm r_k\|\le n_kA_M(\kappa')\}$.

For the Markov chain, the theorem draws the missing data once per iteration from a kernel that leaves $p(\bm z\mid\mathcal{D},\bm\alpha,\bm\Delta)$ invariant, whereas Algorithm~\ref{alg:mcem} averages $B$ stored sweeps. We therefore take the state of iteration $t$ to be the block of the $B$ stored sweeps, so that $\tilde{\bm S}_t$ is a function of the state. The invariant distribution of this block kernel is the joint distribution of $B$ consecutive sweeps of the stationary chain rather than a conditional distribution of the model, but the proof uses the kernel only through its uniform ergodicity, its Lipschitz dependence on $\bm\eta$ and the mean of the statistic under its invariant distribution. Since each stored sweep has $p(\bm z\mid\mathcal{D},\bm\alpha,\bm\Delta)$ as its marginal, this mean is the conditional expectation of the statistic, as in the theorem. The theorem states its conditions on the kernel through a transition density, which on the finite space of assignments is the transition probability, the density with respect to the counting measure. Each conditional of Eq.~\eqref{eq:gibbs} is at least some $\varepsilon_0>0$ uniformly in $\bm\Delta$: each of its $K$ unnormalized weights $(n_{dk}^{-i}+\alpha_k)\exp(\ell_{dik})$ lies between two positive constants, because $\alpha_k>0$, the counts are at most $N_s$ and $|\ell_{dik}|\le|\log C_M(\kappa_k)|+\kappa_k$ is bounded on the parameter space. Every transition probability of one sweep is therefore at least $\varepsilon_0^{N_s}$, and of the block kernel of one E-step at least $\varepsilon_0^{N_sB}$, one factor for each stored sweep. This is a Doeblin condition, so the kernel is uniformly ergodic with constants that do not depend on $\bm\Delta$, and its transition probabilities, being finite sums of products of the smooth conditionals, are Lipschitz in $\bm\eta$ on compact sets. The one remaining condition, the convergence of the perturbation series of the Markov chain, is shown by \citet{kuhn-2004-saem} on the event that $\bm s_t$ never leaves a given compact set. The event that (iv) holds is a countable union of such events, which is why the statement is made on it.

The stronger result of \citet{delyon-1999-saem}, convergence to a local maximum rather than to a stationary point, rests on additional conditions on the stationary points and on the covariance of the statistic. We do not verify the former, and the latter does not hold for the statistic as written, whose covariance is singular because $\sum_kn_k$ and $\sum_k\bm r_k$ are constant. The argument is asymptotic and gives no rate of convergence. The constant $\varepsilon_0^{N_sB}$ is a worst-case lower bound and is not informative about how fast the chain mixes.

We know of no comparable statement for the Monte Carlo EM of \citet{li-2016-integrating}, which Algorithm~\ref{alg:mcem} with $T_0=T$ matches except for the exact solution of the concentration equation and the estimation of $\bm\alpha$: with $\gamma_t\equiv1$ and a fixed number $B$ of stored sweeps, the Monte Carlo error of $\tilde{\bm S}_t$ is not averaged out, and the convergence theory of \citet{fort-2003-mcem} for Monte Carlo EM would instead require $B$ to grow with $t$. Moreover, the closed-form approximation of Eq.~\eqref{eq:kappa-approx} that \citet{li-2016-integrating} use is not the exact maximizer that assumption (iii) requires.
\paragraph{Cost of the full-covariance samplers}
In the collapsed samplers of Gaussian LDA and GSLDA~\citep{das-2015-gaussian,cha-2024-sentencelda} the Gaussian parameters are integrated out, so the posterior predictive density of a topic depends on the sentences currently assigned to it, and the likelihood evaluations cannot be shared across sweeps even in principle. The rank-one Cholesky updates of these samplers avoid refactorizing the scale matrix of a topic after each reassignment, but not the triangular solve of $O(M^2)$ per sentence--topic pair (\appref{app:reference_code}), and the alias sampling of \citet{das-2015-gaussian} reduces the factor $K$ of a sweep but not this factor $M^2$. In vSLDA, by contrast, the log-likelihoods $\ell_{dik}$ depend neither on the current assignments nor on the sweep index within an outer iteration, so caching them in a table of size $O(N_sK)$ would reduce the E-step cost to $O(N_sKM+\zeta N_sK)$. Our implementation evaluates them anew in every sweep, and the costs of Section~\ref{subsec:complexity} and the times of \appref{app:wallclock} are those of this uncached evaluation. Diagonal and isotropic covariance structures, as in the refits of Table~\ref{tab:covariance}, admit density evaluation in $O(M)$ per sentence--topic pair, so the computational contrast drawn in Section~\ref{subsec:complexity} is specific to full covariance matrices.

\section{Implementation Details, Hyperparameters and Random Seeds}
\label{app:hyperparams}
Table~\ref{tab:hyperparams} collects the settings of Section~\ref{subsec:implementation} and of this appendix. The values are held fixed across all conditions unless stated otherwise.
\begin{table}[!t]
\centering
\scriptsize
\caption{Hyperparameters, preprocessing settings and random seeds of the experiments (Section~\ref{subsec:implementation}). $K$ is the number of topics, $M$ the embedding dimension and $i \in \{0,\dots,4\}$ the run index.}
\label{tab:hyperparams}
\begin{tabular}{@{}l >{\raggedright\arraybackslash}p{0.72\linewidth}@{}}
\toprule
Item & Setting \\
\midrule
\multicolumn{2}{@{}l}{\textbf{Common to all models}} \\
\addlinespace[1pt]
Runs and seeds & 5 independent training runs of the topic model per condition; run $i$ trains with seed $42+i$, set jointly for the Python, NumPy and PyTorch generators; classifier random state fixed across runs \\
Topic counts & $K \in \{10, 20, 30\}$ (main experiments); $K \in \{10, 20, 30, 50, 100, 200, 300\}$ on the whole corpus and up to $K=100$ within the coarse categories (Section~\ref{subsec:topic_sweep}; GLDA only within the coarse categories at $K\le30$, and GSLDA within the coarse categories only at $K\le30$) \\
Sentence encoders & \path{all-MiniLM-L6-v2} ($M=384$; main text); \texttt{all-mpnet-base-v2} and \texttt{bge-base-en-v1.5} ($M=768$; \appref{app:encoders}); outputs L2-normalized for vSLDA and GSLDA (except the refits of \appref{app:n_over_m}), unnormalized for ConTM; no instruction prefix for BGE \\
Word embeddings & \texttt{word2vec-google-news-300} ($M=300$) for GLDA, vLDA and ETM, whose vocabularies are the word types with a pretrained vector; L2-normalized for vLDA \\
Preprocessing & sentence segmentation with \texttt{pysbd}; rule-based sentence filter (sentences of fewer than 4 or more than 120 word tokens and fragments without sentence structure dropped; URLs, e-mail addresses and non-ASCII characters removed); tokenization, lowercasing, WordNet lemmatization, numerals mapped to \texttt{<NUM>}, single-character tokens dropped; no stopword or part-of-speech filtering; sentences truncated at the encoder's default maximum length (256/384/512 word pieces) \\
Data split & stratified (by fine category) 6:4 training/test split per corpus \\
Classifiers & scikit-learn logistic regression with the multinomial loss (main text) and RBF-kernel support vector machine (\appref{app:svm}); fixed random state, other settings default \\
Limited supervision & labeled fractions 5, 10, 20, 40, 60, 80, 100\%; below 100\%, 5 stratified subsets per run with distinct seeds \\
Representative words & fold-in sampler with frozen topic parameters for LDA, SentLDA, GLDA, GSLDA, vLDA and vSLDA (one chain, 20 burn-in and 20 retained sweeps, seed 0); each model's own posterior for ETM, ConTM and SAM; words ranked by the NPMI between word and topic under the expected counts \\
Coherence and diversity & $N=10$ representative words for $C_V$ and $C_{\mathrm{NPMI}}$, $N'=25$ for topic diversity; reference corpus of $10^{6}$ articles sampled from \texttt{enwiki-20260401}; sliding window of $s=110$ tokens ($C_V$) and $s=10$ tokens ($C_{\mathrm{NPMI}}$); smoothing constant $\epsilon=10^{-12}$ of the NPMI (Eq.~\eqref{eq:npmi}) \\
Structural diagnostics & diffuse: normalized entropy $H(P(d \mid k))/\ln D$ of the topic's distribution over the $D$ test documents above $0.95$; dead: topic ranked first in fewer than $1\%$ of the documents (\appref{app:entropy_thresholds} for other cut-offs) \\
\midrule
\multicolumn{2}{@{}l}{\textbf{Per model}} \\
\addlinespace[1pt]
vSLDA & one vMF component per topic; $T=10$ outer iterations of the SAEM of Section~\ref{subsec:inference}, with burn-in $T_0=5$ and step sizes $\gamma_t=(t-T_0)^{-1}$ afterwards; $\zeta=20$ Gibbs sweeps per E-step; $B=8$ stored sweeps (the last $B$ of the $\zeta$); concentrations $\kappa_k$ by Newton's method at every M-step, started from Eq.~\eqref{eq:kappa-approx} at the current statistics and capped at $\kappa_{\max}=10^{4}$ ($\kappa_0=10.0$ for a topic holding at most one sentence); Dirichlet parameters $\alpha_k$ initialized to $50/K$ and updated during the burn-in by fixed-point iteration (at most 100 iterations, tolerance $10^{-5}$), fixed afterwards; document-topic distributions of both splits by fold-in with the fitted parameters frozen, one chain, 20 burn-in and 20 retained sweeps, seed 0 (Eq.~\eqref{eq:foldin-theta}) \\
LDA & \texttt{gensim}; 20 passes; at most 50 inference iterations per document; symmetric Dirichlet priors $1/K$ (the library defaults); topic weights below $0.01$ set to zero \\
SentLDA & collapsed Gibbs sampling, 20 sweeps; symmetric Dirichlet priors $1/K$ on the document-topic and topic-word distributions; training documents by the counts of the final sweep, held-out documents by 50 fold-in sweeps \\
SAM & L2-normalized term-frequency vector over the full training vocabulary; variational EM, at most 200 iterations or relative change of the non-constant part of the bound below $10^{-5}$; topic directions initialized at random documents; Dirichlet parameter 1, fixed; concentrations set through their mean resultant lengths, 0.1 for the document likelihood, 0.99 for the prior on the topic directions and 0.01 for the prior on the corpus mean, held fixed; prior mean of the corpus direction fixed to the empirical corpus direction; 100 gradient steps on the variational Dirichlet parameter for held-out documents \\
GLDA & 20 Gibbs sweeps; normal--inverse-Wishart prior with scale matrix $\lambda\bm I$, $\lambda=0.1$, prior strength $\tau_0=0.1$ of the mean, $\nu_0=M$ and $\bm\mu_0$ the mean of the pretrained word vectors; Dirichlet parameter $1/K$, fixed; training documents by the counts of the final sweep, held-out documents by 20 fold-in sweeps \\
vLDA & Monte Carlo EM of Section~\ref{subsec:inference} without averaging ($T_0=T$) on L2-normalized word embeddings; one vMF component per topic; 20 outer iterations; one Gibbs sweep per E-step, $B=1$; Dirichlet parameter fixed at $1/K$; concentrations by Eq.~\eqref{eq:kappa-approx}, capped at $\kappa_{\max}=10^{4}$; document-topic distributions of both splits by the fold-in sampler over the word tokens, as for vSLDA (Eq.~\eqref{eq:foldin-theta}) \\
ETM & 100 epochs; batch size 128; inference network with two hidden layers of 800 ReLU units; Adam, learning rate $2\times10^{-3}$, weight decay $1.2\times10^{-6}$; pretrained word embeddings kept fixed; document-topic distributions of both splits from the mean of the inference network, without sampling \\
ConTM & CombinedTM of the reference implementation with its default architecture; 50 epochs; batch size 64; document-topic distributions averaged over 20 samples \\
GSLDA & 10 Gibbs sweeps; normal--inverse-Wishart prior with scale matrix $\lambda\bm I$, $\lambda=0.1$ (\appref{app:n_over_m}), $\tau_0=0.1$, $\nu_0=M$ and $\bm\mu_0$ the mean of the document mean embeddings; Dirichlet parameter $0.1$, fixed; training documents by the counts of the final sweep, held-out documents by 20 fold-in sweeps \\
\bottomrule
\end{tabular}
\end{table}
\paragraph{Sentence encoders and embeddings}
MiniLM (\path{sentence-transformers/all-MiniLM-L6-v2}) is a Sentence-BERT-type bi-encoder whose six-layer backbone keeps every second layer of a 12-layer MiniLM model distilled from a BERT-base-sized teacher~\citep{wang-2020-minilm} and which is fine-tuned with a contrastive objective on over one billion sentence pairs. It is a standard, openly licensed default for sentence similarity.
In the sentence-level models, each sentence is converted to a sentence embedding and then L2-normalized, except in the refits on unnormalized input of \appref{app:n_over_m}, and GSLDA and vSLDA encode the same sentences with the same encoder within each condition. ConTM applies the same encoder to the concatenated sentences of the document, truncated at the encoder's maximum sequence length and without L2 normalization, so the three models depend on the sentence encoder while the others do not.
The word-embedding-based models (GLDA, vLDA and ETM) use the pretrained \texttt{word2vec-google-news-300} word embeddings ($M=300$) in all conditions, so their results are identical across the sentence-encoder conditions. Their vocabularies are restricted to the word types that have a pretrained vector, so \texttt{<NUM>} and the other out-of-vocabulary types are not modeled. vLDA L2-normalizes the word vectors; GLDA and ETM use them as released.
\paragraph{vSLDA}
The concentration parameters are initialized from the initial random assignments by Eq.~\eqref{eq:kappa-approx} and estimated at every M-step by Newton's method started from Eq.~\eqref{eq:kappa-approx} evaluated at the current averaged statistics, subject to the cap $\kappa_{\max}=10^{4}$ of Algorithm~\ref{alg:mcem}, which no run of the main experiments reached (\appref{app:estimation}). The value $\kappa_0=10.0$ is assigned to a topic holding at most one sentence, which did not occur in the main experiments. A topic to which the final sweep of an E-step assigns no sentence is reseeded before the M-step: one sentence drawn uniformly at random from the topic holding the most sentences is reassigned to it, and the statistics $\tilde{\bm S}_t$ of that iteration are computed from the assignments of the final sweep alone rather than from the $B$ stored sweeps. Algorithm~\ref{alg:mcem} omits this step, which did not occur in the main experiments either. The Dirichlet parameter is initialized symmetrically as $\alpha_0=50/K$ following \citet{griffiths-2004-finding} and updated during the burn-in by Eq.~\eqref{eq:alpha} with at most $T_\alpha=100$ fixed-point iterations per M-step, stopped when the largest relative change falls below $10^{-5}$; each $\alpha_k$ is kept at or above $10^{-3}$, a floor reached only at $K\ge100$ of the topic-count sweep.
The document-topic distribution of a training or held-out document is
\begin{equation}
\tilde\theta_{dk}=\frac{\mathbb{E}[n_{dk}\mid\bm w_d,\bm\Delta,\bm\alpha]}{N_d},
\label{eq:foldin-theta}
\end{equation}
the posterior expected number of sentences of $d$ under topic $k$, the conditional probabilities of Eq.~\eqref{eq:gibbs} averaged over the retained sweeps and summed over the sentences of the document, divided by $N_d$, without the Dirichlet pseudo-counts $\alpha_k$. The pseudo-counts are omitted because the fitted $\sum_k\alpha_k$ is of the same order as the number of sentences of a typical document (Table~\ref{tab:dataset_stats}), so that the posterior mean $(\mathbb{E}[n_{dk}]+\alpha_k)/(N_d+\sum_j\alpha_j)$ would draw a large share of every $\tilde{\bm\theta}_d$ toward the prior; the raw proportion also matches the count-based document-topic features of GSLDA below. The fold-in sampler of Section~\ref{subsec:repwords} is run on the document with the fitted $\bm\Delta$ and $\bm\alpha$ frozen and the settings of the representative-word extraction below (one chain, 20 burn-in sweeps followed by 20 retained sweeps, seed 0). The same estimator serves the training documents of the classifier and the test documents, so that no document-topic distribution of vSLDA is read from the training chain itself.
\paragraph{Baselines}
The baseline models largely follow the standard settings of their original papers. The exact values, held fixed across all conditions, are as follows.
LDA uses the \texttt{gensim} implementation with 20 passes over the corpus, at most 50 inference iterations per document and the library's default symmetric Dirichlet priors ($1/K$). The document-topic distribution of a training or held-out document is the mean of its variational posterior, with topic weights below $0.01$ set to zero, the library's default.
SentLDA is trained by collapsed Gibbs sampling for 20 sweeps with symmetric Dirichlet priors of $1/K$ on both the document-topic and the topic-word distributions. The document-topic proportions of a training document are the topic counts of the final sweep, and those of a held-out document the counts after 50 fold-in sweeps with the topic-word counts held fixed.
SAM observes, for each document, its term-frequency vector over the full training vocabulary of the unit (the vocabulary of LDA, without pruning), L2-normalized to the unit sphere, and is fit by the variational EM of \citet{reisinger-2010-spherical} in our own implementation for at most 200 iterations or until the relative change of the non-constant part of the bound falls below $10^{-5}$; the topic directions are initialized at randomly chosen document directions, the Dirichlet parameter is held fixed at $\alpha_k=1$, and the prior mean of the corpus direction is fixed to the empirical corpus direction.
Because a vMF concentration is meaningful only relative to the dimension, here the vocabulary size, the three concentrations of SAM are set through their mean resultant lengths $A_V$: $A_V(\kappa)=0.1$ for the document likelihood, $A_V(\xi)=0.99$ for the prior on the topic directions and $A_V(\kappa_0)=0.01$ for the prior on the corpus mean, and they are held fixed during training.
The document-topic proportions of a document are the mean of its variational Dirichlet posterior, obtained for a held-out document by 100 gradient steps on its variational Dirichlet parameter with the topic directions held fixed.

GLDA runs 20 Gibbs sweeps with a conjugate normal--inverse-Wishart prior whose scale matrix is $\lambda\bm{I}$ with $\lambda=0.1$, as for GSLDA below, with prior strength $\tau_0=0.1$ of the mean, $\nu_0=M$ degrees of freedom and prior mean $\bm\mu_0$ the mean of all pretrained word vectors; its Dirichlet parameter is fixed at $1/K$. The document-topic proportions of a training document are the topic counts of the final sweep, and those of a held-out document the counts after 20 fold-in sweeps with the topic parameters held fixed.
vLDA is trained by the Monte Carlo EM of \citet{li-2016-integrating} applied to word embeddings, the procedure of Section~\ref{subsec:inference} without the averaging step ($T_0=T$), with a single vMF component per topic, 20 outer iterations, one Gibbs sweep per E-step ($B=1$), the Dirichlet parameter fixed at the symmetric $\alpha_k=1/K$ and the approximation of Eq.~\eqref{eq:kappa-approx} for $\kappa_k$ under the same cap $\kappa_{\max}=10^{4}$, which topics formed by a single frequent word type such as \emph{the} or \emph{be} reach in nearly every run of the main experiments; the document-topic distribution of a training or held-out document is Eq.~\eqref{eq:foldin-theta}, obtained as for vSLDA by the fold-in sampler of Section~\ref{subsec:repwords} run over the word tokens of the document with the fitted topic parameters frozen.
In its original form, the model of \citet{li-2016-integrating} represents each topic as a \emph{mixture} of vMF components.
In this paper we run it with a single component per topic.
This is because mixture components are an orthogonal extension that can be introduced into any of the compared observation models, including vSLDA, and are not the focus of this work.
vLDA and vSLDA then share the vMF observation model and differ only in the observation unit (word embeddings vs.\ sentence embeddings) and in the estimation settings listed above.

ETM is trained for 100 epochs with batch size 128, an inference network with two hidden layers of 800 ReLU units, and the Adam optimizer with learning rate $2\times10^{-3}$ and weight decay $1.2\times10^{-6}$. The pretrained word embeddings are kept fixed during training, i.e., we use the pre-fitted word embedding variant of \citet{dieng-2020-topic}.
ConTM uses the CombinedTM of the reference implementation with its default network architecture, trained for 50 epochs with batch size at most 64, and document-topic distributions are averaged over 20 samples.
GSLDA is trained with 10 Gibbs sweeps over the training corpus. Following \citet{cha-2024-sentencelda}, the scale matrix of its normal--inverse-Wishart prior is $\lambda\bm{I}$ with $\lambda=0.1$ (the prior scale swept in \appref{app:n_over_m}), with prior strength $\tau_0=0.1$ of the mean, $\nu_0=M$ degrees of freedom and prior mean $\bm\mu_0$ the mean of the mean sentence embeddings of the training documents; its Dirichlet parameter is fixed at $0.1$. The document-topic proportions of a training document are the topic counts left by the final sweep, and those of a held-out document are obtained by 20 fold-in sweeps with the topic parameters held fixed. Raising the number of training sweeps from 10 to the 20 used by \citet{cha-2024-sentencelda} changed the accuracy of a run by at most $0.3$ points on the two units we tested, which is inside the run-to-run spread, so we keep the 10-sweep setting throughout. Our implementation evaluates the full quadratic form of the Student-$t$ predictive; the released implementation does not (Section~\ref{subsec:why_gslda}, \appref{app:reference_code}). It ports to sentence embeddings the collapsed Gibbs sampler of the reference Gaussian LDA package,\footnote{\url{https://github.com/markgw/gaussianlda}.} a Python port of the Java implementation of \citet{das-2015-gaussian}, with the per-sentence kernels compiled with Numba as in vSLDA (\appref{app:wallclock}).
\paragraph{Preprocessing}
All methods are trained and evaluated under common preprocessing.
English text is segmented into sentences with \texttt{pysbd}~\citep{sadvilkar-2020-pysbd} and passed through a rule-based filter, which drops sentences of fewer than 4 or more than 120 word tokens and fragments without sentence structure (punctuation-heavy, upper-case noise, parenthetical-only), removes URLs, e-mail addresses and non-ASCII characters from the rest, and excludes the documents left without a sentence. The kept sentences are tokenized into alphabetic and numeric tokens, lowercased and lemmatized with WordNet, numeric tokens are normalized to \texttt{<NUM>}, and single-character tokens are dropped.
Stopword removal and part-of-speech-based token filtering are not applied.
Sentences longer than an encoder's default maximum sequence length (256, 384 and 512 word-piece tokens for MiniLM, MPNet and BGE, respectively) are truncated by the Sentence-Transformers tokenizer. Because the filter removes sentences of more than 120 word tokens, this is rarely triggered for a sentence; the concatenated documents given to ConTM, by contrast, regularly exceed that length and are truncated there.
\paragraph{Representative words and coherence}
The evaluation vocabulary of Section~\ref{subsec:repwords} consists of the word types of the preprocessed corpus of the unit that occur in at least $3$ documents and in at most $70\%$ of them, excluding \texttt{<NUM>}.
The three properties of the expected counts stated in Section~\ref{subsec:repwords} are, in full: by linearity of expectation, $q_{kw}$ is the posterior expected number of occurrences of $w$ generated under topic $k$ when every token inherits the topic of its unit,
\begin{equation}
  q_{kw} = \mathbb{E}\Bigl[\,\textstyle\sum_{d,i}\mathbb{I}[z_{di}=k]\,n(w,u_{di}) \,\Big|\, \mathcal{D}\Bigr];
  \label{eq:induced-expectation}
\end{equation}
because $\sum_k r_{dik}=1$ for every unit, the counts conserve token mass,
\begin{equation}
  \sum_{k=1}^{K} q_{kw} = \sum_{d,i}\Bigl(\sum_{k=1}^{K} r_{dik}\Bigr) n(w,u_{di}) = c_w ,
  \label{eq:induced-mass}
\end{equation}
so observed occurrences are allocated across topics without being created or deleted; and when every unit is a single token, Eq.~\eqref{eq:induced-counts} reduces to
\begin{equation}
  q_{kw} = \sum_{d,i} r_{dik}\,\mathbb{I}[x_{di}=w],
  \label{eq:induced-lda}
\end{equation}
the topic-word count of word-level LDA.
The expected count of Eq.~\eqref{eq:induced-lda} is the hard Gibbs count of word-level LDA when $r_{dik}=\mathbb{I}[z_{di}=k]$ and the variational soft count when $r_{dik}$ is a variational marginal. A sentence contributes to the expected counts of Eq.~\eqref{eq:induced-counts} in proportion to its length, and weighting every sentence equally would break Eq.~\eqref{eq:induced-mass} and up-weight short sentences, whose embeddings are computed from little context.

The fold-in sampler used for the representative-word extraction of Section~\ref{subsec:repwords} uses the same settings for every model to which it applies: a single chain, 20 burn-in sweeps followed by 20 retained sweeps without thinning, whose conditional probabilities are averaged to give $r_{dik}$. The random seed of the sampler is fixed to 0 independently of training.
In this paper we set the number of representative words to $N=10$ for $C_V$ and $N'=25$ for topic diversity.
As the reference corpus we use a uniform random sample, without replacement and with random seed 42, of 1{,}000{,}000 articles of the 20260401 dump of English Wikipedia (\texttt{enwiki-20260401}), and the co-occurrence statistics are computed on this reference corpus.
$C_V$ is computed with our own Palmetto-compatible implementation that follows \citet{roder-2015-coherence} (boolean sliding windows, NPMI-based word vectors with one-set segmentation, cosine similarity and arithmetic-mean aggregation); the \texttt{gensim} library is used only to build the vocabulary.
The sliding window has the standard $C_V$ width $s=110$ tokens, and the smoothing constant of Eq.~\eqref{eq:npmi} is $\epsilon=10^{-12}$.
The $C_{\mathrm{NPMI}}$ scores of \appref{app:coherence_npmi} use the same representative words, reference corpus and $\epsilon$, with the standard $C_{\mathrm{NPMI}}$ window width $s=10$ tokens.
The reference corpus is processed with the same tokenization pipeline as the modeling corpus (Unicode normalization, lowercasing, WordNet lemmatization and normalization of numeric tokens), so that the topic words and the reference co-occurrence counts share an identical lexical normalization.

For ETM, ConTM and SAM, the fold-in sampler above does not apply, since none of them assigns a latent topic to an observed unit: ETM has a native topic-word distribution, the ProdLDA-type decoder of ConTM draws each token from a normalized product of the topic factors rather than from a mixture over topics, and SAM generates the whole document vector as a single vMF observation whose mean direction combines the topic directions (Section~\ref{subsec:spherical}).
To rank their words by the same criterion, we take the unit to be a word token and obtain $r_{dik} \propto \theta_{dk}\,\hat\psi_{kw}$, normalized over $k$, from each model's own variational posterior instead: $\hat{\bm\psi}_k$ is the frozen fitted topic-word distribution of the model (for ConTM, the topic-wise softmax of its decoder weights; for SAM, the positive part of its unit topic direction over the vocabulary, renormalized to sum to one, which discards the negative word weights that SAM can assign) and $\bm\theta_d$ its document-topic distribution (for ETM, $r_{dik}$ is averaged over 100 draws of $\bm\theta_d$ from the logistic-normal posterior of its inference network with seed 0; for ConTM, the 20-sample average above; for SAM, the mean of its variational Dirichlet posterior), so that Eq.~\eqref{eq:induced-counts} and the NPMI ranking apply unchanged.
This quantity is the model's own token-topic posterior for ETM, whose likelihood is the mixture $\sum_k \theta_{dk}\hat\psi_{kw}$, but only a post-hoc mixture reading of the fitted parameters for ConTM and SAM.
\paragraph{Classifiers and random seeds}
For the classification evaluation we use the scikit-learn implementations of logistic regression with the multinomial loss and, in \appref{app:svm}, of the support vector machine, with a random state fixed to the same value in every run and all other settings at their defaults.
A run is one training of a topic model followed by the training and evaluation of the classifier on its document-topic distributions. We report means and standard deviations over 5 independent runs. The runs differ only in the seed of the topic-model training, $42+i$ for run $i=0,\dots,4$, set jointly for the Python, NumPy and PyTorch generators at the start of training, so the classifier receives a different set of features in every run while its own random state does not change.
\paragraph{Practical notes on $K$ and the encoder}
For the choice of the number of topics, few topics of vSLDA are dead even at $K=30$ and its whole-corpus accuracy was still rising at $K=300$ (Section~\ref{subsec:topic_sweep}, \appref{app:entropy_thresholds}), so a large $K$ costs little when the representations feed a classifier, whereas its coherence falls with $K$, so a smaller $K$ is preferable when the topics are to be read.
There is no covariance prior whose scale must be tuned, and the per-iteration cost is linear in the embedding dimension, so a higher-dimensional encoder remains affordable (\appref{app:wallclock}). Substituting the sentence encoder left the classification comparison unchanged but shifted the coherence comparison on 20 Newsgroups (\appref{app:encoders}), so an encoder change should be validated on the target corpus.
The wall-clock training times of vSLDA and GSLDA under these settings, and the convergence of their samplers, are reported in \appref{app:wallclock}.

\section{Wall-Clock Training Time and Convergence of vSLDA and GSLDA}
\label{app:wallclock}
This appendix reports the measured training time of vSLDA and GSLDA, the two sentence-level models whose per-sentence likelihood evaluation differs in its dependence on the embedding dimension $M$ (Section~\ref{subsec:complexity}), together with the convergence of their samplers against wall-clock time.
The measurements use the \textit{computer} unit of 20 Newsgroups (2{,}836 training documents, 24{,}508 sentences) at $K=20$ with the settings of Section~\ref{subsec:implementation} and \appref{app:hyperparams}, and 3 runs per condition that differ only in the random seed.
They were run on the CPU of a single workstation (Intel Core i9-13900K, 24 cores). Its GPU (NVIDIA RTX A6000) was used to compute the sentence embeddings.

Both models run in our own implementations, which follow the same structure: the loop over documents and sweeps is written in Python, the per-sentence sampling is compiled with Numba, and the linear algebra uses NumPy. The likelihood evaluation is compiled with Numba for GSLDA, as a triangular solve per sentence, and is one NumPy matrix product per document for vSLDA. Neither implementation is parallelized beyond the thread pools of these libraries, which are pinned to one thread here. The GSLDA implementation ports the collapsed sampler of the reference Gaussian LDA package (\appref{app:hyperparams}) to sentences and keeps its Cholesky-based rank-one updates of the topic covariances. The $O(M^2)$ triangular solve of its predictive density is evaluated inside the compiled kernel, and the density is recomputed in every sweep rather than cached (\appref{app:estimation}). Neither implementation was tuned for this comparison, and the ratios below are those of two implementations built to the same standard rather than of the algorithms. The direction of the gap and its growth with $M$ follow from the per-pair costs of Section~\ref{subsec:complexity}, whereas the exact factor depends on such implementation choices.
With MiniLM ($M=384$) the runs continue past the training length of the main experiments, to 30 outer iterations for vSLDA and 20 Gibbs sweeps for GSLDA, so that the traces show whether anything changes after that length. With MPNet ($M=768$), they use the training length of the main experiments. Table~\ref{tab:wallclock} reports, for every run, the time up to that training length.
Every run is single-threaded, with the BLAS, OpenMP and Numba thread pools pinned to one thread, and is bound to one physical core, with at most four runs executing at a time. Pinning removes the parallelism of the linear-algebra libraries, so that the comparison reflects the arithmetic of the two samplers as implemented here rather than how well each library parallelizes it.
The per-sweep time of Table~\ref{tab:wallclock} includes the per-iteration work each model does around its sweeps, for vSLDA the M-step, the $\bm\alpha$ update and the likelihood evaluation, spread over the $\zeta=20$ sweeps of an outer iteration, and for GSLDA the likelihood evaluation that follows each sweep.
\begin{table}[!t]
\centering
\footnotesize
\setlength{\tabcolsep}{5pt}
\caption{Wall-clock training time of vSLDA and GSLDA on the \textit{computer} unit of 20 Newsgroups ($K=20$; mean $\pm$ standard deviation over 3 runs). \emph{Per sweep} is the time of one Gibbs sweep over the training sentences, averaged over all recorded sweeps and including the per-iteration work around the sweeps. \emph{Training} is the time of the training length of the main experiments, 10 outer iterations (200 sweeps) for vSLDA and 10 Gibbs sweeps for GSLDA. The setting and what each column includes are described in the text.}
\label{tab:wallclock}
\begin{tabular}{@{}llrrr@{}}
\toprule
Model & Encoder ($M$) & Per sweep (s) & Sweeps & Training (s) \\
\midrule
\textsc{vSLDA} & MiniLM (384) & 0.038 $\pm$ 0.000 & 200 & 7.74 $\pm$ 0.03 \\
\textsc{vSLDA} & MPNet (768) & 0.055 $\pm$ 0.000 & 200 & 11.0 $\pm$ 0.1 \\
\addlinespace[2pt]
GSLDA & MiniLM (384) & 60.2 $\pm$ 11.2 & 10 & 622.7 $\pm$ 82.9 \\
GSLDA & MPNet (768) & 288.0 $\pm$ 0.8 & 10 & 2,880 $\pm$ 8 \\
\bottomrule
\end{tabular}
\end{table}

Training GSLDA takes 80 times as long as training vSLDA under MiniLM and 262 times as long under MPNet (Table~\ref{tab:wallclock}).
The gap widens with the embedding dimension in the direction the per-iteration costs of Section~\ref{subsec:complexity} imply. Doubling $M$ from 384 to 768 multiplies the time of one Gibbs sweep by 1.4 for vSLDA and by 4.8 for GSLDA.
The factor for GSLDA is somewhat above the 4 that a cost quadratic in $M$ implies, whereas the factor for vSLDA stays below the 2 that a cost linear in $M$ implies, because at this corpus size the parts of a sweep that do not depend on $M$, above all sampling from the $K$-dimensional conditional distributions, are a substantial share of its cost.
\begin{figure}[!t]
\centering
\includegraphics{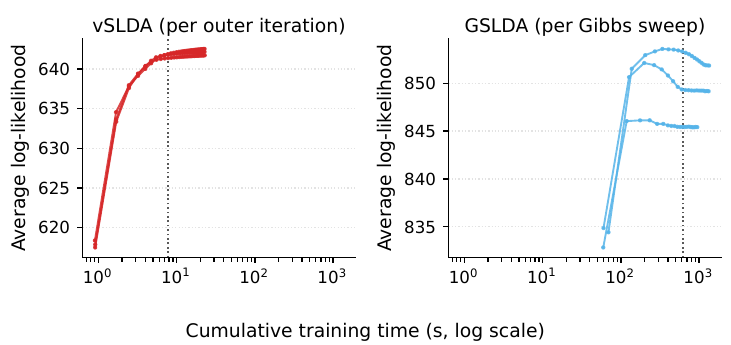}
\caption{\textbf{Sampler convergence against wall-clock time (20 Newsgroups/\textit{computer}, $K=20$, MiniLM).} Each panel shows the average log-likelihood of the training sentences against the cumulative training time in seconds, on a logarithmic scale, for vSLDA (left) and GSLDA (right), with one line per run (3 runs). The vMF likelihood is a density on the unit hypersphere and the GSLDA likelihood a Student-$t$ posterior predictive density on $\mathbb{R}^M$, so the two are not comparable in value. Each panel therefore carries its own vertical axis and only the horizontal axis is shared. The dotted rule marks the training length of the main experiments, 10 outer iterations for vSLDA and 10 Gibbs sweeps for GSLDA. Both samplers reach their plateau within that length.}
\label{fig:convergence}
\end{figure}

Figure~\ref{fig:convergence} places the two samplers on a common time axis, on which a whole vSLDA run precedes the completion of the first GSLDA sweep.
Both have reached their plateau within the training length of the main experiments and neither moves afterwards. The vSLDA traces are flat from about the fifth outer iteration and unchanged out to 30 iterations, and the GSLDA traces are flat from about the third sweep and unchanged out to 20 sweeps.
For GSLDA this is evidence independent of the accuracy comparison of \appref{app:hyperparams} that the 10-sweep setting is not what separates our GSLDA results from those of \citet{cha-2024-sentencelda}.
The GSLDA traces also rise to an early maximum and then settle slightly below it, the likelihood signature of the concentration of sentences on a few topics discussed in Section~\ref{subsec:why_gslda}. Because that trace is not monotone, it is read here only for the time at which it stops changing.
\FloatBarrier

\section{Per-Condition Results}
\label{app:per_condition}
Tables~\ref{tab:cls_logreg_minilm}, \ref{tab:topic_coherence_c_v}, \ref{tab:topic_diversity} and \ref{tab:topic_quality} list, for every condition of the main experiments (ten experimental units at $K\in\{10,20,30\}$, MiniLM, 5 runs), the classification accuracy, the $C_V$ coherence, the topic diversity and their product that Table~\ref{tab:results_summary} summarizes by dataset and number of topics. The column groups of the tables follow Table~\ref{tab:model_map}: the upper header row gives the observation unit of each model (word, sentence or document), and the lower row groups the models into the vocabulary-based LDA, SentLDA and SAM, whose observations are counts or term frequencies over the vocabulary, and the embedding-based models, which observe word or sentence embeddings (Section~\ref{subsec:baselines}). SAM stands outside the factorial design of Section~\ref{subsec:baselines}, its observation being the term-frequency vector of a document. With the vMF likelihood placed on the document instead of the sentence, its mean accuracy is $22.4$ and $17.0$ points below that of vSLDA on 20 Newsgroups and NYT, and it is ahead of LDA, which shares its vocabulary, in only 9 of the 30 conditions, so a directional likelihood over document vectors does not recover the advantage that vSLDA obtains from sentence embeddings. GLDA reaches a topic diversity of exactly 1.000 in all conditions because its expected counts assign each word type to a single topic, so its representative-word lists are disjoint by construction. On coherence, the document-level SAM is above LDA in 19 of the 30 conditions and above SentLDA in 13, and it is the second-highest model in 2, and SentLDA is ahead of GSLDA in every condition. On topic diversity, SAM lies between 0.542 and 1.000 and falls below 0.98 in 11 of the 30 conditions, GSLDA falls below 0.98 in 19 of them, and ConTM is low overall.
\begin{sidewaystable}
\centering
\captionsetup{font=footnotesize}
\footnotesize
\setlength{\tabcolsep}{2pt}
\caption{Classification accuracy (\%) of logistic regression with document-topic distributions as features. Values are means~$\pm$~standard deviations (in smaller type) over 5 independent runs. The sentence encoder is MiniLM. \textbf{Bold}/\underline{underline} indicate the best/second-best score within each dataset--topic-count--category row.}
\label{tab:cls_logreg_minilm}
\footnotesize
\colcellreset
\colcellfit{\ms{44.93}{2.42}}
\colcellfit{\underline{\ms{48.33}{0.77}}}
\colcellfit{\textbf{\ms{50.11}{1.64}}}
\colcellfit{\ms{59.42}{10.05}}
\colcellfit{LDA}
\colcellfit{SentLDA}
\colcellfit{SAM}
\colcellfit{GLDA}
\colcellfit{vLDA}
\colcellfit{ETM}
\colcellfit{ConTM}
\colcellfit{GSLDA}
\colcellfit{\textsc{vSLDA}}
\colcellfit{\colcite[\tiny]{Blei, 2003}}
\colcellfit{\colcite[\tiny]{Balikas, 2016}}
\colcellfit{\colcite[\tiny]{Reisinger, 2010}}
\colcellfit{\colcite[\tiny]{Das, 2015}}
\colcellfit{\colcite[\tiny]{Li, 2016}}
\colcellfit{\colcite[\tiny]{Dieng, 2020}}
\colcellfit{\colcite[\tiny]{Bianchi, 2021}}
\colcellfit{\colcite[\tiny]{Cha, 2024}}
\colcellfit{(proposed)}
\colcellfit{{\scriptsize word-level}}
\colcellfit{{\scriptsize sentence-level}}
\colcellfit{{\scriptsize document-level}}
\adjustbox{max width=\textheight}{%
\begin{tabular}{c c l *{9}{>{\centering\arraybackslash}p{\colcellwidth}}}
\toprule
\multirow[b]{4}{*}{\rotatebox[origin=c]{90}{Dataset}} & \multirow[b]{4}{*}{\rotatebox[origin=c]{90}{\#Topics}} & \multirow[b]{4}{*}{Category} & LDA & SentLDA & SAM & GLDA & vLDA & ETM & ConTM & GSLDA & \textsc{vSLDA} \\
 & & & \colcite[\tiny]{Blei, 2003} & \colcite[\tiny]{Balikas, 2016} & \colcite[\tiny]{Reisinger, 2010} & \colcite[\tiny]{Das, 2015} & \colcite[\tiny]{Li, 2016} & \colcite[\tiny]{Dieng, 2020} & \colcite[\tiny]{Bianchi, 2021} & \colcite[\tiny]{Cha, 2024} & (proposed) \\
\cmidrule(lr){4-4} \cmidrule(lr){5-5} \cmidrule(lr){6-6} \cmidrule(lr){7-9} \cmidrule(lr){10-10} \cmidrule(lr){11-12}
 & & & \multicolumn{1}{c}{\scriptsize word-level} & \multicolumn{1}{c}{\scriptsize sentence-level} & \multicolumn{1}{c}{\scriptsize document-level} & \multicolumn{3}{c}{\scriptsize word-level} & \multicolumn{1}{c}{\scriptsize document-level} & \multicolumn{2}{c}{\scriptsize sentence-level} \\
\cmidrule(lr){4-6} \cmidrule(lr){7-12}
 & & & \multicolumn{3}{c}{\scriptsize vocabulary-based} & \multicolumn{6}{c}{\scriptsize embedding-based} \\
\midrule
\multirow{18}{*}{\rotatebox[origin=c]{90}{20news}} & \multirow{6}{*}{10} & Computer & \ms{44.93}{2.42} & \underline{\ms{48.33}{0.77}} & \ms{30.31}{0.33} & \ms{27.93}{0.53} & \ms{27.99}{1.45} & \ms{42.59}{1.67} & \ms{39.54}{0.27} & \ms{44.79}{2.25} & \textbf{\ms{50.11}{1.64}} \\
 &  & Ride & \ms{63.81}{2.04} & \ms{68.80}{2.03} & \ms{56.72}{1.21} & \ms{57.47}{1.60} & \ms{58.51}{0.93} & \underline{\ms{69.28}{5.73}} & \ms{67.63}{5.61} & \ms{54.59}{3.11} & \textbf{\ms{79.57}{1.22}} \\
 &  & Sports & \ms{73.52}{3.14} & \underline{\ms{82.33}{2.28}} & \ms{56.98}{1.06} & \ms{56.97}{3.27} & \ms{55.69}{0.83} & \ms{74.46}{4.38} & \ms{70.25}{5.18} & \ms{54.00}{1.57} & \textbf{\ms{85.19}{0.92}} \\
 &  & Science & \ms{63.57}{9.24} & \underline{\ms{75.81}{3.85}} & \ms{38.88}{0.40} & \ms{39.03}{1.63} & \ms{36.23}{4.83} & \ms{75.02}{2.56} & \ms{48.82}{7.73} & \ms{72.67}{4.34} & \textbf{\ms{81.26}{0.54}} \\
 &  & Religion & \ms{51.92}{1.81} & \underline{\ms{53.35}{1.26}} & \ms{47.01}{0.41} & \ms{43.39}{0.47} & \ms{46.27}{1.82} & \ms{51.73}{1.90} & \ms{46.80}{1.22} & \ms{46.95}{2.37} & \textbf{\ms{55.99}{0.72}} \\
 &  & Politics & \ms{62.74}{4.32} & \underline{\ms{68.20}{2.79}} & \ms{41.16}{0.49} & \ms{43.65}{1.76} & \ms{51.37}{0.40} & \ms{67.92}{1.93} & \ms{46.18}{6.34} & \ms{57.89}{7.22} & \textbf{\ms{75.38}{0.92}} \\
\cmidrule(lr){2-12}
 & \multirow{6}{*}{20} & Computer & \ms{48.52}{2.59} & \underline{\ms{51.82}{1.60}} & \ms{38.72}{0.82} & \ms{32.64}{0.97} & \ms{32.18}{1.83} & \ms{45.61}{2.41} & \ms{40.33}{0.82} & \ms{46.01}{1.96} & \textbf{\ms{55.67}{2.06}} \\
 &  & Ride & \ms{66.64}{2.59} & \underline{\ms{71.97}{2.31}} & \ms{70.06}{0.71} & \ms{62.11}{2.95} & \ms{60.35}{1.13} & \ms{61.01}{1.39} & \ms{67.52}{3.18} & \ms{55.49}{2.88} & \textbf{\ms{79.20}{0.81}} \\
 &  & Sports & \ms{71.59}{5.01} & \underline{\ms{82.62}{1.70}} & \ms{61.31}{2.28} & \ms{59.92}{1.30} & \ms{61.85}{1.17} & \ms{74.99}{3.30} & \ms{78.18}{2.48} & \ms{53.44}{0.84} & \textbf{\ms{86.51}{1.10}} \\
 &  & Science & \ms{71.32}{2.97} & \ms{76.74}{1.29} & \ms{44.52}{2.99} & \ms{41.81}{1.62} & \ms{49.48}{4.00} & \underline{\ms{77.74}{0.89}} & \ms{58.73}{4.28} & \ms{72.24}{4.11} & \textbf{\ms{82.51}{0.75}} \\
 &  & Religion & \ms{51.17}{1.86} & \underline{\ms{53.60}{1.41}} & \ms{48.59}{0.75} & \ms{43.52}{0.59} & \ms{47.14}{1.88} & \ms{46.33}{0.82} & \ms{47.19}{2.25} & \ms{46.44}{1.78} & \textbf{\ms{58.70}{1.09}} \\
 &  & Politics & \ms{63.34}{5.17} & \underline{\ms{69.95}{2.39}} & \ms{48.67}{2.69} & \ms{47.54}{2.29} & \ms{53.48}{2.04} & \ms{67.86}{1.70} & \ms{56.96}{6.55} & \ms{55.42}{6.56} & \textbf{\ms{75.14}{0.87}} \\
\cmidrule(lr){2-12}
 & \multirow{6}{*}{30} & Computer & \ms{49.90}{1.99} & \underline{\ms{52.70}{2.18}} & \ms{41.62}{0.36} & \ms{34.58}{1.56} & \ms{33.05}{1.31} & \ms{46.77}{1.79} & \ms{42.78}{2.01} & \ms{44.67}{2.94} & \textbf{\ms{57.02}{1.61}} \\
 &  & Ride & \ms{67.41}{2.33} & \ms{69.01}{1.94} & \ms{68.37}{0.75} & \ms{61.41}{2.35} & \ms{63.46}{3.10} & \ms{63.20}{3.49} & \underline{\ms{70.29}{5.37}} & \ms{53.73}{0.95} & \textbf{\ms{80.21}{1.38}} \\
 &  & Sports & \ms{73.79}{2.53} & \underline{\ms{80.45}{1.96}} & \ms{58.61}{2.93} & \ms{61.18}{1.19} & \ms{63.91}{1.90} & \ms{78.10}{3.11} & \ms{80.27}{2.86} & \ms{53.49}{1.21} & \textbf{\ms{86.37}{0.72}} \\
 &  & Science & \ms{69.10}{4.27} & \ms{77.47}{1.03} & \ms{48.52}{0.36} & \ms{43.73}{1.89} & \ms{59.29}{6.74} & \underline{\ms{78.69}{1.31}} & \ms{65.12}{2.07} & \ms{70.90}{3.65} & \textbf{\ms{82.29}{0.66}} \\
 &  & Religion & \ms{52.33}{1.88} & \underline{\ms{54.82}{0.19}} & \ms{49.64}{0.35} & \ms{43.54}{1.05} & \ms{49.06}{1.68} & \ms{45.08}{1.30} & \ms{48.61}{1.63} & \ms{46.01}{0.34} & \textbf{\ms{59.77}{0.33}} \\
 &  & Politics & \ms{66.99}{4.36} & \underline{\ms{70.27}{1.66}} & \ms{54.29}{1.38} & \ms{45.55}{2.87} & \ms{55.70}{2.83} & \ms{68.69}{2.88} & \ms{64.42}{1.59} & \ms{55.16}{6.61} & \textbf{\ms{76.05}{0.70}} \\
\midrule
\multirow{12}{*}{\rotatebox[origin=c]{90}{NYT}} & \multirow{4}{*}{10} & Arts & \ms{53.84}{6.46} & \underline{\ms{78.56}{5.11}} & \ms{58.03}{1.53} & \ms{40.48}{0.68} & \ms{42.50}{2.86} & \ms{63.99}{4.12} & \ms{56.68}{3.29} & \ms{39.80}{0.24} & \textbf{\ms{90.00}{3.56}} \\
 &  & Business & \ms{68.09}{2.36} & \underline{\ms{74.91}{3.23}} & \ms{74.70}{1.68} & \ms{52.37}{2.93} & \ms{49.47}{1.25} & \ms{51.70}{5.79} & \ms{70.64}{4.64} & \ms{67.18}{2.59} & \textbf{\ms{81.07}{1.44}} \\
 &  & Politics & \ms{52.27}{4.73} & \underline{\ms{65.40}{3.62}} & \ms{61.62}{1.33} & \ms{42.42}{0.00} & \ms{42.67}{0.39} & \ms{42.42}{0.00} & \ms{56.52}{2.32} & \ms{49.80}{2.14} & \textbf{\ms{70.65}{1.12}} \\
 &  & Sports & \textbf{\ms{89.43}{3.28}} & \underline{\ms{84.34}{5.17}} & \ms{64.73}{0.44} & \ms{36.78}{1.40} & \ms{31.94}{1.89} & \ms{75.58}{4.10} & \ms{62.93}{3.00} & \ms{64.64}{3.25} & \ms{82.09}{1.47} \\
\cmidrule(lr){2-12}
 & \multirow{4}{*}{20} & Arts & \ms{53.80}{7.15} & \underline{\ms{83.89}{0.96}} & \ms{68.99}{1.20} & \ms{40.14}{0.40} & \ms{49.71}{7.62} & \ms{64.47}{1.94} & \ms{60.15}{2.48} & \ms{39.80}{0.24} & \textbf{\ms{91.11}{0.66}} \\
 &  & Business & \ms{70.48}{4.39} & \underline{\ms{80.66}{3.97}} & \ms{74.05}{1.10} & \ms{48.04}{0.83} & \ms{51.66}{2.92} & \ms{54.40}{4.15} & \ms{70.23}{3.42} & \ms{66.51}{2.40} & \textbf{\ms{83.82}{1.59}} \\
 &  & Politics & \ms{58.69}{1.85} & \underline{\ms{74.75}{2.43}} & \ms{60.50}{2.05} & \ms{42.42}{0.00} & \ms{43.94}{1.49} & \ms{42.42}{0.00} & \ms{56.52}{1.74} & \ms{47.88}{0.38} & \textbf{\ms{77.17}{1.76}} \\
 &  & Sports & \textbf{\ms{95.30}{2.12}} & \underline{\ms{94.92}{0.81}} & \ms{80.41}{1.44} & \ms{36.38}{2.17} & \ms{36.08}{3.68} & \ms{81.39}{1.94} & \ms{78.07}{1.32} & \ms{76.65}{4.85} & \ms{93.58}{0.86} \\
\cmidrule(lr){2-12}
 & \multirow{4}{*}{30} & Arts & \ms{53.13}{9.51} & \underline{\ms{78.75}{0.69}} & \ms{60.48}{2.09} & \ms{40.00}{0.56} & \ms{46.11}{7.54} & \ms{59.42}{10.05} & \ms{61.92}{2.42} & \ms{39.47}{0.38} & \textbf{\ms{90.82}{0.85}} \\
 &  & Business & \ms{73.38}{2.97} & \underline{\ms{83.36}{2.51}} & \ms{70.13}{1.10} & \ms{47.22}{0.31} & \ms{52.11}{1.99} & \ms{51.30}{5.38} & \ms{71.15}{2.17} & \ms{62.60}{5.48} & \textbf{\ms{84.68}{2.17}} \\
 &  & Politics & \ms{61.97}{1.28} & \underline{\ms{74.65}{2.28}} & \ms{56.72}{1.84} & \ms{42.42}{0.00} & \ms{44.70}{1.40} & \ms{42.42}{0.00} & \ms{57.22}{0.72} & \ms{43.58}{2.32} & \textbf{\ms{79.60}{1.42}} \\
 &  & Sports & \textbf{\ms{95.23}{2.86}} & \underline{\ms{95.00}{0.45}} & \ms{85.42}{0.36} & \ms{36.41}{0.85} & \ms{41.13}{2.72} & \ms{84.12}{2.87} & \ms{83.18}{0.93} & \ms{73.26}{2.02} & \ms{94.73}{0.50} \\
\bottomrule
\end{tabular}
}
\end{sidewaystable}

\begin{sidewaystable}
\centering
\captionsetup{font=footnotesize}
\footnotesize
\setlength{\tabcolsep}{2pt}
\caption{Topic coherence measured by $C_V$ using the top 10 topic words and English Wikipedia as the reference corpus. Values are means~$\pm$~standard deviations (in smaller type) over 5 independent runs. The sentence encoder is MiniLM. \textbf{Bold}/\underline{underline} indicate the best/second-best score within each dataset--topic-count--category row.}
\label{tab:topic_coherence_c_v}
\footnotesize
\colcellreset
\colcellfit{\ms{0.345}{0.018}}
\colcellfit{\underline{\ms{0.431}{0.032}}}
\colcellfit{\textbf{\ms{0.525}{0.026}}}
\colcellfit{LDA}
\colcellfit{SentLDA}
\colcellfit{SAM}
\colcellfit{GLDA}
\colcellfit{vLDA}
\colcellfit{ETM}
\colcellfit{ConTM}
\colcellfit{GSLDA}
\colcellfit{\textsc{vSLDA}}
\colcellfit{\colcite[\tiny]{Blei, 2003}}
\colcellfit{\colcite[\tiny]{Balikas, 2016}}
\colcellfit{\colcite[\tiny]{Reisinger, 2010}}
\colcellfit{\colcite[\tiny]{Das, 2015}}
\colcellfit{\colcite[\tiny]{Li, 2016}}
\colcellfit{\colcite[\tiny]{Dieng, 2020}}
\colcellfit{\colcite[\tiny]{Bianchi, 2021}}
\colcellfit{\colcite[\tiny]{Cha, 2024}}
\colcellfit{(proposed)}
\colcellfit{{\scriptsize word-level}}
\colcellfit{{\scriptsize sentence-level}}
\colcellfit{{\scriptsize document-level}}
\adjustbox{max width=\textheight}{%
\begin{tabular}{c c l *{9}{>{\centering\arraybackslash}p{\colcellwidth}}}
\toprule
\multirow[b]{4}{*}{\rotatebox[origin=c]{90}{Dataset}} & \multirow[b]{4}{*}{\rotatebox[origin=c]{90}{\#Topics}} & \multirow[b]{4}{*}{Category} & LDA & SentLDA & SAM & GLDA & vLDA & ETM & ConTM & GSLDA & \textsc{vSLDA} \\
 & & & \colcite[\tiny]{Blei, 2003} & \colcite[\tiny]{Balikas, 2016} & \colcite[\tiny]{Reisinger, 2010} & \colcite[\tiny]{Das, 2015} & \colcite[\tiny]{Li, 2016} & \colcite[\tiny]{Dieng, 2020} & \colcite[\tiny]{Bianchi, 2021} & \colcite[\tiny]{Cha, 2024} & (proposed) \\
\cmidrule(lr){4-4} \cmidrule(lr){5-5} \cmidrule(lr){6-6} \cmidrule(lr){7-9} \cmidrule(lr){10-10} \cmidrule(lr){11-12}
 & & & \multicolumn{1}{c}{\scriptsize word-level} & \multicolumn{1}{c}{\scriptsize sentence-level} & \multicolumn{1}{c}{\scriptsize document-level} & \multicolumn{3}{c}{\scriptsize word-level} & \multicolumn{1}{c}{\scriptsize document-level} & \multicolumn{2}{c}{\scriptsize sentence-level} \\
\cmidrule(lr){4-6} \cmidrule(lr){7-12}
 & & & \multicolumn{3}{c}{\scriptsize vocabulary-based} & \multicolumn{6}{c}{\scriptsize embedding-based} \\
\midrule
\multirow{18}{*}{\rotatebox[origin=c]{90}{20news}} & \multirow{6}{*}{10} & Computer & \ms{0.345}{0.018} & \ms{0.363}{0.015} & \ms{0.326}{0.008} & \ms{0.430}{0.010} & \underline{\ms{0.431}{0.032}} & \textbf{\ms{0.525}{0.026}} & \ms{0.192}{0.021} & \ms{0.297}{0.032} & \ms{0.429}{0.013} \\
 &  & Ride & \ms{0.277}{0.007} & \ms{0.354}{0.035} & \ms{0.333}{0.004} & \ms{0.406}{0.009} & \ms{0.383}{0.014} & \underline{\ms{0.495}{0.016}} & \ms{0.215}{0.016} & \ms{0.252}{0.020} & \textbf{\ms{0.513}{0.017}} \\
 &  & Sports & \ms{0.289}{0.017} & \ms{0.315}{0.039} & \ms{0.368}{0.008} & \ms{0.394}{0.012} & \underline{\ms{0.454}{0.029}} & \textbf{\ms{0.514}{0.028}} & \ms{0.228}{0.009} & \ms{0.257}{0.014} & \ms{0.417}{0.029} \\
 &  & Science & \ms{0.437}{0.052} & \ms{0.485}{0.033} & \ms{0.349}{0.011} & \ms{0.402}{0.013} & \ms{0.354}{0.022} & \underline{\ms{0.573}{0.014}} & \ms{0.252}{0.028} & \ms{0.383}{0.038} & \textbf{\ms{0.616}{0.032}} \\
 &  & Religion & \ms{0.343}{0.009} & \ms{0.365}{0.027} & \ms{0.364}{0.009} & \ms{0.430}{0.007} & \ms{0.380}{0.035} & \underline{\ms{0.556}{0.017}} & \ms{0.210}{0.011} & \ms{0.276}{0.017} & \textbf{\ms{0.589}{0.022}} \\
 &  & Politics & \ms{0.417}{0.048} & \ms{0.379}{0.018} & \ms{0.364}{0.010} & \ms{0.397}{0.008} & \ms{0.391}{0.039} & \underline{\ms{0.518}{0.013}} & \ms{0.228}{0.021} & \ms{0.270}{0.024} & \textbf{\ms{0.556}{0.064}} \\
\cmidrule(lr){2-12}
 & \multirow{6}{*}{20} & Computer & \ms{0.300}{0.008} & \ms{0.285}{0.019} & \ms{0.330}{0.023} & \ms{0.402}{0.004} & \underline{\ms{0.421}{0.023}} & \textbf{\ms{0.507}{0.019}} & \ms{0.194}{0.009} & \ms{0.213}{0.012} & \ms{0.371}{0.016} \\
 &  & Ride & \ms{0.263}{0.010} & \ms{0.276}{0.016} & \ms{0.321}{0.007} & \ms{0.391}{0.006} & \ms{0.386}{0.012} & \textbf{\ms{0.482}{0.030}} & \ms{0.228}{0.004} & \ms{0.254}{0.015} & \underline{\ms{0.450}{0.030}} \\
 &  & Sports & \ms{0.273}{0.010} & \ms{0.302}{0.010} & \ms{0.330}{0.006} & \ms{0.370}{0.004} & \underline{\ms{0.449}{0.009}} & \textbf{\ms{0.513}{0.033}} & \ms{0.238}{0.011} & \ms{0.247}{0.008} & \ms{0.383}{0.010} \\
 &  & Science & \ms{0.360}{0.014} & \ms{0.407}{0.022} & \ms{0.354}{0.016} & \ms{0.382}{0.005} & \ms{0.388}{0.020} & \underline{\ms{0.528}{0.004}} & \ms{0.254}{0.009} & \ms{0.290}{0.026} & \textbf{\ms{0.530}{0.031}} \\
 &  & Religion & \ms{0.314}{0.021} & \ms{0.320}{0.013} & \ms{0.352}{0.006} & \ms{0.415}{0.007} & \ms{0.387}{0.012} & \underline{\ms{0.527}{0.005}} & \ms{0.232}{0.010} & \ms{0.252}{0.005} & \textbf{\ms{0.528}{0.013}} \\
 &  & Politics & \ms{0.326}{0.010} & \ms{0.346}{0.029} & \ms{0.350}{0.018} & \ms{0.382}{0.004} & \ms{0.388}{0.024} & \textbf{\ms{0.505}{0.023}} & \ms{0.236}{0.026} & \ms{0.217}{0.010} & \underline{\ms{0.496}{0.014}} \\
\cmidrule(lr){2-12}
 & \multirow{6}{*}{30} & Computer & \ms{0.283}{0.007} & \ms{0.252}{0.010} & \ms{0.337}{0.016} & \underline{\ms{0.394}{0.006}} & \ms{0.394}{0.018} & \textbf{\ms{0.515}{0.007}} & \ms{0.189}{0.010} & \ms{0.195}{0.009} & \ms{0.354}{0.016} \\
 &  & Ride & \ms{0.267}{0.018} & \ms{0.283}{0.014} & \ms{0.332}{0.009} & \ms{0.376}{0.006} & \ms{0.407}{0.006} & \textbf{\ms{0.488}{0.014}} & \ms{0.238}{0.008} & \ms{0.264}{0.007} & \underline{\ms{0.421}{0.016}} \\
 &  & Sports & \ms{0.284}{0.013} & \ms{0.287}{0.009} & \ms{0.318}{0.005} & \ms{0.359}{0.005} & \underline{\ms{0.462}{0.007}} & \textbf{\ms{0.534}{0.011}} & \ms{0.223}{0.014} & \ms{0.264}{0.007} & \ms{0.382}{0.014} \\
 &  & Science & \ms{0.345}{0.013} & \ms{0.339}{0.019} & \ms{0.327}{0.006} & \ms{0.376}{0.003} & \ms{0.392}{0.016} & \textbf{\ms{0.513}{0.010}} & \ms{0.267}{0.015} & \ms{0.276}{0.012} & \underline{\ms{0.484}{0.031}} \\
 &  & Religion & \ms{0.292}{0.014} & \ms{0.308}{0.016} & \ms{0.336}{0.008} & \ms{0.403}{0.006} & \ms{0.386}{0.021} & \textbf{\ms{0.515}{0.025}} & \ms{0.224}{0.005} & \ms{0.240}{0.008} & \underline{\ms{0.489}{0.003}} \\
 &  & Politics & \ms{0.330}{0.004} & \ms{0.294}{0.029} & \ms{0.375}{0.018} & \ms{0.370}{0.003} & \ms{0.362}{0.009} & \textbf{\ms{0.506}{0.021}} & \ms{0.247}{0.006} & \ms{0.215}{0.011} & \underline{\ms{0.473}{0.023}} \\
\midrule
\multirow{12}{*}{\rotatebox[origin=c]{90}{NYT}} & \multirow{4}{*}{10} & Arts & \ms{0.364}{0.026} & \underline{\ms{0.495}{0.044}} & \ms{0.391}{0.022} & \ms{0.424}{0.005} & \ms{0.374}{0.018} & \ms{0.493}{0.045} & \ms{0.272}{0.018} & \ms{0.339}{0.011} & \textbf{\ms{0.585}{0.029}} \\
 &  & Business & \ms{0.409}{0.031} & \ms{0.472}{0.030} & \underline{\ms{0.495}{0.029}} & \ms{0.399}{0.007} & \ms{0.323}{0.036} & \ms{0.425}{0.022} & \ms{0.300}{0.032} & \ms{0.370}{0.016} & \textbf{\ms{0.565}{0.043}} \\
 &  & Politics & \ms{0.433}{0.041} & \ms{0.496}{0.023} & \underline{\ms{0.506}{0.024}} & \ms{0.397}{0.009} & \ms{0.355}{0.023} & \ms{0.439}{0.036} & \ms{0.335}{0.028} & \ms{0.311}{0.022} & \textbf{\ms{0.647}{0.039}} \\
 &  & Sports & \ms{0.619}{0.030} & \textbf{\ms{0.650}{0.034}} & \ms{0.580}{0.035} & \ms{0.421}{0.006} & \ms{0.465}{0.029} & \ms{0.581}{0.012} & \ms{0.366}{0.022} & \ms{0.498}{0.053} & \underline{\ms{0.635}{0.033}} \\
\cmidrule(lr){2-12}
 & \multirow{4}{*}{20} & Arts & \ms{0.333}{0.009} & \ms{0.443}{0.020} & \ms{0.377}{0.015} & \ms{0.397}{0.008} & \ms{0.383}{0.025} & \underline{\ms{0.458}{0.024}} & \ms{0.293}{0.012} & \ms{0.299}{0.007} & \textbf{\ms{0.495}{0.018}} \\
 &  & Business & \ms{0.358}{0.029} & \ms{0.430}{0.010} & \ms{0.419}{0.022} & \ms{0.385}{0.005} & \ms{0.346}{0.014} & \underline{\ms{0.441}{0.034}} & \ms{0.294}{0.025} & \ms{0.333}{0.019} & \textbf{\ms{0.555}{0.031}} \\
 &  & Politics & \ms{0.374}{0.017} & \underline{\ms{0.462}{0.036}} & \ms{0.431}{0.014} & \ms{0.387}{0.005} & \ms{0.326}{0.017} & \ms{0.456}{0.039} & \ms{0.338}{0.032} & \ms{0.295}{0.007} & \textbf{\ms{0.573}{0.028}} \\
 &  & Sports & \ms{0.551}{0.046} & \underline{\ms{0.591}{0.035}} & \ms{0.533}{0.023} & \ms{0.388}{0.007} & \ms{0.383}{0.042} & \ms{0.587}{0.009} & \ms{0.401}{0.014} & \ms{0.448}{0.017} & \textbf{\ms{0.602}{0.024}} \\
\cmidrule(lr){2-12}
 & \multirow{4}{*}{30} & Arts & \ms{0.309}{0.020} & \ms{0.377}{0.015} & \ms{0.299}{0.015} & \ms{0.387}{0.004} & \ms{0.357}{0.022} & \underline{\ms{0.459}{0.031}} & \ms{0.298}{0.015} & \ms{0.284}{0.009} & \textbf{\ms{0.461}{0.019}} \\
 &  & Business & \ms{0.348}{0.022} & \ms{0.408}{0.018} & \ms{0.318}{0.012} & \ms{0.377}{0.003} & \ms{0.336}{0.026} & \underline{\ms{0.434}{0.018}} & \ms{0.277}{0.020} & \ms{0.342}{0.015} & \textbf{\ms{0.519}{0.012}} \\
 &  & Politics & \ms{0.371}{0.017} & \ms{0.420}{0.024} & \ms{0.368}{0.021} & \ms{0.376}{0.003} & \ms{0.341}{0.028} & \underline{\ms{0.436}{0.017}} & \ms{0.341}{0.022} & \ms{0.273}{0.011} & \textbf{\ms{0.530}{0.022}} \\
 &  & Sports & \ms{0.506}{0.037} & \ms{0.504}{0.007} & \ms{0.488}{0.015} & \ms{0.376}{0.005} & \ms{0.377}{0.011} & \textbf{\ms{0.576}{0.011}} & \ms{0.388}{0.013} & \ms{0.382}{0.032} & \underline{\ms{0.552}{0.011}} \\
\bottomrule
\end{tabular}
}
\end{sidewaystable}

\begin{sidewaystable}
\centering
\captionsetup{font=footnotesize}
\footnotesize
\setlength{\tabcolsep}{2pt}
\caption{Topic diversity measured as the proportion of unique words among the top 25 words of all topics. Values are means~$\pm$~standard deviations (in smaller type) over 5 independent runs. The sentence encoder is MiniLM. The metric is left unmarked because it is near its ceiling for most models and is read here as a check against redundant topics rather than as a comparison between them.}
\label{tab:topic_diversity}
\footnotesize
\colcellreset
\colcellfit{\ms{0.998}{0.003}}
\colcellfit{LDA}
\colcellfit{SentLDA}
\colcellfit{SAM}
\colcellfit{GLDA}
\colcellfit{vLDA}
\colcellfit{ETM}
\colcellfit{ConTM}
\colcellfit{GSLDA}
\colcellfit{\textsc{vSLDA}}
\colcellfit{\colcite[\tiny]{Blei, 2003}}
\colcellfit{\colcite[\tiny]{Balikas, 2016}}
\colcellfit{\colcite[\tiny]{Reisinger, 2010}}
\colcellfit{\colcite[\tiny]{Das, 2015}}
\colcellfit{\colcite[\tiny]{Li, 2016}}
\colcellfit{\colcite[\tiny]{Dieng, 2020}}
\colcellfit{\colcite[\tiny]{Bianchi, 2021}}
\colcellfit{\colcite[\tiny]{Cha, 2024}}
\colcellfit{(proposed)}
\colcellfit{{\scriptsize word-level}}
\colcellfit{{\scriptsize sentence-level}}
\colcellfit{{\scriptsize document-level}}
\adjustbox{max width=\textheight}{%
\begin{tabular}{c c l *{9}{>{\centering\arraybackslash}p{\colcellwidth}}}
\toprule
\multirow[b]{4}{*}{\rotatebox[origin=c]{90}{Dataset}} & \multirow[b]{4}{*}{\rotatebox[origin=c]{90}{\#Topics}} & \multirow[b]{4}{*}{Category} & LDA & SentLDA & SAM & GLDA & vLDA & ETM & ConTM & GSLDA & \textsc{vSLDA} \\
 & & & \colcite[\tiny]{Blei, 2003} & \colcite[\tiny]{Balikas, 2016} & \colcite[\tiny]{Reisinger, 2010} & \colcite[\tiny]{Das, 2015} & \colcite[\tiny]{Li, 2016} & \colcite[\tiny]{Dieng, 2020} & \colcite[\tiny]{Bianchi, 2021} & \colcite[\tiny]{Cha, 2024} & (proposed) \\
\cmidrule(lr){4-4} \cmidrule(lr){5-5} \cmidrule(lr){6-6} \cmidrule(lr){7-9} \cmidrule(lr){10-10} \cmidrule(lr){11-12}
 & & & \multicolumn{1}{c}{\scriptsize word-level} & \multicolumn{1}{c}{\scriptsize sentence-level} & \multicolumn{1}{c}{\scriptsize document-level} & \multicolumn{3}{c}{\scriptsize word-level} & \multicolumn{1}{c}{\scriptsize document-level} & \multicolumn{2}{c}{\scriptsize sentence-level} \\
\cmidrule(lr){4-6} \cmidrule(lr){7-12}
 & & & \multicolumn{3}{c}{\scriptsize vocabulary-based} & \multicolumn{6}{c}{\scriptsize embedding-based} \\
\midrule
\multirow{18}{*}{\rotatebox[origin=c]{90}{20news}} & \multirow{6}{*}{10} & Computer & \ms{0.998}{0.003} & \ms{1.000}{0.000} & \ms{0.989}{0.010} & \ms{1.000}{0.000} & \ms{0.801}{0.042} & \ms{1.000}{0.000} & \ms{0.784}{0.055} & \ms{0.998}{0.002} & \ms{1.000}{0.000} \\
 &  & Ride & \ms{0.995}{0.005} & \ms{1.000}{0.000} & \ms{1.000}{0.000} & \ms{1.000}{0.000} & \ms{0.722}{0.050} & \ms{0.973}{0.007} & \ms{0.730}{0.012} & \ms{0.990}{0.010} & \ms{1.000}{0.000} \\
 &  & Sports & \ms{0.996}{0.004} & \ms{1.000}{0.000} & \ms{1.000}{0.000} & \ms{1.000}{0.000} & \ms{0.785}{0.038} & \ms{0.970}{0.007} & \ms{0.778}{0.013} & \ms{0.986}{0.007} & \ms{1.000}{0.000} \\
 &  & Science & \ms{0.998}{0.002} & \ms{1.000}{0.000} & \ms{0.996}{0.006} & \ms{1.000}{0.000} & \ms{0.675}{0.094} & \ms{0.998}{0.002} & \ms{0.798}{0.026} & \ms{1.000}{0.000} & \ms{0.999}{0.002} \\
 &  & Religion & \ms{1.000}{0.000} & \ms{1.000}{0.000} & \ms{0.998}{0.002} & \ms{1.000}{0.000} & \ms{0.650}{0.087} & \ms{0.994}{0.004} & \ms{0.738}{0.022} & \ms{0.998}{0.003} & \ms{0.998}{0.003} \\
 &  & Politics & \ms{0.998}{0.003} & \ms{0.999}{0.002} & \ms{0.996}{0.003} & \ms{1.000}{0.000} & \ms{0.761}{0.071} & \ms{0.994}{0.007} & \ms{0.717}{0.042} & \ms{1.000}{0.000} & \ms{0.999}{0.002} \\
\cmidrule(lr){2-12}
 & \multirow{6}{*}{20} & Computer & \ms{0.991}{0.007} & \ms{0.999}{0.001} & \ms{0.964}{0.008} & \ms{1.000}{0.000} & \ms{0.674}{0.039} & \ms{0.996}{0.001} & \ms{0.716}{0.019} & \ms{0.991}{0.004} & \ms{0.994}{0.005} \\
 &  & Ride & \ms{0.971}{0.009} & \ms{0.997}{0.003} & \ms{0.989}{0.010} & \ms{1.000}{0.000} & \ms{0.548}{0.051} & \ms{0.870}{0.010} & \ms{0.652}{0.025} & \ms{0.949}{0.012} & \ms{0.995}{0.003} \\
 &  & Sports & \ms{0.975}{0.014} & \ms{0.998}{0.002} & \ms{0.990}{0.002} & \ms{1.000}{0.000} & \ms{0.564}{0.032} & \ms{0.879}{0.021} & \ms{0.682}{0.023} & \ms{0.938}{0.003} & \ms{0.994}{0.002} \\
 &  & Science & \ms{0.982}{0.006} & \ms{0.999}{0.002} & \ms{0.976}{0.006} & \ms{1.000}{0.000} & \ms{0.638}{0.051} & \ms{0.982}{0.008} & \ms{0.723}{0.022} & \ms{0.989}{0.005} & \ms{0.998}{0.001} \\
 &  & Religion & \ms{0.984}{0.008} & \ms{0.998}{0.001} & \ms{0.993}{0.006} & \ms{1.000}{0.000} & \ms{0.563}{0.030} & \ms{0.953}{0.016} & \ms{0.617}{0.039} & \ms{0.956}{0.005} & \ms{0.997}{0.001} \\
 &  & Politics & \ms{0.988}{0.006} & \ms{1.000}{0.001} & \ms{0.984}{0.001} & \ms{1.000}{0.000} & \ms{0.599}{0.030} & \ms{0.944}{0.012} & \ms{0.657}{0.033} & \ms{0.992}{0.003} & \ms{0.996}{0.001} \\
\cmidrule(lr){2-12}
 & \multirow{6}{*}{30} & Computer & \ms{0.975}{0.013} & \ms{0.998}{0.002} & \ms{0.935}{0.002} & \ms{1.000}{0.000} & \ms{0.567}{0.060} & \ms{0.964}{0.013} & \ms{0.645}{0.043} & \ms{0.974}{0.004} & \ms{0.994}{0.005} \\
 &  & Ride & \ms{0.941}{0.003} & \ms{0.986}{0.002} & \ms{0.972}{0.007} & \ms{1.000}{0.000} & \ms{0.546}{0.043} & \ms{0.809}{0.027} & \ms{0.589}{0.014} & \ms{0.898}{0.008} & \ms{0.978}{0.003} \\
 &  & Sports & \ms{0.955}{0.006} & \ms{0.984}{0.005} & \ms{0.970}{0.011} & \ms{1.000}{0.000} & \ms{0.538}{0.050} & \ms{0.804}{0.008} & \ms{0.590}{0.010} & \ms{0.873}{0.008} & \ms{0.982}{0.004} \\
 &  & Science & \ms{0.977}{0.007} & \ms{0.997}{0.002} & \ms{0.961}{0.006} & \ms{1.000}{0.000} & \ms{0.616}{0.041} & \ms{0.936}{0.011} & \ms{0.677}{0.014} & \ms{0.972}{0.003} & \ms{0.995}{0.001} \\
 &  & Religion & \ms{0.963}{0.010} & \ms{0.995}{0.002} & \ms{0.980}{0.008} & \ms{1.000}{0.000} & \ms{0.501}{0.059} & \ms{0.887}{0.015} & \ms{0.571}{0.012} & \ms{0.922}{0.005} & \ms{0.991}{0.005} \\
 &  & Politics & \ms{0.972}{0.009} & \ms{0.999}{0.001} & \ms{0.985}{0.003} & \ms{1.000}{0.000} & \ms{0.495}{0.031} & \ms{0.893}{0.016} & \ms{0.598}{0.010} & \ms{0.973}{0.008} & \ms{0.995}{0.001} \\
\midrule
\multirow{12}{*}{\rotatebox[origin=c]{90}{NYT}} & \multirow{4}{*}{10} & Arts & \ms{0.998}{0.003} & \ms{1.000}{0.000} & \ms{0.999}{0.002} & \ms{1.000}{0.000} & \ms{0.709}{0.038} & \ms{0.928}{0.015} & \ms{0.838}{0.044} & \ms{0.734}{0.030} & \ms{0.999}{0.002} \\
 &  & Business & \ms{0.986}{0.019} & \ms{1.000}{0.000} & \ms{0.997}{0.002} & \ms{1.000}{0.000} & \ms{0.667}{0.056} & \ms{0.907}{0.023} & \ms{0.918}{0.031} & \ms{0.977}{0.012} & \ms{0.999}{0.002} \\
 &  & Politics & \ms{0.994}{0.005} & \ms{1.000}{0.000} & \ms{0.998}{0.002} & \ms{1.000}{0.000} & \ms{0.681}{0.083} & \ms{0.908}{0.030} & \ms{0.938}{0.009} & \ms{0.979}{0.009} & \ms{0.999}{0.002} \\
 &  & Sports & \ms{1.000}{0.000} & \ms{0.999}{0.002} & \ms{0.994}{0.005} & \ms{1.000}{0.000} & \ms{0.781}{0.071} & \ms{1.000}{0.000} & \ms{0.985}{0.017} & \ms{1.000}{0.000} & \ms{0.999}{0.002} \\
\cmidrule(lr){2-12}
 & \multirow{4}{*}{20} & Arts & \ms{0.960}{0.026} & \ms{0.998}{0.002} & \ms{0.999}{0.001} & \ms{1.000}{0.000} & \ms{0.660}{0.039} & \ms{0.807}{0.022} & \ms{0.761}{0.016} & \ms{0.403}{0.026} & \ms{0.994}{0.003} \\
 &  & Business & \ms{0.884}{0.059} & \ms{0.996}{0.003} & \ms{0.982}{0.025} & \ms{1.000}{0.000} & \ms{0.600}{0.028} & \ms{0.795}{0.030} & \ms{0.778}{0.047} & \ms{0.919}{0.010} & \ms{0.993}{0.006} \\
 &  & Politics & \ms{0.934}{0.023} & \ms{0.997}{0.002} & \ms{0.957}{0.038} & \ms{1.000}{0.000} & \ms{0.574}{0.038} & \ms{0.744}{0.022} & \ms{0.833}{0.031} & \ms{0.927}{0.009} & \ms{0.996}{0.003} \\
 &  & Sports & \ms{0.995}{0.006} & \ms{0.996}{0.003} & \ms{0.985}{0.005} & \ms{1.000}{0.000} & \ms{0.590}{0.059} & \ms{0.999}{0.002} & \ms{0.898}{0.018} & \ms{0.995}{0.002} & \ms{0.995}{0.004} \\
\cmidrule(lr){2-12}
 & \multirow{4}{*}{30} & Arts & \ms{0.863}{0.036} & \ms{0.979}{0.024} & \ms{0.755}{0.087} & \ms{1.000}{0.000} & \ms{0.575}{0.048} & \ms{0.730}{0.027} & \ms{0.715}{0.029} & \ms{0.281}{0.011} & \ms{0.991}{0.003} \\
 &  & Business & \ms{0.835}{0.070} & \ms{0.987}{0.004} & \ms{0.542}{0.039} & \ms{1.000}{0.000} & \ms{0.562}{0.043} & \ms{0.656}{0.029} & \ms{0.684}{0.022} & \ms{0.793}{0.044} & \ms{0.981}{0.005} \\
 &  & Politics & \ms{0.886}{0.021} & \ms{0.988}{0.003} & \ms{0.582}{0.062} & \ms{1.000}{0.000} & \ms{0.546}{0.040} & \ms{0.653}{0.022} & \ms{0.722}{0.032} & \ms{0.864}{0.009} & \ms{0.986}{0.005} \\
 &  & Sports & \ms{0.987}{0.004} & \ms{0.995}{0.005} & \ms{0.978}{0.007} & \ms{1.000}{0.000} & \ms{0.562}{0.006} & \ms{0.992}{0.006} & \ms{0.859}{0.009} & \ms{0.975}{0.005} & \ms{0.989}{0.002} \\
\bottomrule
\end{tabular}
}
\end{sidewaystable}

\begin{sidewaystable}
\centering
\captionsetup{font=footnotesize}
\footnotesize
\setlength{\tabcolsep}{2pt}
\caption{Topic quality, the product of $C_V$ coherence (Table~\ref{tab:topic_coherence_c_v}) and topic diversity (Table~\ref{tab:topic_diversity}), formed within each run and then averaged. The construction follows \citet{dieng-2020-topic}. Values are means~$\pm$~standard deviations (in smaller type) over 5 independent runs. The sentence encoder is MiniLM. \textbf{Bold}/\underline{underline} indicate the best/second-best score within each dataset--topic-count--category row.}
\label{tab:topic_quality}
\footnotesize
\colcellreset
\colcellfit{\ms{0.345}{0.018}}
\colcellfit{\underline{\ms{0.430}{0.010}}}
\colcellfit{\textbf{\ms{0.525}{0.026}}}
\colcellfit{LDA}
\colcellfit{SentLDA}
\colcellfit{SAM}
\colcellfit{GLDA}
\colcellfit{vLDA}
\colcellfit{ETM}
\colcellfit{ConTM}
\colcellfit{GSLDA}
\colcellfit{\textsc{vSLDA}}
\colcellfit{\colcite[\tiny]{Blei, 2003}}
\colcellfit{\colcite[\tiny]{Balikas, 2016}}
\colcellfit{\colcite[\tiny]{Reisinger, 2010}}
\colcellfit{\colcite[\tiny]{Das, 2015}}
\colcellfit{\colcite[\tiny]{Li, 2016}}
\colcellfit{\colcite[\tiny]{Dieng, 2020}}
\colcellfit{\colcite[\tiny]{Bianchi, 2021}}
\colcellfit{\colcite[\tiny]{Cha, 2024}}
\colcellfit{(proposed)}
\colcellfit{{\scriptsize word-level}}
\colcellfit{{\scriptsize sentence-level}}
\colcellfit{{\scriptsize document-level}}
\adjustbox{max width=\textheight}{%
\begin{tabular}{c c l *{9}{>{\centering\arraybackslash}p{\colcellwidth}}}
\toprule
\multirow[b]{4}{*}{\rotatebox[origin=c]{90}{Dataset}} & \multirow[b]{4}{*}{\rotatebox[origin=c]{90}{\#Topics}} & \multirow[b]{4}{*}{Category} & LDA & SentLDA & SAM & GLDA & vLDA & ETM & ConTM & GSLDA & \textsc{vSLDA} \\
 & & & \colcite[\tiny]{Blei, 2003} & \colcite[\tiny]{Balikas, 2016} & \colcite[\tiny]{Reisinger, 2010} & \colcite[\tiny]{Das, 2015} & \colcite[\tiny]{Li, 2016} & \colcite[\tiny]{Dieng, 2020} & \colcite[\tiny]{Bianchi, 2021} & \colcite[\tiny]{Cha, 2024} & (proposed) \\
\cmidrule(lr){4-4} \cmidrule(lr){5-5} \cmidrule(lr){6-6} \cmidrule(lr){7-9} \cmidrule(lr){10-10} \cmidrule(lr){11-12}
 & & & \multicolumn{1}{c}{\scriptsize word-level} & \multicolumn{1}{c}{\scriptsize sentence-level} & \multicolumn{1}{c}{\scriptsize document-level} & \multicolumn{3}{c}{\scriptsize word-level} & \multicolumn{1}{c}{\scriptsize document-level} & \multicolumn{2}{c}{\scriptsize sentence-level} \\
\cmidrule(lr){4-6} \cmidrule(lr){7-12}
 & & & \multicolumn{3}{c}{\scriptsize vocabulary-based} & \multicolumn{6}{c}{\scriptsize embedding-based} \\
\midrule
\multirow{18}{*}{\rotatebox[origin=c]{90}{20news}} & \multirow{6}{*}{10} & Computer & \ms{0.345}{0.018} & \ms{0.363}{0.015} & \ms{0.323}{0.009} & \underline{\ms{0.430}{0.010}} & \ms{0.347}{0.044} & \textbf{\ms{0.525}{0.026}} & \ms{0.151}{0.017} & \ms{0.296}{0.032} & \ms{0.429}{0.013} \\
 &  & Ride & \ms{0.276}{0.006} & \ms{0.354}{0.035} & \ms{0.333}{0.004} & \ms{0.406}{0.009} & \ms{0.276}{0.024} & \underline{\ms{0.482}{0.015}} & \ms{0.157}{0.013} & \ms{0.249}{0.021} & \textbf{\ms{0.513}{0.017}} \\
 &  & Sports & \ms{0.288}{0.016} & \ms{0.315}{0.039} & \ms{0.368}{0.008} & \ms{0.394}{0.012} & \ms{0.357}{0.038} & \textbf{\ms{0.498}{0.030}} & \ms{0.177}{0.007} & \ms{0.253}{0.013} & \underline{\ms{0.417}{0.029}} \\
 &  & Science & \ms{0.436}{0.053} & \ms{0.485}{0.033} & \ms{0.348}{0.010} & \ms{0.402}{0.013} & \ms{0.241}{0.045} & \underline{\ms{0.572}{0.014}} & \ms{0.201}{0.022} & \ms{0.383}{0.038} & \textbf{\ms{0.615}{0.033}} \\
 &  & Religion & \ms{0.343}{0.009} & \ms{0.365}{0.027} & \ms{0.363}{0.009} & \ms{0.430}{0.007} & \ms{0.249}{0.054} & \underline{\ms{0.553}{0.016}} & \ms{0.155}{0.012} & \ms{0.276}{0.017} & \textbf{\ms{0.588}{0.022}} \\
 &  & Politics & \ms{0.416}{0.047} & \ms{0.378}{0.018} & \ms{0.363}{0.009} & \ms{0.397}{0.008} & \ms{0.300}{0.056} & \underline{\ms{0.515}{0.015}} & \ms{0.164}{0.023} & \ms{0.270}{0.024} & \textbf{\ms{0.556}{0.064}} \\
\cmidrule(lr){2-12}
 & \multirow{6}{*}{20} & Computer & \ms{0.297}{0.006} & \ms{0.285}{0.020} & \ms{0.318}{0.023} & \underline{\ms{0.402}{0.004}} & \ms{0.284}{0.032} & \textbf{\ms{0.505}{0.020}} & \ms{0.139}{0.010} & \ms{0.212}{0.012} & \ms{0.368}{0.015} \\
 &  & Ride & \ms{0.256}{0.011} & \ms{0.275}{0.016} & \ms{0.318}{0.005} & \ms{0.391}{0.006} & \ms{0.212}{0.025} & \underline{\ms{0.420}{0.023}} & \ms{0.149}{0.008} & \ms{0.241}{0.014} & \textbf{\ms{0.448}{0.030}} \\
 &  & Sports & \ms{0.266}{0.010} & \ms{0.301}{0.011} & \ms{0.327}{0.005} & \ms{0.370}{0.004} & \ms{0.253}{0.018} & \textbf{\ms{0.451}{0.033}} & \ms{0.162}{0.005} & \ms{0.232}{0.008} & \underline{\ms{0.380}{0.010}} \\
 &  & Science & \ms{0.353}{0.014} & \ms{0.407}{0.022} & \ms{0.346}{0.014} & \ms{0.382}{0.005} & \ms{0.249}{0.029} & \underline{\ms{0.519}{0.005}} & \ms{0.184}{0.011} & \ms{0.287}{0.026} & \textbf{\ms{0.529}{0.031}} \\
 &  & Religion & \ms{0.308}{0.023} & \ms{0.319}{0.013} & \ms{0.349}{0.006} & \ms{0.415}{0.007} & \ms{0.218}{0.016} & \underline{\ms{0.502}{0.013}} & \ms{0.143}{0.012} & \ms{0.241}{0.004} & \textbf{\ms{0.526}{0.013}} \\
 &  & Politics & \ms{0.322}{0.009} & \ms{0.346}{0.029} & \ms{0.345}{0.018} & \ms{0.382}{0.004} & \ms{0.233}{0.024} & \underline{\ms{0.477}{0.028}} & \ms{0.156}{0.023} & \ms{0.216}{0.010} & \textbf{\ms{0.494}{0.014}} \\
\cmidrule(lr){2-12}
 & \multirow{6}{*}{30} & Computer & \ms{0.276}{0.008} & \ms{0.251}{0.009} & \ms{0.315}{0.015} & \underline{\ms{0.394}{0.006}} & \ms{0.224}{0.032} & \textbf{\ms{0.497}{0.008}} & \ms{0.122}{0.014} & \ms{0.190}{0.009} & \ms{0.352}{0.017} \\
 &  & Ride & \ms{0.251}{0.016} & \ms{0.279}{0.014} & \ms{0.323}{0.010} & \ms{0.376}{0.006} & \ms{0.222}{0.020} & \underline{\ms{0.395}{0.022}} & \ms{0.140}{0.007} & \ms{0.237}{0.008} & \textbf{\ms{0.412}{0.015}} \\
 &  & Sports & \ms{0.271}{0.013} & \ms{0.283}{0.010} & \ms{0.309}{0.006} & \ms{0.359}{0.005} & \ms{0.249}{0.027} & \textbf{\ms{0.430}{0.010}} & \ms{0.132}{0.009} & \ms{0.230}{0.006} & \underline{\ms{0.375}{0.015}} \\
 &  & Science & \ms{0.337}{0.012} & \ms{0.338}{0.019} & \ms{0.314}{0.007} & \ms{0.376}{0.003} & \ms{0.242}{0.024} & \underline{\ms{0.480}{0.014}} & \ms{0.181}{0.012} & \ms{0.268}{0.012} & \textbf{\ms{0.482}{0.031}} \\
 &  & Religion & \ms{0.281}{0.012} & \ms{0.306}{0.016} & \ms{0.329}{0.006} & \ms{0.403}{0.006} & \ms{0.195}{0.032} & \underline{\ms{0.457}{0.026}} & \ms{0.128}{0.005} & \ms{0.221}{0.007} & \textbf{\ms{0.485}{0.003}} \\
 &  & Politics & \ms{0.321}{0.006} & \ms{0.293}{0.029} & \ms{0.370}{0.018} & \ms{0.370}{0.003} & \ms{0.179}{0.012} & \underline{\ms{0.451}{0.019}} & \ms{0.148}{0.004} & \ms{0.209}{0.012} & \textbf{\ms{0.471}{0.023}} \\
\midrule
\multirow{12}{*}{\rotatebox[origin=c]{90}{NYT}} & \multirow{4}{*}{10} & Arts & \ms{0.363}{0.026} & \underline{\ms{0.495}{0.044}} & \ms{0.391}{0.022} & \ms{0.424}{0.005} & \ms{0.266}{0.025} & \ms{0.458}{0.040} & \ms{0.228}{0.014} & \ms{0.249}{0.010} & \textbf{\ms{0.585}{0.029}} \\
 &  & Business & \ms{0.404}{0.035} & \ms{0.472}{0.030} & \underline{\ms{0.493}{0.029}} & \ms{0.399}{0.007} & \ms{0.217}{0.040} & \ms{0.386}{0.029} & \ms{0.276}{0.038} & \ms{0.361}{0.017} & \textbf{\ms{0.564}{0.043}} \\
 &  & Politics & \ms{0.430}{0.042} & \ms{0.496}{0.023} & \underline{\ms{0.505}{0.023}} & \ms{0.397}{0.009} & \ms{0.244}{0.044} & \ms{0.400}{0.044} & \ms{0.315}{0.029} & \ms{0.305}{0.023} & \textbf{\ms{0.647}{0.040}} \\
 &  & Sports & \ms{0.619}{0.030} & \textbf{\ms{0.650}{0.034}} & \ms{0.577}{0.034} & \ms{0.421}{0.006} & \ms{0.364}{0.052} & \ms{0.581}{0.012} & \ms{0.360}{0.027} & \ms{0.498}{0.053} & \underline{\ms{0.635}{0.033}} \\
\cmidrule(lr){2-12}
 & \multirow{4}{*}{20} & Arts & \ms{0.320}{0.014} & \underline{\ms{0.442}{0.020}} & \ms{0.377}{0.015} & \ms{0.397}{0.008} & \ms{0.253}{0.028} & \ms{0.370}{0.026} & \ms{0.223}{0.009} & \ms{0.121}{0.010} & \textbf{\ms{0.493}{0.018}} \\
 &  & Business & \ms{0.318}{0.041} & \underline{\ms{0.428}{0.009}} & \ms{0.412}{0.029} & \ms{0.385}{0.005} & \ms{0.207}{0.014} & \ms{0.351}{0.038} & \ms{0.229}{0.023} & \ms{0.306}{0.020} & \textbf{\ms{0.552}{0.033}} \\
 &  & Politics & \ms{0.349}{0.020} & \underline{\ms{0.460}{0.036}} & \ms{0.412}{0.012} & \ms{0.387}{0.005} & \ms{0.188}{0.020} & \ms{0.341}{0.037} & \ms{0.281}{0.021} & \ms{0.274}{0.008} & \textbf{\ms{0.571}{0.027}} \\
 &  & Sports & \ms{0.549}{0.047} & \underline{\ms{0.589}{0.036}} & \ms{0.525}{0.022} & \ms{0.388}{0.007} & \ms{0.228}{0.049} & \ms{0.587}{0.010} & \ms{0.361}{0.020} & \ms{0.445}{0.016} & \textbf{\ms{0.599}{0.026}} \\
\cmidrule(lr){2-12}
 & \multirow{4}{*}{30} & Arts & \ms{0.267}{0.024} & \ms{0.370}{0.021} & \ms{0.226}{0.032} & \underline{\ms{0.387}{0.004}} & \ms{0.206}{0.028} & \ms{0.335}{0.023} & \ms{0.214}{0.014} & \ms{0.080}{0.005} & \textbf{\ms{0.457}{0.020}} \\
 &  & Business & \ms{0.291}{0.036} & \underline{\ms{0.403}{0.017}} & \ms{0.172}{0.009} & \ms{0.377}{0.003} & \ms{0.190}{0.029} & \ms{0.284}{0.005} & \ms{0.190}{0.017} & \ms{0.271}{0.018} & \textbf{\ms{0.509}{0.009}} \\
 &  & Politics & \ms{0.329}{0.019} & \underline{\ms{0.415}{0.023}} & \ms{0.215}{0.035} & \ms{0.376}{0.003} & \ms{0.187}{0.029} & \ms{0.285}{0.012} & \ms{0.246}{0.017} & \ms{0.236}{0.011} & \textbf{\ms{0.523}{0.023}} \\
 &  & Sports & \ms{0.500}{0.038} & \ms{0.501}{0.008} & \ms{0.477}{0.013} & \ms{0.376}{0.005} & \ms{0.212}{0.006} & \textbf{\ms{0.572}{0.010}} & \ms{0.333}{0.011} & \ms{0.372}{0.031} & \underline{\ms{0.546}{0.010}} \\
\bottomrule
\end{tabular}
}
\end{sidewaystable}

\FloatBarrier

\section{Per-Category Sample-Efficiency Results}
\label{app:sample_eff}
Figures~\ref{fig:sample_eff_20news_minilm_logreg} (20 Newsgroups, two pages) and \ref{fig:sample_eff_nyt_minilm_logreg} (NYT) report the per-category sample-efficiency curves at $K=10$, $20$ and $30$ under the protocol of Section~\ref{subsec:sample_eff} (MiniLM sentence embeddings, logistic regression classifier).
On 20 Newsgroups, vSLDA attains the highest mean accuracy at every labeled fraction in every panel at all three topic counts.
At $K=20$, on \textit{computer}, \textit{sports}, \textit{science} and \textit{politics}, ETM loses between $12.5$ and $17.6$ accuracy points as the labeled fraction falls from 100 to 5\%, whereas vSLDA loses between $3.8$ and $6.6$ and SentLDA between $2.7$ and $4.6$. On \textit{ride} and \textit{religion}, ETM stays far below the leading models and loses about as much as they do, and at $K=10$ and $K=30$ the accuracy of ETM again declines more steeply than that of the sentence-level models.
The NYT results at the smallest fractions are less uniform. On \textit{arts}, vSLDA is the most accurate model at every fraction at all three topic counts. On \textit{business} and \textit{politics}, SentLDA is above vSLDA at 5\% of the labels at all three topic counts and close to or above it up to 20\% at $K=20$ and $K=30$, and on \textit{sports} LDA remains among the leading methods at all fractions and all three topic counts.

Because each topic model is trained without supervision on the full training corpus, Section~\ref{subsec:sample_eff} and these figures measure how effectively a linear classifier trained on few labeled documents can exploit each fixed representation, not the ability of the topic model to learn from few labels.
On NYT, the share of its full-supervision accuracy that vSLDA retains with 5\% of the labels is lowest on the units with the fewest documents per fine category, \textit{politics} and \textit{business} (Table~\ref{tab:dataset_stats}), so the accuracy there reflects the variance of the classifier as much as the representation. That SentLDA is ahead on these two units at the smallest fractions (pooled over $K$, 5--10\% of the labels on \textit{business}, 5--20\% on \textit{politics}) may follow from the documents of vSLDA mixing more topics than those of SentLDA on NYT (\appref{app:entropy_thresholds}), since a representation spread over more topics offers a linear classifier more directions to fit from few examples.

\begin{figure}[!htbp]
\centering
\includegraphics{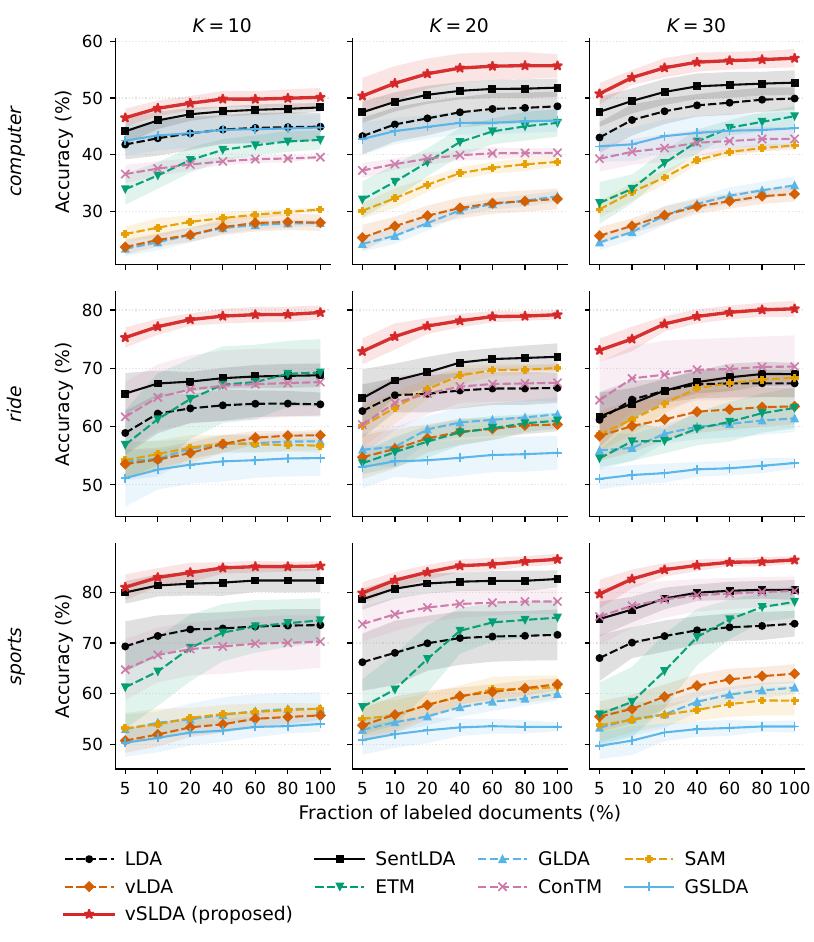}
\caption{\textbf{Sample efficiency per category on 20 Newsgroups (MiniLM).} Rows are the coarse categories \textit{computer}, \textit{ride} and \textit{sports} (the other three follow on the next page) and columns the topic counts $K=10$, $20$ and $30$. The vertical axis is the accuracy (\%) of fine-category classification by logistic regression on the document-topic distributions, and the horizontal axis is the fraction (\%) of training documents whose labels were used to train the classifier. The three panels of a row share the vertical range. Each point is the mean over 5 runs and, below 100\%, 5 labeled subsets drawn per run ($n=25$; $n=5$ at 100\%), shaded bands are $\pm1$ standard deviation over the same values (Section~\ref{subsec:metrics}), and the nine models of the main experiments are compared. Solid lines are the sentence-level models (SentLDA, GSLDA, vSLDA) and dashed lines the word- and document-level models.}
\label{fig:sample_eff_20news_minilm_logreg}
\end{figure}

\begin{figure}[!htbp]
\ContinuedFloat
\centering
\includegraphics{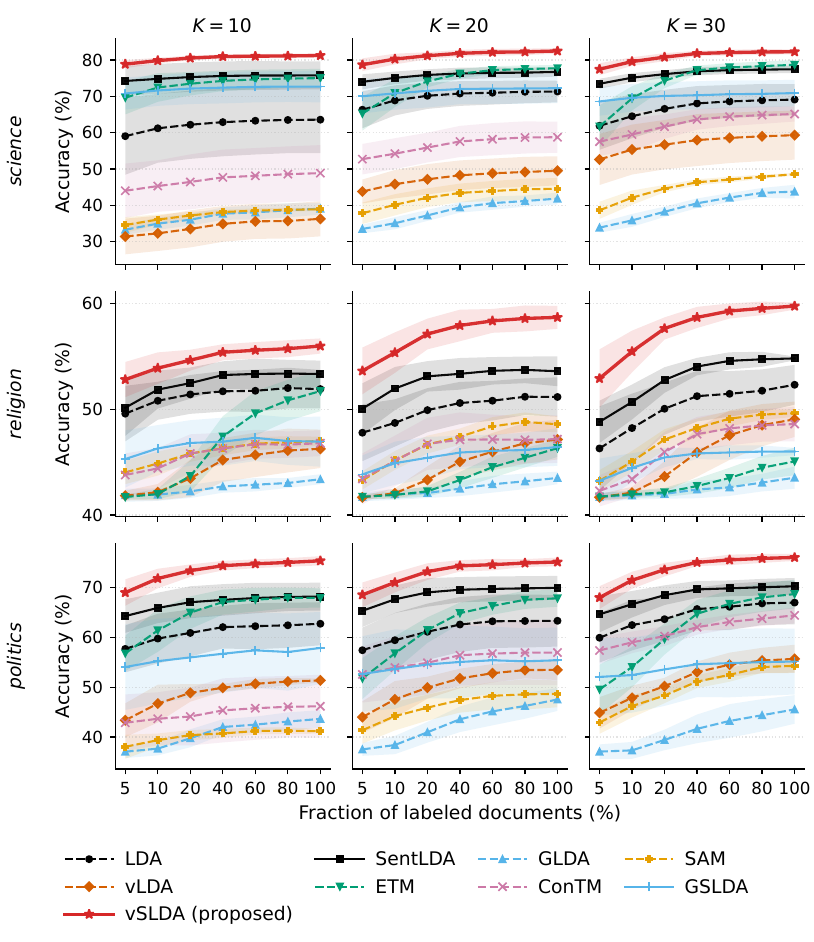}
\caption[]{\textbf{Sample efficiency per category on 20 Newsgroups (MiniLM), continued.} Rows are the coarse categories \textit{science}, \textit{religion} and \textit{politics} and columns the topic counts $K=10$, $20$ and $30$; axes, points, bands and line styles as on the previous page.}
\end{figure}

\begin{figure}[!htbp]
\centering
\includegraphics{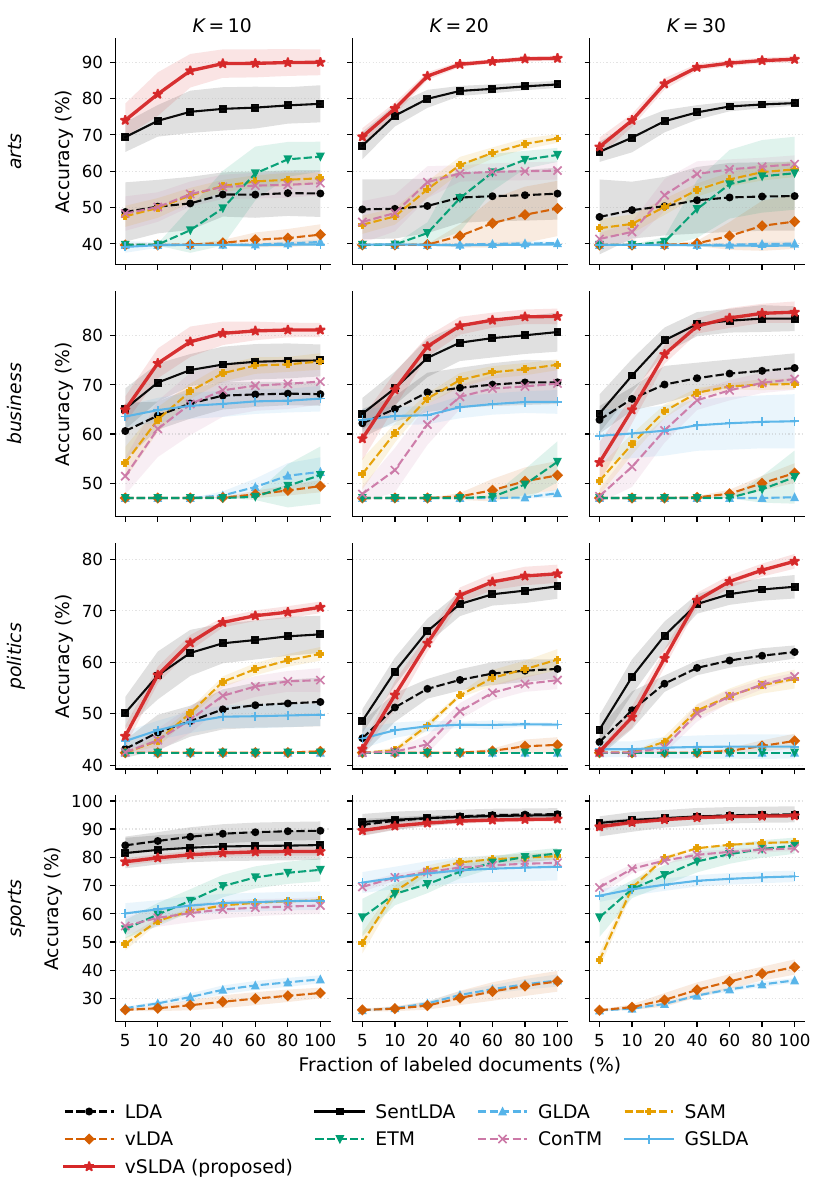}
\caption{\textbf{Sample efficiency per category on NYT (MiniLM).} Rows are the four coarse categories and columns the topic counts $K=10$, $20$ and $30$; axes, points, bands and line styles as in Figure~\ref{fig:sample_eff_20news_minilm_logreg}.}
\label{fig:sample_eff_nyt_minilm_logreg}
\end{figure}
\FloatBarrier

\section{Whole-Corpus Topic Diversity and the Per-Category Topic-Count Sweep}
\label{app:topic_sweep_categories}
\begin{figure}[!t]
\centering
\includegraphics{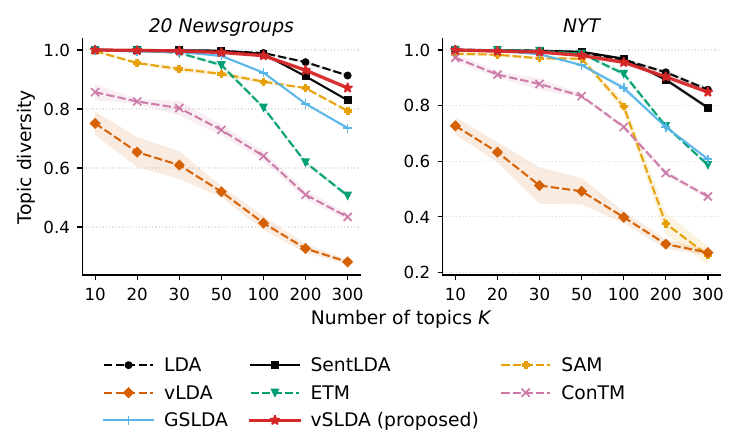}
\caption{\textbf{Topic diversity in the topic-count sweep on the whole corpus (MiniLM).} Left: 20 Newsgroups; right: NYT. The vertical axis is topic diversity, the horizontal axis the number of topics $K$; accuracy, $C_V$ coherence and topic quality on the same sweep are in Figure~\ref{fig:topic_sweep}. Each point is the mean over 5 runs ($n=5$), shaded bands are $\pm1$ standard deviation, and the eight models with runs across the whole grid are compared. Solid lines are the sentence-level models (SentLDA, GSLDA, vSLDA) and dashed lines the word- and document-level models.}
\label{fig:topic_sweep_diversity}
\end{figure}

\paragraph{Whole-corpus topic diversity}
Figure~\ref{fig:topic_sweep_diversity} shows the topic diversity of the whole-corpus sweep of Section~\ref{subsec:topic_sweep}, the one metric that Figure~\ref{fig:topic_sweep} omits; the other three are in that figure.
Topic diversity falls with $K$ for every model, but at different rates, and the decline of ETM is what confines its topic-quality lead in Section~\ref{subsec:topic_sweep} to a band of intermediate topic counts. ETM keeps its diversity at or above $0.98$ up to $K=30$ on both corpora and then loses it faster than LDA, SentLDA and vSLDA, falling to $0.805$ and $0.914$ at $K=100$ on 20 Newsgroups and NYT and to $0.506$ and $0.586$ at $K=300$. vSLDA stands at $0.981$ and $0.955$ at $K=100$ and at $0.871$ and $0.848$ at $K=300$. At $K=300$, LDA, SentLDA and vSLDA remain above $0.79$ on both corpora, and SAM stays at $0.793$ on 20 Newsgroups but falls to $0.261$ on NYT. The coherence of SAM falls from $0.396$ to $0.285$ on 20 Newsgroups and from $0.595$ to $0.322$ on NYT over the sweep, and its accuracy on NYT falls to $55.2$ at $K=300$ from its peak of $74.6$ at $K=50$. At $K=300$ on NYT, vSLDA, LDA and ETM lie within $0.030$ of one another on topic quality.

\begin{figure}[!t]
\centering
\includegraphics{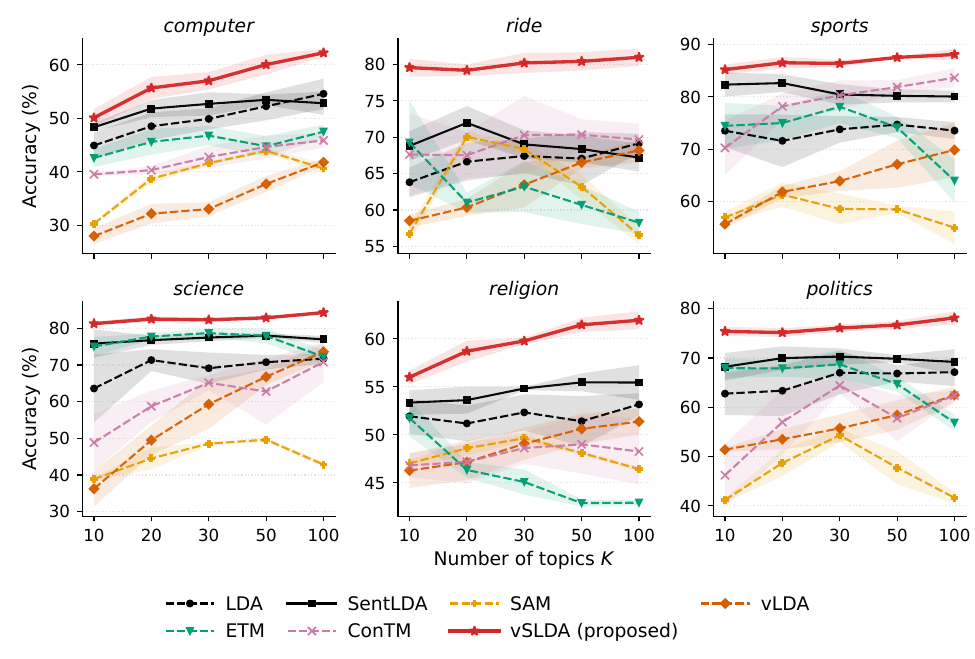}
\caption{\textbf{Topic-count sweep of classification accuracy within the coarse categories of 20 Newsgroups (MiniLM).} Each panel is one coarse category. The vertical axis is the accuracy (\%) of fine-category classification by logistic regression on the document-topic distributions, and the horizontal axis is the number of topics $K$ (the per-category runs stop at $K=100$). Each point is the mean over 5 runs ($n=5$), shaded bands are $\pm1$ standard deviation, and the seven models with runs across the grid are compared. Solid lines are the sentence-level models (SentLDA, vSLDA) and dashed lines the word- and document-level models.}
\label{fig:topic_sweep_cat_20news}
\end{figure}

\begin{figure}[!t]
\centering
\includegraphics{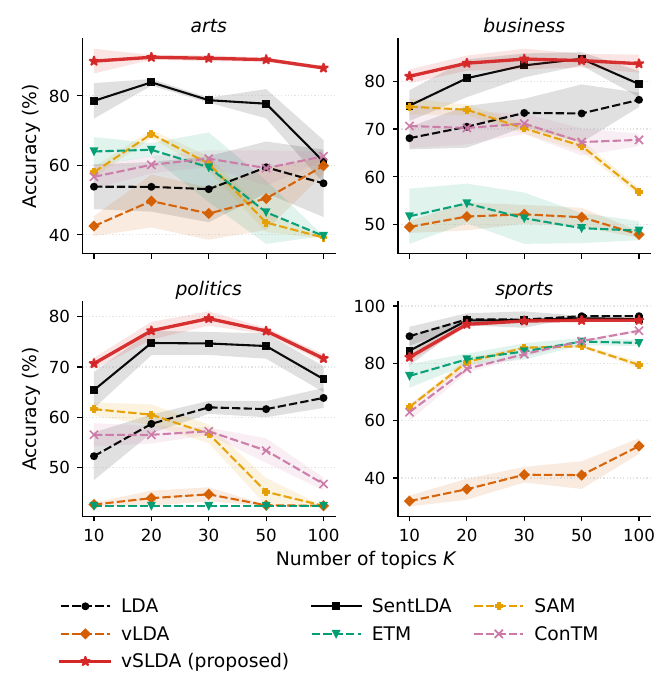}
\caption{\textbf{Topic-count sweep of classification accuracy within the coarse categories of NYT (MiniLM).} Each panel is one coarse category. The vertical axis is the accuracy (\%) of fine-category classification by logistic regression on the document-topic distributions, and the horizontal axis is the number of topics $K$ (the per-category runs stop at $K=100$). Each point is the mean over 5 runs ($n=5$), shaded bands are $\pm1$ standard deviation, and the seven models with runs across the grid are compared. Solid lines are the sentence-level models (SentLDA, vSLDA) and dashed lines the word- and document-level models.}
\label{fig:topic_sweep_cat_nyt}
\end{figure}

\begin{figure}[!t]
\centering
\includegraphics{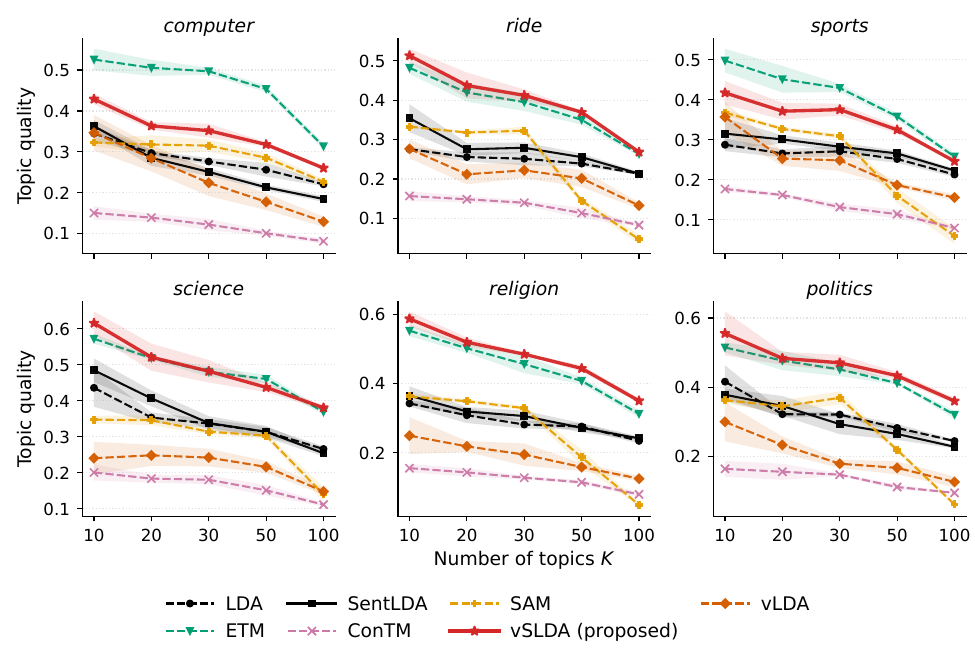}
\caption{\textbf{Topic-count sweep of topic quality within the coarse categories of 20 Newsgroups (MiniLM).} Each panel is one coarse category. The vertical axis is topic quality, the product of $C_V$ coherence and topic diversity (as in Table~\ref{tab:topic_quality}), and the horizontal axis is the number of topics $K$ (the per-category runs stop at $K=100$). Each point is the mean over 5 runs ($n=5$), shaded bands are $\pm1$ standard deviation, and the seven models with runs across the grid are compared. Solid lines are the sentence-level models (SentLDA, vSLDA) and dashed lines the word- and document-level models.}
\label{fig:topic_sweep_qual_20news}
\end{figure}

\begin{figure}[!t]
\centering
\includegraphics{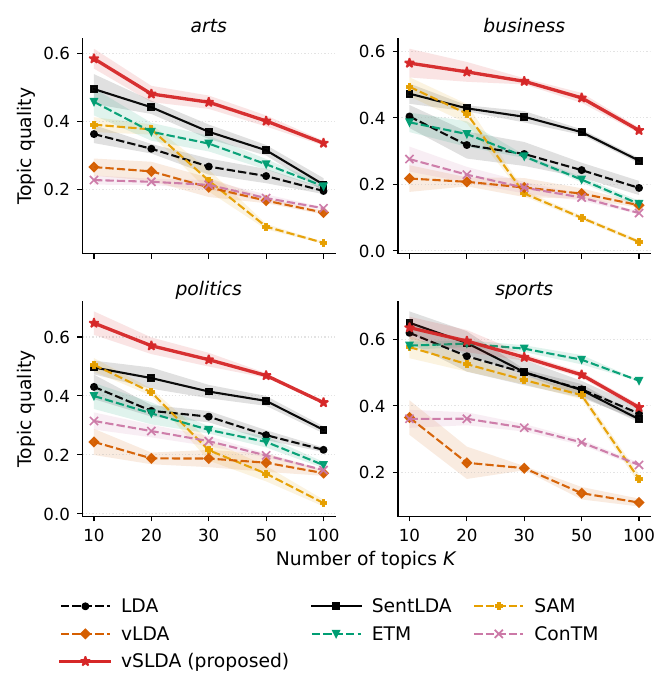}
\caption{\textbf{Topic-count sweep of topic quality within the coarse categories of NYT (MiniLM).} Each panel is one coarse category. The vertical axis is topic quality, the product of $C_V$ coherence and topic diversity (as in Table~\ref{tab:topic_quality}), and the horizontal axis is the number of topics $K$ (the per-category runs stop at $K=100$). Each point is the mean over 5 runs ($n=5$), shaded bands are $\pm1$ standard deviation, and the seven models with runs across the grid are compared. Solid lines are the sentence-level models (SentLDA, vSLDA) and dashed lines the word- and document-level models.}
\label{fig:topic_sweep_qual_nyt}
\end{figure}

Figures~\ref{fig:topic_sweep_cat_20news} and \ref{fig:topic_sweep_cat_nyt} report the same sweep on the coarse-category units of the main experiments, where the runs reach $K=100$, for classification accuracy, and Figures~\ref{fig:topic_sweep_qual_20news} and \ref{fig:topic_sweep_qual_nyt} for topic quality, under the protocols of Sections~\ref{subsec:cls_results} and \ref{subsec:coh_results}.
\paragraph{Classification}
Extending the grid beyond the topic counts of Table~\ref{tab:cls_logreg_minilm} keeps the ordering reported there. vSLDA attains the highest mean accuracy at $K=50$ on eight of the ten units and at $K=100$ on nine, including all six of 20 Newsgroups at both counts.
The exceptions are both on NYT.
On \textit{sports}, LDA is the highest at $K=50$ ($96.42$ against $94.97$) and at $K=100$ ($96.43$ against $94.98$), which continues the pattern discussed in Section~\ref{subsec:when_lda}. On \textit{business}, SentLDA is the highest at $K=50$ ($84.73$ against $84.38$), by less than the run-to-run standard deviation of either model, after vSLDA led at the three smaller topic counts and before it leads again at $K=100$.
The contrast with the whole-corpus unit of Section~\ref{subsec:topic_sweep}, where LDA and SentLDA lead at the smaller topic counts, is a property of the unit rather than of the topic count alone.
\paragraph{Topic coherence, diversity and topic quality}
The coherence side reproduces, within the coarse categories, the movement that Section~\ref{subsec:topic_sweep} reports on the corpus as a whole.
Under $C_V$ alone, the number of units on which vSLDA attains the highest score falls from 7 of 10 at $K=10$ to 6, 3, 2 and 0 as $K$ grows to 20, 30, 50 and 100, while ETM rises from 2 to 4, 7, 8 and 10 and SAM leads on none at any count. Of the three topic counts of Table~\ref{tab:topic_coherence_c_v}, the two smaller favor vSLDA by this measure and $K=30$ already reverses it.
Topic diversity again separates vSLDA from ETM. At $K=50$, it stays at or above $0.921$ for vSLDA and $0.913$ for SentLDA on every unit, whereas ETM falls as low as $0.498$, ConTM to $0.459$, vLDA to $0.420$ and SAM to $0.218$.
On their product, vSLDA leads on 7 units at $K=10$ and on 8, 7, 6 and 7 at $K=20$, $30$, $50$ and $100$, against 2, 2, 3, 4 and 3 for ETM, while SAM leads on none.
At $K=50$, vSLDA leads on topic quality on \textit{politics}, \textit{religion} and \textit{ride} of 20 Newsgroups and on \textit{arts}, \textit{business} and \textit{politics} of NYT, and ETM on the remaining three 20 Newsgroups units and on NYT \textit{sports}. The unit \textit{science} is the only one whose leader changes between $K=30$ and $K=50$, and it returns to vSLDA at $K=100$.

\FloatBarrier

\section{Results with a Nonlinear Classifier}
\label{app:svm}
Table~\ref{tab:cls_svm_minilm} repeats the full-data classification experiment of Table~\ref{tab:cls_logreg_minilm} with the downstream classifier replaced by a support vector machine with an RBF kernel, for which we use the scikit-learn implementation with its default settings.
The rest of the protocol is unchanged: the same MiniLM sentence embeddings, document-topic distributions $\bm\theta_d$ as features, training and test data and 5 independent runs.
The comparison is largely unaffected by this change. vSLDA attains the best accuracy in 27 of the 30 conditions, and the three exceptions are the same as in Table~\ref{tab:cls_logreg_minilm}: NYT \textit{sports} at $K=10$, $20$ and $30$, where LDA is ahead. It is ahead of the best baseline in all five paired runs in 18 conditions (20 with logistic regression), and at the unit level it is ahead of SentLDA on 9 of the 10 units, by $5.04$ points on average (median $4.54$; exact $p=0.004$), a repeat of the primary comparison of Section~\ref{subsec:cls_results} under another classifier rather than a further formal test.
The mean ranks over the 30 conditions are $1.17$ for vSLDA, $2.37$ for SentLDA, $3.90$ for LDA, $4.90$ for ETM, $5.13$ for ConTM, $5.37$ for SAM, $6.80$ for vLDA, $7.03$ for GSLDA and $8.33$ for GLDA. The ordering agrees with the one obtained with logistic regression except that two nearly tied pairs exchange places (vLDA and GSLDA, ConTM and ETM).
This indicates that the comparison reported in Section~\ref{subsec:cls_results} does not rest on the downstream classifier being a linear model.
\begin{sidewaystable}
\centering
\captionsetup{font=footnotesize}
\footnotesize
\setlength{\tabcolsep}{2pt}
\caption{Classification accuracy (\%) of a support vector machine with document-topic distributions as features. The protocol is that of Table~\ref{tab:cls_logreg_minilm} with the downstream classifier replaced. Values are means~$\pm$~standard deviations (in smaller type) over 5 independent runs. The sentence encoder is MiniLM. \textbf{Bold}/\underline{underline} indicate the best/second-best score within each dataset--topic-count--category row.}
\label{tab:cls_svm_minilm}
\footnotesize
\colcellreset
\colcellfit{\ms{45.82}{3.31}}
\colcellfit{\underline{\ms{47.59}{1.18}}}
\colcellfit{\textbf{\ms{50.08}{1.30}}}
\colcellfit{\ms{64.20}{10.25}}
\colcellfit{LDA}
\colcellfit{SentLDA}
\colcellfit{SAM}
\colcellfit{GLDA}
\colcellfit{vLDA}
\colcellfit{ETM}
\colcellfit{ConTM}
\colcellfit{GSLDA}
\colcellfit{\textsc{vSLDA}}
\colcellfit{\colcite[\tiny]{Blei, 2003}}
\colcellfit{\colcite[\tiny]{Balikas, 2016}}
\colcellfit{\colcite[\tiny]{Reisinger, 2010}}
\colcellfit{\colcite[\tiny]{Das, 2015}}
\colcellfit{\colcite[\tiny]{Li, 2016}}
\colcellfit{\colcite[\tiny]{Dieng, 2020}}
\colcellfit{\colcite[\tiny]{Bianchi, 2021}}
\colcellfit{\colcite[\tiny]{Cha, 2024}}
\colcellfit{(proposed)}
\colcellfit{{\scriptsize word-level}}
\colcellfit{{\scriptsize sentence-level}}
\colcellfit{{\scriptsize document-level}}
\adjustbox{max width=\textheight}{%
\begin{tabular}{c c l *{9}{>{\centering\arraybackslash}p{\colcellwidth}}}
\toprule
\multirow[b]{4}{*}{\rotatebox[origin=c]{90}{Dataset}} & \multirow[b]{4}{*}{\rotatebox[origin=c]{90}{\#Topics}} & \multirow[b]{4}{*}{Category} & LDA & SentLDA & SAM & GLDA & vLDA & ETM & ConTM & GSLDA & \textsc{vSLDA} \\
 & & & \colcite[\tiny]{Blei, 2003} & \colcite[\tiny]{Balikas, 2016} & \colcite[\tiny]{Reisinger, 2010} & \colcite[\tiny]{Das, 2015} & \colcite[\tiny]{Li, 2016} & \colcite[\tiny]{Dieng, 2020} & \colcite[\tiny]{Bianchi, 2021} & \colcite[\tiny]{Cha, 2024} & (proposed) \\
\cmidrule(lr){4-4} \cmidrule(lr){5-5} \cmidrule(lr){6-6} \cmidrule(lr){7-9} \cmidrule(lr){10-10} \cmidrule(lr){11-12}
 & & & \multicolumn{1}{c}{\scriptsize word-level} & \multicolumn{1}{c}{\scriptsize sentence-level} & \multicolumn{1}{c}{\scriptsize document-level} & \multicolumn{3}{c}{\scriptsize word-level} & \multicolumn{1}{c}{\scriptsize document-level} & \multicolumn{2}{c}{\scriptsize sentence-level} \\
\cmidrule(lr){4-6} \cmidrule(lr){7-12}
 & & & \multicolumn{3}{c}{\scriptsize vocabulary-based} & \multicolumn{6}{c}{\scriptsize embedding-based} \\
\midrule
\multirow{18}{*}{\rotatebox[origin=c]{90}{20news}} & \multirow{6}{*}{10} & Computer & \ms{45.82}{3.31} & \underline{\ms{47.59}{1.18}} & \ms{31.44}{0.29} & \ms{28.59}{0.87} & \ms{28.74}{1.68} & \ms{43.41}{2.01} & \ms{40.77}{0.47} & \ms{44.95}{2.22} & \textbf{\ms{50.08}{1.30}} \\
 &  & Ride & \ms{62.72}{2.51} & \ms{68.35}{1.96} & \ms{59.09}{1.19} & \ms{57.71}{1.88} & \ms{58.48}{1.02} & \underline{\ms{69.79}{5.32}} & \ms{68.08}{5.34} & \ms{54.93}{2.42} & \textbf{\ms{79.17}{1.45}} \\
 &  & Sports & \ms{72.88}{2.40} & \underline{\ms{81.82}{2.46}} & \ms{58.90}{0.99} & \ms{57.96}{2.77} & \ms{56.87}{0.35} & \ms{72.61}{5.15} & \ms{72.08}{4.79} & \ms{53.82}{1.41} & \textbf{\ms{85.06}{0.38}} \\
 &  & Science & \ms{64.20}{10.25} & \ms{75.62}{3.96} & \ms{39.71}{0.30} & \ms{39.67}{2.03} & \ms{37.82}{4.73} & \underline{\ms{76.10}{2.16}} & \ms{50.57}{6.93} & \ms{72.54}{4.04} & \textbf{\ms{81.00}{0.45}} \\
 &  & Religion & \ms{51.81}{1.58} & \ms{53.41}{1.59} & \ms{48.55}{0.40} & \ms{43.82}{0.86} & \ms{47.97}{1.84} & \underline{\ms{54.82}{0.82}} & \ms{46.82}{1.48} & \ms{47.25}{2.00} & \textbf{\ms{56.03}{0.41}} \\
 &  & Politics & \ms{62.49}{4.38} & \underline{\ms{67.68}{3.24}} & \ms{43.75}{0.52} & \ms{44.44}{1.64} & \ms{50.90}{0.80} & \ms{66.53}{2.20} & \ms{46.57}{5.38} & \ms{56.98}{7.23} & \textbf{\ms{74.98}{0.99}} \\
\cmidrule(lr){2-12}
 & \multirow{6}{*}{20} & Computer & \ms{48.37}{2.77} & \underline{\ms{51.16}{1.36}} & \ms{39.05}{1.25} & \ms{33.49}{0.80} & \ms{33.70}{2.30} & \ms{46.94}{2.41} & \ms{40.89}{1.93} & \ms{45.49}{1.97} & \textbf{\ms{55.67}{2.67}} \\
 &  & Ride & \ms{66.83}{1.47} & \ms{71.49}{1.58} & \underline{\ms{73.55}{1.06}} & \ms{63.12}{3.13} & \ms{59.22}{1.15} & \ms{62.13}{2.84} & \ms{69.63}{2.01} & \ms{54.56}{2.79} & \textbf{\ms{78.03}{0.72}} \\
 &  & Sports & \ms{71.38}{5.21} & \underline{\ms{82.22}{2.04}} & \ms{62.89}{1.59} & \ms{60.75}{1.83} & \ms{61.85}{1.93} & \ms{69.56}{5.71} & \ms{78.50}{2.75} & \ms{53.95}{1.52} & \textbf{\ms{85.25}{1.12}} \\
 &  & Science & \ms{71.78}{3.36} & \ms{76.59}{1.17} & \ms{46.16}{2.96} & \ms{42.36}{2.01} & \ms{52.34}{4.00} & \underline{\ms{78.06}{0.34}} & \ms{60.91}{4.11} & \ms{72.08}{4.66} & \textbf{\ms{81.81}{1.09}} \\
 &  & Religion & \ms{51.64}{1.03} & \underline{\ms{53.35}{1.04}} & \ms{49.89}{0.61} & \ms{45.59}{1.39} & \ms{49.19}{1.63} & \ms{51.26}{1.28} & \ms{47.14}{2.91} & \ms{45.99}{1.27} & \textbf{\ms{58.62}{0.73}} \\
 &  & Politics & \ms{64.36}{5.38} & \underline{\ms{69.81}{2.34}} & \ms{52.22}{3.35} & \ms{49.34}{1.67} & \ms{55.06}{2.57} & \ms{63.81}{2.60} & \ms{58.56}{6.41} & \ms{54.65}{7.30} & \textbf{\ms{74.19}{0.93}} \\
\cmidrule(lr){2-12}
 & \multirow{6}{*}{30} & Computer & \ms{50.12}{1.90} & \underline{\ms{52.07}{2.35}} & \ms{42.95}{1.10} & \ms{35.65}{1.92} & \ms{35.31}{1.64} & \ms{48.20}{2.47} & \ms{43.95}{2.59} & \ms{44.31}{2.50} & \textbf{\ms{56.63}{1.56}} \\
 &  & Ride & \ms{67.36}{2.35} & \ms{68.35}{1.78} & \underline{\ms{70.88}{1.23}} & \ms{62.96}{1.96} & \ms{64.13}{2.18} & \ms{66.35}{5.36} & \ms{70.69}{5.00} & \ms{53.39}{0.58} & \textbf{\ms{80.21}{1.00}} \\
 &  & Sports & \ms{74.22}{2.82} & \ms{79.68}{1.59} & \ms{61.20}{1.83} & \ms{62.06}{1.58} & \ms{65.73}{1.90} & \ms{69.61}{2.55} & \underline{\ms{81.31}{2.51}} & \ms{53.34}{0.61} & \textbf{\ms{85.44}{1.42}} \\
 &  & Science & \ms{70.41}{4.25} & \ms{77.26}{0.87} & \ms{50.71}{0.55} & \ms{45.95}{2.32} & \ms{61.60}{4.99} & \underline{\ms{78.02}{1.36}} & \ms{67.20}{1.39} & \ms{70.98}{3.85} & \textbf{\ms{81.66}{0.79}} \\
 &  & Religion & \ms{52.37}{2.84} & \underline{\ms{54.03}{1.01}} & \ms{50.85}{0.47} & \ms{45.80}{1.82} & \ms{51.32}{1.86} & \ms{49.32}{2.17} & \ms{48.74}{2.11} & \ms{45.52}{0.66} & \textbf{\ms{59.51}{0.86}} \\
 &  & Politics & \ms{68.23}{3.58} & \underline{\ms{70.23}{1.49}} & \ms{57.59}{1.81} & \ms{47.99}{3.56} & \ms{58.54}{3.27} & \ms{62.17}{2.12} & \ms{65.41}{1.80} & \ms{55.38}{6.14} & \textbf{\ms{75.66}{0.64}} \\
\midrule
\multirow{12}{*}{\rotatebox[origin=c]{90}{NYT}} & \multirow{4}{*}{10} & Arts & \ms{52.21}{9.74} & \underline{\ms{78.70}{6.58}} & \ms{61.11}{1.63} & \ms{46.10}{5.84} & \ms{50.43}{8.21} & \ms{62.40}{5.07} & \ms{57.26}{5.02} & \ms{39.66}{0.00} & \textbf{\ms{90.00}{2.71}} \\
 &  & Business & \ms{71.86}{3.12} & \ms{76.44}{4.18} & \underline{\ms{77.61}{2.52}} & \ms{58.93}{1.21} & \ms{64.89}{1.26} & \ms{64.37}{6.73} & \ms{71.04}{5.40} & \ms{69.41}{1.19} & \textbf{\ms{83.15}{1.95}} \\
 &  & Politics & \ms{56.97}{5.78} & \underline{\ms{67.22}{4.59}} & \ms{65.30}{0.69} & \ms{42.57}{0.30} & \ms{45.20}{3.40} & \ms{43.43}{1.05} & \ms{59.85}{2.62} & \ms{49.65}{3.30} & \textbf{\ms{71.92}{2.14}} \\
 &  & Sports & \textbf{\ms{90.73}{2.87}} & \underline{\ms{84.64}{5.06}} & \ms{67.54}{0.15} & \ms{48.09}{2.14} & \ms{42.72}{2.22} & \ms{81.93}{4.20} & \ms{64.99}{2.84} & \ms{66.78}{3.12} & \ms{83.26}{1.52} \\
\cmidrule(lr){2-12}
 & \multirow{4}{*}{20} & Arts & \ms{53.75}{5.08} & \underline{\ms{83.32}{1.11}} & \ms{72.12}{0.89} & \ms{45.67}{3.36} & \ms{63.46}{4.26} & \ms{62.69}{5.97} & \ms{61.88}{3.11} & \ms{39.66}{0.00} & \textbf{\ms{91.25}{0.39}} \\
 &  & Business & \ms{73.28}{6.30} & \underline{\ms{82.54}{2.50}} & \ms{80.20}{0.69} & \ms{61.93}{2.60} & \ms{71.45}{3.37} & \ms{65.60}{0.80} & \ms{75.62}{4.32} & \ms{68.29}{1.86} & \textbf{\ms{84.18}{1.36}} \\
 &  & Politics & \ms{64.74}{2.02} & \underline{\ms{76.06}{2.38}} & \ms{67.17}{2.48} & \ms{42.42}{0.00} & \ms{50.05}{5.62} & \ms{43.28}{0.74} & \ms{62.68}{3.57} & \ms{47.42}{0.46} & \textbf{\ms{80.20}{1.60}} \\
 &  & Sports & \textbf{\ms{95.96}{1.95}} & \underline{\ms{94.73}{0.64}} & \ms{84.23}{0.79} & \ms{56.57}{5.33} & \ms{55.23}{6.00} & \ms{89.18}{1.26} & \ms{81.95}{1.15} & \ms{77.87}{4.53} & \ms{94.01}{0.89} \\
\cmidrule(lr){2-12}
 & \multirow{4}{*}{30} & Arts & \ms{55.19}{8.36} & \underline{\ms{78.70}{2.30}} & \ms{66.06}{1.16} & \ms{47.45}{3.78} & \ms{65.86}{4.82} & \ms{58.60}{10.91} & \ms{67.55}{3.94} & \ms{39.66}{0.00} & \textbf{\ms{91.25}{0.64}} \\
 &  & Business & \ms{78.12}{3.36} & \underline{\ms{84.94}{1.92}} & \ms{74.76}{0.47} & \ms{61.02}{3.19} & \ms{76.54}{1.47} & \ms{62.80}{3.45} & \ms{77.46}{2.89} & \ms{63.36}{5.49} & \textbf{\ms{85.80}{1.79}} \\
 &  & Politics & \ms{68.89}{1.79} & \underline{\ms{77.78}{2.63}} & \ms{63.54}{3.09} & \ms{42.83}{0.69} & \ms{55.86}{1.97} & \ms{42.82}{0.81} & \ms{67.07}{1.78} & \ms{43.43}{2.02} & \textbf{\ms{82.83}{1.42}} \\
 &  & Sports & \textbf{\ms{95.66}{2.82}} & \ms{94.77}{0.43} & \ms{88.44}{0.50} & \ms{61.61}{2.25} & \ms{64.52}{2.63} & \ms{90.96}{1.49} & \ms{86.50}{0.72} & \ms{74.76}{2.13} & \underline{\ms{94.94}{0.59}} \\
\bottomrule
\end{tabular}
}
\end{sidewaystable}

\section{Coherence Measures and Results under \texorpdfstring{$C_{\mathrm{NPMI}}$}{C\_NPMI}}
\label{app:coherence_npmi}
This appendix defines the coherence measures of Section~\ref{subsec:metrics} and repeats the comparison of Table~\ref{tab:topic_coherence_c_v} under $C_{\mathrm{NPMI}}$.
\paragraph{Measures}
The normalized pointwise mutual information (NPMI) of two words $w_i$ and $w_j$ under a probability estimate $p$ is
\begin{equation}
  \mathrm{npmi}_{p}(w_i,w_j)
  \;=\;
  \frac{\log\bigl( (p(w_i,w_j)+\epsilon) \,/\, (p(w_i)\, p(w_j)) \bigr)}{-\log\bigl(p(w_i,w_j)+\epsilon\bigr)},
  \label{eq:npmi}
\end{equation}
where $\epsilon$ is a small constant. The representative words of Section~\ref{subsec:repwords} are ranked by Eq.~\eqref{eq:npmi} evaluated under the plug-in estimates of the expected counts, with the topic in place of one of the two words.
For $C_V$~\citep{roder-2015-coherence}, the probabilities $p_{\mathrm{sw}}$ of words and word pairs are estimated from boolean sliding windows of width $s$ over the reference corpus. Each representative word $w_{ki}$ of topic $k$ is mapped to the vector $\bm v_{ki}=\bigl(\mathrm{npmi}_{p_{\mathrm{sw}}}(w_{ki},w_{kj})\bigr)_{j=1}^{N}$ of its NPMI values (Eq.~\eqref{eq:npmi}) against the representative words, and, with $\bm v_k=\sum_{i=1}^{N}\bm v_{ki}$, the score of topic $k$ and of a model are
\begin{equation}
  C_V(k)
  \;=\;
  \frac{1}{N}\sum_{i=1}^{N}\cos\bigl(\bm v_{ki},\, \bm v_k\bigr),
  \qquad
  C_V=\frac{1}{K}\sum_{k=1}^{K} C_V(k).
  \label{eq:cv}
\end{equation}
Automated coherence measures are imperfect proxies for human judgment~\citep{hoyle-2021-broken}, so the comparison is repeated below with a second measure.
$C_{\mathrm{NPMI}}$ averages the NPMI of \citet{bouma-2009-npmi} over the pairs of representative words of a topic,
\begin{equation}
  C_{\mathrm{NPMI}}(k)
  \;=\;
  \frac{2}{N(N-1)}\sum_{i<j} \mathrm{npmi}_{p_{\mathrm{sw}}}(w_{ki},w_{kj}),
  \label{eq:cnpmi}
\end{equation}
rather than comparing their co-occurrence profiles as $C_V$ does in Eq.~\eqref{eq:cv}. It is one of the measures commonly reported alongside $C_V$.
\paragraph{Results under the NPMI-based measure}
Table~\ref{tab:topic_coherence_c_npmi} repeats the comparison of Table~\ref{tab:topic_coherence_c_v} under the $C_{\mathrm{NPMI}}$ of Eq.~\eqref{eq:cnpmi}, with only the sliding-window width changed (\appref{app:hyperparams}).
\begin{sidewaystable}
\centering
\captionsetup{font=footnotesize}
\footnotesize
\setlength{\tabcolsep}{2pt}
\caption{Topic coherence measured by $C_{\mathrm{NPMI}}$ using the top 10 topic words and English Wikipedia as the reference corpus. The protocol is that of Table~\ref{tab:topic_coherence_c_v} with the coherence measure replaced. Values are means~$\pm$~standard deviations (in smaller type) over 5 independent runs. The sentence encoder is MiniLM. \textbf{Bold}/\underline{underline} indicate the best/second-best score within each dataset--topic-count--category row.}
\label{tab:topic_coherence_c_npmi}
\footnotesize
\colcellreset
\colcellfit{\ms{-0.098}{0.030}}
\colcellfit{\underline{\ms{0.008}{0.003}}}
\colcellfit{\textbf{\ms{0.043}{0.012}}}
\colcellfit{\underline{\ms{-0.002}{0.004}}}
\colcellfit{\ms{0.001}{0.005}}
\colcellfit{\textbf{\ms{-0.004}{0.002}}}
\colcellfit{LDA}
\colcellfit{SentLDA}
\colcellfit{SAM}
\colcellfit{GLDA}
\colcellfit{vLDA}
\colcellfit{ETM}
\colcellfit{ConTM}
\colcellfit{GSLDA}
\colcellfit{\textsc{vSLDA}}
\colcellfit{\colcite[\tiny]{Blei, 2003}}
\colcellfit{\colcite[\tiny]{Balikas, 2016}}
\colcellfit{\colcite[\tiny]{Reisinger, 2010}}
\colcellfit{\colcite[\tiny]{Das, 2015}}
\colcellfit{\colcite[\tiny]{Li, 2016}}
\colcellfit{\colcite[\tiny]{Dieng, 2020}}
\colcellfit{\colcite[\tiny]{Bianchi, 2021}}
\colcellfit{\colcite[\tiny]{Cha, 2024}}
\colcellfit{(proposed)}
\colcellfit{{\scriptsize word-level}}
\colcellfit{{\scriptsize sentence-level}}
\colcellfit{{\scriptsize document-level}}
\adjustbox{max width=\textheight}{%
\begin{tabular}{c c l *{9}{>{\centering\arraybackslash}p{\colcellwidth}}}
\toprule
\multirow[b]{4}{*}{\rotatebox[origin=c]{90}{Dataset}} & \multirow[b]{4}{*}{\rotatebox[origin=c]{90}{\#Topics}} & \multirow[b]{4}{*}{Category} & LDA & SentLDA & SAM & GLDA & vLDA & ETM & ConTM & GSLDA & \textsc{vSLDA} \\
 & & & \colcite[\tiny]{Blei, 2003} & \colcite[\tiny]{Balikas, 2016} & \colcite[\tiny]{Reisinger, 2010} & \colcite[\tiny]{Das, 2015} & \colcite[\tiny]{Li, 2016} & \colcite[\tiny]{Dieng, 2020} & \colcite[\tiny]{Bianchi, 2021} & \colcite[\tiny]{Cha, 2024} & (proposed) \\
\cmidrule(lr){4-4} \cmidrule(lr){5-5} \cmidrule(lr){6-6} \cmidrule(lr){7-9} \cmidrule(lr){10-10} \cmidrule(lr){11-12}
 & & & \multicolumn{1}{c}{\scriptsize word-level} & \multicolumn{1}{c}{\scriptsize sentence-level} & \multicolumn{1}{c}{\scriptsize document-level} & \multicolumn{3}{c}{\scriptsize word-level} & \multicolumn{1}{c}{\scriptsize document-level} & \multicolumn{2}{c}{\scriptsize sentence-level} \\
\cmidrule(lr){4-6} \cmidrule(lr){7-12}
 & & & \multicolumn{3}{c}{\scriptsize vocabulary-based} & \multicolumn{6}{c}{\scriptsize embedding-based} \\
\midrule
\multirow{18}{*}{\rotatebox[origin=c]{90}{20news}} & \multirow{6}{*}{10} & Computer & \ms{-0.098}{0.030} & \ms{-0.048}{0.020} & \ms{-0.066}{0.018} & \underline{\ms{0.008}{0.003}} & \ms{-0.130}{0.077} & \textbf{\ms{0.043}{0.012}} & \ms{-0.294}{0.016} & \ms{-0.244}{0.046} & \ms{-0.063}{0.014} \\
 &  & Ride & \ms{-0.160}{0.013} & \ms{-0.110}{0.037} & \ms{-0.051}{0.014} & \underline{\ms{0.004}{0.003}} & \ms{-0.180}{0.005} & \textbf{\ms{0.023}{0.012}} & \ms{-0.290}{0.048} & \ms{-0.201}{0.025} & \ms{-0.019}{0.025} \\
 &  & Sports & \ms{-0.118}{0.020} & \ms{-0.090}{0.026} & \ms{-0.021}{0.006} & \underline{\ms{-0.002}{0.004}} & \ms{-0.121}{0.077} & \textbf{\ms{0.024}{0.012}} & \ms{-0.234}{0.012} & \ms{-0.161}{0.021} & \ms{-0.041}{0.023} \\
 &  & Science & \ms{-0.045}{0.033} & \ms{-0.029}{0.025} & \ms{-0.045}{0.014} & \ms{0.001}{0.005} & \ms{-0.195}{0.018} & \underline{\ms{0.064}{0.005}} & \ms{-0.215}{0.042} & \ms{-0.123}{0.035} & \textbf{\ms{0.069}{0.023}} \\
 &  & Religion & \ms{-0.057}{0.011} & \ms{-0.045}{0.009} & \ms{-0.040}{0.002} & \ms{0.017}{0.004} & \ms{-0.268}{0.060} & \underline{\ms{0.061}{0.010}} & \ms{-0.226}{0.048} & \ms{-0.154}{0.021} & \textbf{\ms{0.062}{0.019}} \\
 &  & Politics & \ms{-0.042}{0.041} & \ms{-0.051}{0.020} & \ms{-0.037}{0.009} & \ms{-0.000}{0.002} & \ms{-0.164}{0.096} & \textbf{\ms{0.031}{0.006}} & \ms{-0.190}{0.021} & \ms{-0.135}{0.021} & \underline{\ms{0.026}{0.061}} \\
\cmidrule(lr){2-12}
 & \multirow{6}{*}{20} & Computer & \ms{-0.126}{0.013} & \ms{-0.145}{0.032} & \ms{-0.072}{0.043} & \underline{\ms{-0.001}{0.002}} & \ms{-0.057}{0.047} & \textbf{\ms{0.038}{0.010}} & \ms{-0.302}{0.032} & \ms{-0.331}{0.032} & \ms{-0.136}{0.018} \\
 &  & Ride & \ms{-0.133}{0.016} & \ms{-0.175}{0.013} & \ms{-0.073}{0.008} & \underline{\ms{-0.000}{0.002}} & \ms{-0.095}{0.006} & \textbf{\ms{0.017}{0.014}} & \ms{-0.251}{0.018} & \ms{-0.181}{0.018} & \ms{-0.039}{0.015} \\
 &  & Sports & \ms{-0.119}{0.017} & \ms{-0.109}{0.007} & \ms{-0.057}{0.010} & \underline{\ms{-0.009}{0.001}} & \ms{-0.069}{0.003} & \textbf{\ms{0.028}{0.013}} & \ms{-0.192}{0.021} & \ms{-0.133}{0.008} & \ms{-0.053}{0.015} \\
 &  & Science & \ms{-0.086}{0.017} & \ms{-0.072}{0.022} & \ms{-0.040}{0.010} & \ms{-0.008}{0.002} & \ms{-0.104}{0.011} & \textbf{\ms{0.044}{0.003}} & \ms{-0.195}{0.013} & \ms{-0.197}{0.036} & \underline{\ms{-0.008}{0.026}} \\
 &  & Religion & \ms{-0.079}{0.014} & \ms{-0.107}{0.010} & \ms{-0.040}{0.005} & \ms{0.010}{0.003} & \ms{-0.154}{0.027} & \textbf{\ms{0.045}{0.003}} & \ms{-0.202}{0.018} & \ms{-0.179}{0.012} & \underline{\ms{0.016}{0.004}} \\
 &  & Politics & \ms{-0.102}{0.010} & \ms{-0.083}{0.035} & \ms{-0.044}{0.016} & \underline{\ms{-0.005}{0.001}} & \ms{-0.126}{0.032} & \textbf{\ms{0.026}{0.009}} & \ms{-0.183}{0.026} & \ms{-0.201}{0.022} & \ms{-0.013}{0.017} \\
\cmidrule(lr){2-12}
 & \multirow{6}{*}{30} & Computer & \ms{-0.157}{0.009} & \ms{-0.211}{0.012} & \ms{-0.071}{0.009} & \underline{\ms{-0.006}{0.002}} & \ms{-0.069}{0.028} & \textbf{\ms{0.043}{0.003}} & \ms{-0.296}{0.032} & \ms{-0.346}{0.014} & \ms{-0.150}{0.024} \\
 &  & Ride & \ms{-0.131}{0.011} & \ms{-0.150}{0.009} & \ms{-0.067}{0.012} & \underline{\ms{-0.007}{0.002}} & \ms{-0.059}{0.005} & \textbf{\ms{0.020}{0.010}} & \ms{-0.236}{0.022} & \ms{-0.145}{0.006} & \ms{-0.045}{0.018} \\
 &  & Sports & \ms{-0.107}{0.008} & \ms{-0.107}{0.009} & \ms{-0.066}{0.008} & \underline{\ms{-0.013}{0.003}} & \ms{-0.033}{0.003} & \textbf{\ms{0.040}{0.008}} & \ms{-0.207}{0.017} & \ms{-0.119}{0.003} & \ms{-0.054}{0.008} \\
 &  & Science & \ms{-0.097}{0.016} & \ms{-0.125}{0.021} & \ms{-0.071}{0.006} & \underline{\ms{-0.009}{0.002}} & \ms{-0.079}{0.010} & \textbf{\ms{0.034}{0.005}} & \ms{-0.198}{0.027} & \ms{-0.187}{0.014} & \ms{-0.039}{0.017} \\
 &  & Religion & \ms{-0.108}{0.018} & \ms{-0.108}{0.008} & \ms{-0.058}{0.009} & \underline{\ms{0.006}{0.002}} & \ms{-0.121}{0.022} & \textbf{\ms{0.036}{0.011}} & \ms{-0.162}{0.012} & \ms{-0.175}{0.007} & \ms{-0.000}{0.005} \\
 &  & Politics & \ms{-0.105}{0.010} & \ms{-0.141}{0.016} & \ms{-0.035}{0.014} & \underline{\ms{-0.009}{0.001}} & \ms{-0.118}{0.005} & \textbf{\ms{0.026}{0.011}} & \ms{-0.181}{0.008} & \ms{-0.236}{0.010} & \ms{-0.016}{0.005} \\
\midrule
\multirow{12}{*}{\rotatebox[origin=c]{90}{NYT}} & \multirow{4}{*}{10} & Arts & \ms{-0.083}{0.030} & \ms{0.003}{0.032} & \ms{-0.036}{0.010} & \underline{\ms{0.006}{0.003}} & \ms{-0.207}{0.040} & \ms{0.004}{0.026} & \ms{-0.161}{0.019} & \ms{-0.085}{0.015} & \textbf{\ms{0.054}{0.019}} \\
 &  & Business & \ms{-0.010}{0.020} & \ms{0.011}{0.022} & \underline{\ms{0.039}{0.016}} & \ms{0.003}{0.002} & \ms{-0.270}{0.043} & \ms{-0.020}{0.012} & \ms{-0.095}{0.018} & \ms{-0.053}{0.008} & \textbf{\ms{0.056}{0.019}} \\
 &  & Politics & \ms{-0.007}{0.027} & \ms{0.035}{0.014} & \underline{\ms{0.044}{0.013}} & \ms{-0.003}{0.003} & \ms{-0.229}{0.046} & \ms{-0.018}{0.018} & \ms{-0.070}{0.018} & \ms{-0.102}{0.019} & \textbf{\ms{0.116}{0.019}} \\
 &  & Sports & \underline{\ms{0.088}{0.016}} & \textbf{\ms{0.105}{0.029}} & \ms{0.080}{0.018} & \ms{0.006}{0.002} & \ms{-0.085}{0.032} & \ms{0.076}{0.009} & \ms{-0.057}{0.011} & \ms{0.025}{0.033} & \ms{0.085}{0.029} \\
\cmidrule(lr){2-12}
 & \multirow{4}{*}{20} & Arts & \ms{-0.094}{0.006} & \ms{-0.026}{0.013} & \ms{-0.053}{0.011} & \underline{\ms{-0.003}{0.004}} & \ms{-0.163}{0.068} & \ms{-0.009}{0.016} & \ms{-0.139}{0.008} & \ms{-0.103}{0.006} & \textbf{\ms{0.001}{0.012}} \\
 &  & Business & \ms{-0.057}{0.020} & \ms{-0.011}{0.009} & \ms{-0.014}{0.011} & \underline{\ms{-0.003}{0.002}} & \ms{-0.214}{0.038} & \ms{-0.018}{0.018} & \ms{-0.108}{0.013} & \ms{-0.067}{0.010} & \textbf{\ms{0.058}{0.023}} \\
 &  & Politics & \ms{-0.050}{0.006} & \underline{\ms{0.003}{0.016}} & \ms{-0.013}{0.011} & \ms{-0.006}{0.002} & \ms{-0.286}{0.037} & \ms{-0.014}{0.019} & \ms{-0.069}{0.016} & \ms{-0.106}{0.002} & \textbf{\ms{0.061}{0.023}} \\
 &  & Sports & \ms{0.044}{0.025} & \ms{0.072}{0.021} & \ms{0.058}{0.013} & \ms{-0.003}{0.002} & \ms{-0.181}{0.085} & \textbf{\ms{0.081}{0.006}} & \ms{-0.025}{0.011} & \ms{-0.018}{0.008} & \underline{\ms{0.079}{0.017}} \\
\cmidrule(lr){2-12}
 & \multirow{4}{*}{30} & Arts & \ms{-0.112}{0.011} & \ms{-0.055}{0.014} & \ms{-0.102}{0.012} & \textbf{\ms{-0.004}{0.002}} & \ms{-0.196}{0.053} & \underline{\ms{-0.014}{0.016}} & \ms{-0.133}{0.005} & \ms{-0.113}{0.010} & \ms{-0.017}{0.012} \\
 &  & Business & \ms{-0.065}{0.012} & \ms{-0.021}{0.011} & \ms{-0.091}{0.014} & \underline{\ms{-0.007}{0.000}} & \ms{-0.213}{0.066} & \ms{-0.023}{0.013} & \ms{-0.110}{0.025} & \ms{-0.058}{0.009} & \textbf{\ms{0.037}{0.007}} \\
 &  & Politics & \ms{-0.051}{0.010} & \ms{-0.026}{0.013} & \ms{-0.047}{0.007} & \underline{\ms{-0.009}{0.002}} & \ms{-0.246}{0.061} & \ms{-0.021}{0.009} & \ms{-0.073}{0.012} & \ms{-0.107}{0.004} & \textbf{\ms{0.030}{0.017}} \\
 &  & Sports & \ms{0.014}{0.021} & \ms{0.020}{0.005} & \ms{0.028}{0.008} & \ms{-0.009}{0.002} & \ms{-0.202}{0.044} & \textbf{\ms{0.075}{0.005}} & \ms{-0.045}{0.011} & \ms{-0.060}{0.017} & \underline{\ms{0.042}{0.012}} \\
\bottomrule
\end{tabular}
}
\end{sidewaystable}

As under $C_V$, vSLDA is higher than SentLDA and GSLDA on all ten experimental units, by $0.063$ and $0.146$ on average (exact $p=0.002$ against both).
The dataset dependence of Section~\ref{subsec:coherence_doclen} also persists. At the unit level, ETM is ahead on all six 20 Newsgroups units, by $0.064$ on average (exact $p=0.031$), and vSLDA on three of the four NYT units, by $0.042$ on average (exact $p=0.250$).

GLDA, mid-ranked under $C_V$, is the second-highest model in 18 of the 30 conditions under $C_{\mathrm{NPMI}}$ and the highest in one.
Its scores lie between $-0.013$ and $0.017$ in every condition, near the zero that Eq.~\eqref{eq:npmi} assigns to words co-occurring at the rate expected under independence. Its representative words are largely frequent general vocabulary, whose pairs co-occur at about that rate, whereas a pair of the rarer, more specific words of vSLDA that is never observed together in a reference window receives a strongly negative NPMI and lowers the average, so $C_{\mathrm{NPMI}}$ can rank a topic of the first kind above one of the second.

\section{Results with Additional Sentence Encoders}
\label{app:encoders}
This appendix repeats the experiments of Tables~\ref{tab:cls_logreg_minilm}, \ref{tab:topic_coherence_c_v} and \ref{tab:topic_quality} with the MiniLM encoder of Section~\ref{subsec:implementation} replaced by two further encoders. The rest of the protocol is unchanged. Both are widely used, openly licensed English sentence encoders.
\texttt{all-mpnet-base-v2} ($M=768$) keeps the fine-tuning objective, training pairs and mean pooling of MiniLM but uses the 12-layer MPNet backbone (about 110M parameters) pretrained by masked and permuted language modeling~\citep{song-2020-mpnet},\footnote{\url{https://huggingface.co/sentence-transformers/all-mpnet-base-v2}, revision \texttt{e8c3b32}.} so it varies backbone, capacity and dimensionality at a fixed training recipe.
\texttt{bge-base-en-v1.5} ($M=768$) matches MPNet in capacity and dimensionality but follows a different pipeline: a BERT-base backbone pretrained with a retrieval-oriented autoencoding objective (RetroMAE), fine-tuned in two contrastive stages on English text pairs, the second with a retrieval instruction attached to the query side, and pooled at the \texttt{[CLS]} token~\citep{xiao-2024-cpack},\footnote{\url{https://huggingface.co/BAAI/bge-base-en-v1.5}, revision \texttt{a5beb1e}.} so it varies training recipe and pooling at a fixed model size.
Following its model card, we use no instruction prefix for BGE.
All three are English-only encoder-only models with 23M to 110M parameters, so we do not claim robustness to instruction-tuned or decoder-based embedding models.

Table~\ref{tab:encoder_summary} aggregates the conditions of each dataset into a mean score and a mean rank, in the form of Table~\ref{tab:results_summary} with the topic counts pooled and the encoder varied instead.
Only ConTM, GSLDA and vSLDA depend on the sentence encoder, GSLDA and vSLDA through the embeddings of the individual sentences and ConTM through the embedding of the whole document.
\paragraph{Classification}
On classification, the table confirms the comparison of Section~\ref{subsec:cls_results} under both encoders, with the mean accuracy of vSLDA over the conditions of a dataset nearly unchanged by the substitution (Panel~a). One point is not visible in it.
Across the three encoders, vSLDA is not the highest in nine cases: six at $K=10$, the NYT \textit{sports} category at $K=20$ and $K=30$ under MiniLM (Section~\ref{subsec:when_lda}) and 20 Newsgroups/\textit{computer} at $K=30$ under BGE. The model ahead of it is LDA or SentLDA in every case.
Because MPNet and BGE vary different factors relative to MiniLM (above), this agreement indicates that the comparison is not an artifact of any one of them, within the class of encoder-only English sentence encoders considered here.
\paragraph{Topic coherence and diversity}
Under every encoder, the split between vSLDA and ETM of Section~\ref{subsec:coh_results} keeps its direction on coherence but not its size, and on topic quality it keeps its direction only on NYT.
On NYT, vSLDA attains the best mean rank on both measures under all three encoders, being the highest in 10, 9 and 7 of the 12 conditions on coherence and in 10, 10 and 8 of them on topic quality with MiniLM, MPNet and BGE, respectively.
On 20 Newsgroups the comparison moves towards ETM as the encoder is changed: ETM is the highest on coherence in 12, 17 and 18 of the 18 conditions under MiniLM, MPNet and BGE, and on topic quality, on which vSLDA is the highest in 12 of the 18 under MiniLM, ETM is the highest in 15 under MPNet and in all 18 under BGE.
ETM does not read sentence embeddings and so obtains the same scores under the three encoders, at a mean $C_V$ of $0.518$ on 20 Newsgroups, whereas the mean $C_V$ of vSLDA falls from MiniLM to MPNet to BGE on both datasets (Panel~b), on NYT nonetheless remaining the highest.
With the classification accuracy of vSLDA nearly unchanged by the same substitution, the coherence of its topics is the more sensitive of the two properties to the encoder.
GLDA and vLDA, which read word embeddings and so do not depend on the sentence encoder either, are also ahead of vSLDA in more of the 20 Newsgroups conditions as its coherence falls, which, together with the shift towards ETM, accounts for its lower mean rank there under MPNet and BGE in Panels~b and~c of Table~\ref{tab:encoder_summary}.

Among the models that read the sentence encoder, vSLDA is ahead of ConTM in all 30 conditions on both measures and of GSLDA on $C_V$ in all but one (20 Newsgroups \textit{sports} at $K=30$ under BGE), and it is ahead of SentLDA on $C_V$ in all 18 conditions of 20 Newsgroups and in 11, 11 and 8 of the 12 NYT conditions under MiniLM, MPNet and BGE.

The unit-level pattern of Section~\ref{subsec:coherence_doclen} shifts as a whole: the $C_V$ difference between vSLDA and ETM, averaged over the three topic counts, is lower under BGE than under MiniLM on every one of the ten experimental units while their ordering is largely preserved, and it is positive on six units under MiniLM, four under MPNet and two under BGE.

The structural diagnostics of \appref{app:entropy_thresholds} show the same pattern under the three encoders. Under MPNet and BGE, vSLDA has no diffuse topic in any run and between $2.6$ and $8.4$\% dead topics pooled over the three topic counts, against $3.8$ and $4.7$\% on 20 Newsgroups and NYT under MiniLM.
\begin{table}[!t]
\centering
\footnotesize
\setlength{\tabcolsep}{4pt}
\caption{Effect of the sentence encoder on (a) classification accuracy (\%), (b) topic coherence $C_V$ and (c) topic quality, in the layout of Table~\ref{tab:results_summary} with the topic counts pooled ($K \in \{10, 20, 30\}$) and the encoder varied. Only the three models that read embeddings from the sentence encoder are listed; the other six do not read them, so their scores do not depend on the encoder and are those of Tables~\ref{tab:cls_logreg_minilm}, \ref{tab:topic_coherence_c_v} and \ref{tab:topic_quality}. Each cell is the mean over the conditions of a dataset (18 for 20 Newsgroups, 12 for NYT) of the model's per-condition score, followed in parentheses by its mean rank among all nine compared models over the same conditions. The protocol is that of those three tables, panel by panel, with the sentence encoder replaced. The column headers give the embedding dimensionality $M$ of each encoder. Values are means over 5 independent runs. Comparisons are within a column, not across columns; scores carry no best/second-best mark, because the highest score of a column may belong to one of the six models not listed.}
\label{tab:encoder_summary}
\adjustbox{max width=\textwidth}{%
\begin{tabular}{@{}l *{3}{c}@{\hspace{7pt}} *{3}{c}@{}}
\toprule
\multirow[b]{3}{*}{Model} & \multicolumn{3}{c}{20news} & \multicolumn{3}{c}{NYT} \\
\cmidrule(lr){2-4} \cmidrule(lr){5-7}
 & \makecell{MiniLM\\{\scriptsize $M=384$}} & \makecell{MPNet\\{\scriptsize $M=768$}} & \makecell{BGE\\{\scriptsize $M=768$}} & \makecell{MiniLM\\{\scriptsize $M=384$}} & \makecell{MPNet\\{\scriptsize $M=768$}} & \makecell{BGE\\{\scriptsize $M=768$}} \\
 & \multicolumn{3}{c}{\scriptsize score (mean rank)} & \multicolumn{3}{c}{\scriptsize score (mean rank)} \\
\midrule
\multicolumn{7}{@{}l}{\textbf{(a) Classification accuracy}} \\
\addlinespace[1pt]
ConTM {\scriptsize(Bianchi, 2021)} & \sr{57.76}{5.11} & \sr{61.92}{4.11} & \sr{55.63}{5.33} & \sr{65.43}{4.83} & \sr{66.91}{4.50} & \sr{65.46}{4.67} \\
GSLDA {\scriptsize(Cha, 2024)} & \sr{54.66}{6.50} & \sr{50.89}{7.28} & \sr{39.01}{9.00} & \sr{55.93}{7.00} & \sr{49.86}{7.50} & \sr{45.02}{8.25} \\
\textsc{vSLDA} {\scriptsize(proposed)} & \sr{72.61}{1.00} & \sr{73.38}{1.06} & \sr{72.29}{1.11} & \sr{84.94}{1.50} & \sr{84.88}{1.17} & \sr{85.07}{1.17} \\
\midrule
\multicolumn{7}{@{}l}{\textbf{(b) Topic coherence}} \\
\addlinespace[1pt]
ConTM {\scriptsize(Bianchi, 2021)} & \sr{0.227}{8.89} & \sr{0.234}{8.78} & \sr{0.226}{8.67} & \sr{0.325}{8.08} & \sr{0.336}{7.67} & \sr{0.341}{7.50} \\
GSLDA {\scriptsize(Cha, 2024)} & \sr{0.259}{8.00} & \sr{0.256}{7.89} & \sr{0.247}{7.89} & \sr{0.348}{7.67} & \sr{0.330}{8.42} & \sr{0.303}{8.83} \\
\textsc{vSLDA} {\scriptsize(proposed)} & \sr{0.471}{2.17} & \sr{0.443}{2.61} & \sr{0.390}{3.33} & \sr{0.560}{1.17} & \sr{0.530}{1.25} & \sr{0.503}{1.83} \\
\midrule
\multicolumn{7}{@{}l}{\textbf{(c) Topic quality}} \\
\addlinespace[1pt]
ConTM {\scriptsize(Bianchi, 2021)} & \sr{0.155}{9.00} & \sr{0.164}{9.00} & \sr{0.151}{8.83} & \sr{0.271}{7.75} & \sr{0.280}{7.17} & \sr{0.281}{7.17} \\
GSLDA {\scriptsize(Cha, 2024)} & \sr{0.251}{7.44} & \sr{0.240}{7.44} & \sr{0.193}{8.06} & \sr{0.293}{7.33} & \sr{0.235}{8.17} & \sr{0.130}{8.67} \\
\textsc{vSLDA} {\scriptsize(proposed)} & \sr{0.469}{1.50} & \sr{0.441}{2.06} & \sr{0.388}{2.72} & \sr{0.557}{1.17} & \sr{0.526}{1.17} & \sr{0.499}{1.75} \\
\bottomrule
\end{tabular}}
\end{table}

\FloatBarrier

\section{Sensitivity of vSLDA to Its Hyperparameters}
\label{app:hyper_sensitivity}
This appendix examines how the results of vSLDA depend on seven settings other than the number of topics. Six are the settings whose values Section~\ref{subsec:implementation} fixes: the initial Dirichlet parameter $\alpha_0$, the number of Gibbs sweeps per E-step $\zeta$, the number of stored sweeps $B$, the number of outer iterations $T$, and the burn-in $T_0$ and step-size exponent $a$ of the stochastic approximation of Section~\ref{subsec:inference}. The seventh is the concentration $\kappa_0$ that Algorithm~\ref{alg:mcem} assigns to a topic holding at most one sentence.
Each setting is varied one at a time from the values of the main experiments ($\kappa_0=10$, $\alpha_0=50/K$, $\zeta=20$, $B=8$, $T=10$, $T_0=5$, $a=1$), with the other six held fixed, at $K=20$ with MiniLM embeddings on the coarse-category units of both datasets (six on 20 Newsgroups, four on NYT), with three runs per condition that differ only in the random seed. The main setting is read from the main runs with the same three seeds.

\begin{table}[!t]
\centering
\footnotesize
\setlength{\tabcolsep}{4pt}
\caption{Sensitivity of \textsc{vSLDA} to its own hyperparameters. Classification accuracy (\%, logistic regression), $C_V$ topic coherence, topic diversity and number of empty topics out of $K=20$ as one hyperparameter at a time is varied from the setting of the main experiments ($\kappa_0=10$, $\alpha_0=50/K=2.5$, $\zeta=20$, $B=8$, $T=10$, $T_0=5$, $a=1$; the row marked ``main'' in each block), at $K=20$ with MiniLM embeddings. $\kappa_0$ is the concentration assigned to a topic holding at most one sentence, $\alpha_0$ the initial Dirichlet parameter, $\zeta$ the number of Gibbs sweeps per E-step, $B$ the number of stored sweeps, $T$ the number of outer iterations, $T_0$ the burn-in after which the statistics are averaged and $\bm\alpha$ is held fixed ($T_0=T$: no averaging), and $a$ the exponent of the step sizes $\gamma_t=(t-T_0)^{-a}$. Values are means over the coarse-category units of each dataset and 3 runs per condition; the main setting is read from the main runs with the same seeds and therefore differs slightly from Table~\ref{tab:cls_logreg_minilm}. An empty topic is one to which no sentence is assigned at the last Gibbs sweep of training.}
\label{tab:hyper_sensitivity}
\begin{tabular}{@{}l *{4}{c} @{\hspace{7pt}} *{4}{c}@{}}
\toprule
\multirow[b]{2}{*}{Setting} & \multicolumn{4}{c}{20 Newsgroups} & \multicolumn{4}{c}{NYT} \\
\cmidrule(lr){2-5} \cmidrule(lr){6-9}
 & Acc.\ (\%) & $C_V$ & Div. & Empty & Acc.\ (\%) & $C_V$ & Div. & Empty \\
\midrule
\multicolumn{9}{@{}l}{\textbf{Fallback concentration $\kappa_0$}} \\
\addlinespace[1pt]
\quad $1$ & 73.50 & 0.456 & 0.996 & 0.0 & 86.35 & 0.553 & 0.995 & 0.0 \\
\quad $10$ (main) & 73.51 & 0.457 & 0.996 & 0.0 & 86.27 & 0.554 & 0.995 & 0.0 \\
\quad $100$ & 73.51 & 0.457 & 0.996 & 0.0 & 86.33 & 0.554 & 0.995 & 0.0 \\
\quad $1000$ & 73.51 & 0.457 & 0.996 & 0.0 & 86.33 & 0.554 & 0.995 & 0.0 \\
\midrule
\multicolumn{9}{@{}l}{\textbf{Initial Dirichlet parameter $\alpha_0$}} \\
\addlinespace[1pt]
\quad $0.1$ & 73.21 & 0.465 & 0.997 & 0.0 & 86.42 & 0.561 & 0.996 & 0.0 \\
\quad $1$ & 73.46 & 0.463 & 0.996 & 0.0 & 86.89 & 0.557 & 0.996 & 0.0 \\
\quad $50/K = 2.5$ (main) & 73.51 & 0.457 & 0.996 & 0.0 & 86.27 & 0.554 & 0.995 & 0.0 \\
\quad $10$ & 73.37 & 0.453 & 0.995 & 0.0 & 86.50 & 0.549 & 0.994 & 0.0 \\
\midrule
\multicolumn{9}{@{}l}{\textbf{Gibbs sweeps per E-step $\zeta$}} \\
\addlinespace[1pt]
\quad $8$ & 73.21 & 0.455 & 0.996 & 0.0 & 86.27 & 0.552 & 0.994 & 0.0 \\
\quad $10$ & 73.05 & 0.458 & 0.996 & 0.0 & 86.32 & 0.547 & 0.994 & 0.0 \\
\quad $20$ (main) & 73.51 & 0.457 & 0.996 & 0.0 & 86.27 & 0.554 & 0.995 & 0.0 \\
\quad $40$ & 73.51 & 0.450 & 0.996 & 0.0 & 86.09 & 0.560 & 0.994 & 0.0 \\
\quad $80$ & 73.28 & 0.451 & 0.997 & 0.0 & 87.04 & 0.552 & 0.995 & 0.0 \\
\midrule
\multicolumn{9}{@{}l}{\textbf{Stored sweeps $B$}} \\
\addlinespace[1pt]
\quad $1$ & 72.99 & 0.457 & 0.997 & 0.0 & 86.23 & 0.551 & 0.994 & 0.0 \\
\quad $2$ & 73.31 & 0.456 & 0.996 & 0.0 & 85.86 & 0.547 & 0.994 & 0.0 \\
\quad $4$ & 73.49 & 0.457 & 0.996 & 0.0 & 86.03 & 0.557 & 0.994 & 0.0 \\
\quad $8$ (main) & 73.51 & 0.457 & 0.996 & 0.0 & 86.27 & 0.554 & 0.995 & 0.0 \\
\quad $16$ & 73.39 & 0.458 & 0.996 & 0.0 & 86.59 & 0.554 & 0.995 & 0.0 \\
\quad $20$ & 73.37 & 0.456 & 0.996 & 0.0 & 86.35 & 0.559 & 0.994 & 0.0 \\
\midrule
\multicolumn{9}{@{}l}{\textbf{Outer iterations $T$}} \\
\addlinespace[1pt]
\quad $1$ & 70.42 & 0.358 & 0.989 & 0.0 & 77.72 & 0.407 & 0.991 & 0.0 \\
\quad $2$ & 71.73 & 0.394 & 0.995 & 0.0 & 81.84 & 0.460 & 0.993 & 0.0 \\
\quad $5$ & 73.37 & 0.436 & 0.997 & 0.0 & 85.09 & 0.534 & 0.994 & 0.0 \\
\quad $10$ (main) & 73.51 & 0.457 & 0.996 & 0.0 & 86.27 & 0.554 & 0.995 & 0.0 \\
\quad $20$ & 73.42 & 0.461 & 0.996 & 0.0 & 86.67 & 0.558 & 0.995 & 0.0 \\
\quad $30$ & 73.38 & 0.462 & 0.995 & 0.0 & 86.79 & 0.560 & 0.995 & 0.0 \\
\midrule
\multicolumn{9}{@{}l}{\textbf{Burn-in iterations $T_0$}} \\
\addlinespace[1pt]
\quad $1$ & 73.07 & 0.425 & 0.996 & 0.0 & 84.68 & 0.519 & 0.993 & 0.0 \\
\quad $2$ & 73.44 & 0.436 & 0.997 & 0.0 & 85.25 & 0.532 & 0.994 & 0.0 \\
\quad $3$ & 73.37 & 0.442 & 0.997 & 0.0 & 85.87 & 0.541 & 0.995 & 0.0 \\
\quad $5$ (main) & 73.51 & 0.457 & 0.996 & 0.0 & 86.27 & 0.554 & 0.995 & 0.0 \\
\quad $8$ & 73.50 & 0.464 & 0.995 & 0.0 & 86.92 & 0.560 & 0.995 & 0.0 \\
\quad $10$ & 73.53 & 0.465 & 0.996 & 0.0 & 86.92 & 0.556 & 0.995 & 0.0 \\
\midrule
\multicolumn{9}{@{}l}{\textbf{Step-size exponent $a$}} \\
\addlinespace[1pt]
\quad $0.6$ & 73.39 & 0.461 & 0.996 & 0.0 & 86.60 & 0.559 & 0.995 & 0.0 \\
\quad $0.7$ & 73.55 & 0.461 & 0.996 & 0.0 & 86.56 & 0.558 & 0.995 & 0.0 \\
\quad $0.8$ & 73.51 & 0.461 & 0.996 & 0.0 & 86.48 & 0.557 & 0.995 & 0.0 \\
\quad $0.9$ & 73.64 & 0.459 & 0.996 & 0.0 & 86.39 & 0.555 & 0.995 & 0.0 \\
\quad $1$ (main) & 73.51 & 0.457 & 0.996 & 0.0 & 86.27 & 0.554 & 0.995 & 0.0 \\
\bottomrule
\end{tabular}
\end{table}

The setting $B$ is varied at $\zeta=20$ and $\zeta$ at $B=8$, so that $B\le\zeta$ throughout.
$T_0=T$ is the Monte Carlo EM without averaging, the algorithm of \citet{li-2016-integrating} with the Newton solution for $\kappa_k$ and with $\bm\alpha$ re-estimated at every iteration, and the grid of $a$ covers the interval $(\tfrac12,1]$ on which the convergence statement of \appref{app:estimation} holds.
Table~\ref{tab:hyper_sensitivity} reports, for every point of the grid, the classification accuracy of logistic regression on the document-topic distributions, $C_V$ topic coherence, topic diversity and the number of empty topics, each averaged over the units of a dataset and the three runs.
Five of the seven settings leave the results essentially unchanged.
The $\kappa_0$ block is flat because the fallback never applied: no run of these experiments contained a topic holding at most one sentence.
That block therefore measures something else, namely how much the pooled accuracy moves between separate trainings of the same condition from the same seeds, which is $0.01$ points on 20 Newsgroups and $0.07$ on NYT.
Differences of that size elsewhere in the table cannot be read as sensitivity to the hyperparameter.
Varying $\alpha_0$ from $0.1$ to $10$ changes the accuracy by at most $0.30$ points on 20 Newsgroups and $0.61$ on NYT. The re-estimation of $\bm\alpha$ at every burn-in iteration (Eq.~\eqref{eq:alpha}) leaves the estimate nearly independent of the start.
The number of Gibbs sweeps per E-step $\zeta$, from $8$ to $80$, and the number of stored sweeps $B$, from $1$ to $20$, change the accuracy by at most $0.77$ and $0.52$ points on either dataset. The sufficient statistics of Eq.~\eqref{eq:suffstats} are averages over the sentences of a topic, of which there are hundreds to thousands on these units at $K=20$, so the Monte Carlo error that $\zeta$ and $B$ control is small beside the variation between runs.
Varying the step-size exponent $a$ over $(\tfrac12,1]$ changes the accuracy by at most $0.13$ points on 20 Newsgroups and $0.33$ on NYT, and $C_V$ by at most $0.005$: at this training length the shape of the averaging weights matters little once the burn-in is fixed.
Over these five settings no swept value changes the accuracy by more than $0.8$ points, against a run-to-run standard deviation of the same order in Table~\ref{tab:cls_logreg_minilm}. The topic measures move as little, with topic diversity above $0.99$ and no empty topic at every swept value.

Two settings matter, and only when they are small: the number of outer iterations $T$ and the burn-in $T_0$. Setting $T=1$ costs $3.1$ accuracy points on 20 Newsgroups and $8.6$ on NYT, $T=2$ recovers about half of that and $T=5$ most of it, and from $T=10$, the training length of the main experiments, on the accuracy is flat.
Topic coherence does not move past that point either, from $C_V=0.457$ at $T=10$ to $0.462$ at $T=30$ on 20 Newsgroups and from $0.554$ to $0.560$ on NYT.
This is consistent with the training log-likelihood traces of \appref{app:wallclock}, which are flat from about the fifth iteration onwards.
The burn-in acts through the same early iterations. With $T_0=1$, $\bm\alpha$ is frozen after a single outer iteration and the averaged statistics include the iterations that follow it, which costs $0.4$ accuracy points on 20 Newsgroups and $1.6$ on NYT and lowers $C_V$ to $0.425$ and $0.519$; the loss shrinks as $T_0$ grows and is within the run-to-run variation from $T_0=5$ on. At the other end, $T_0=T$ (no averaging) differs from the main setting by $0.0$ and $0.6$ points in accuracy and by at most $0.008$ in $C_V$ ($0.465$ and $0.556$), so at this training length the averaging that makes the convergence statement of \appref{app:estimation} available neither costs nor buys accuracy.

Because each setting is varied on its own, the table bounds the sensitivity along the seven axes through the setting of the main experiments, not over their joint space.

\section{Structural Diagnostics of the Document-Topic Distributions}
\label{app:entropy_thresholds}
\begin{figure}[!t]
\centering
\includegraphics{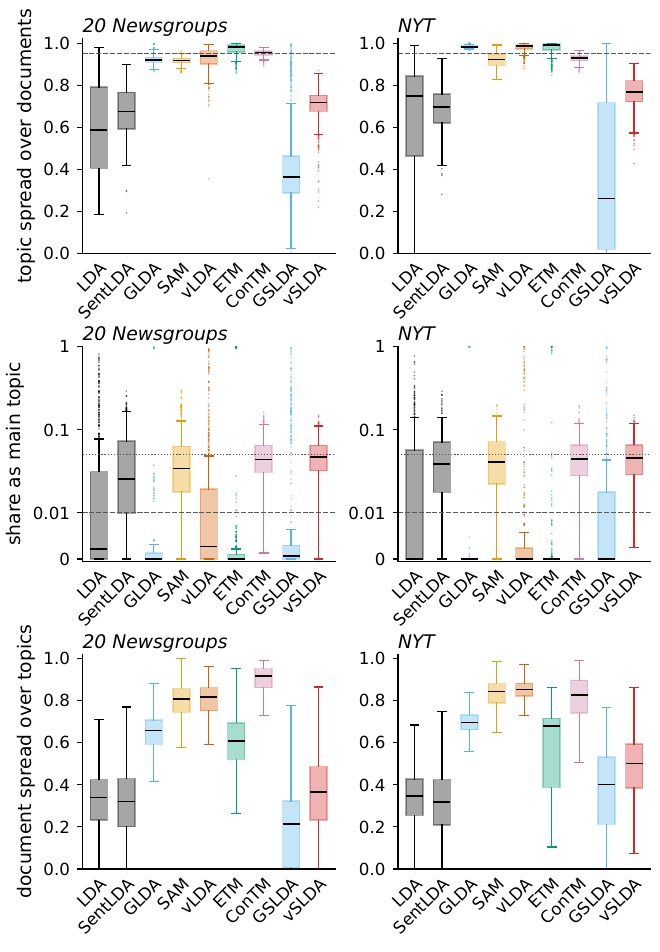}
\caption{\textbf{Per-topic and per-document distributions of the structural diagnostics ($K=20$, MiniLM).} Rows, from top: the normalized document entropy $H(P(d \mid k))/\ln D$ of each topic (``topic spread over documents''), the rank-1 document fraction of each topic (``share as main topic''; the axis is linear below $0.01$ and logarithmic above), and the normalized entropy $H(\bm\theta_d)/\ln K$ of each document (``document spread over topics''); columns: 20 Newsgroups (left) and NYT (right). The models are named on the horizontal axis. Each box pools the topics (or documents) of all coarse categories of the dataset and all 5 runs: boxes span the quartiles, whiskers extend to 1.5 interquartile ranges and points mark topics beyond the whiskers (not drawn in the bottom row). Dashed lines mark the cut-offs of Table~\ref{tab:entropy_summary}, $0.95$ for diffuse and $0.01$ for dead topics, and the dotted line the equal-use value $1/K$.}
\label{fig:entropy_boxplot}
\end{figure}

This appendix defines three structural diagnostics of the document-topic distributions, reports the percentages of diffuse and dead topics at the main and at other cut-offs, shows the underlying per-topic and per-document distributions and offers a reading of the contrasts.
\paragraph{Diagnostics}
A model can score well on classification and coherence while leaving part of its topics unused or letting one topic absorb a share of every document, and neither score reveals such a structure, so three diagnostics of the document-topic distributions of the $D$ test documents of a unit are added: (i) the normalized document entropy $H^{\mathrm{doc}}_k = H(P(d\mid k))/\ln D$ of topic $k$, where $P(d\mid k)=\theta_{dk}/\sum_{d'}\theta_{d'k}$ (a value of $x$ corresponds to a uniform spread over $D^x$ documents); (ii) its rank-1 document fraction $\rho_k$, the share of documents whose largest entry of $\bm\theta_d$ is $k$ ($1/K$ at equal use); and (iii) the normalized entropy $H^{\mathrm{top}}_d = H(\bm\theta_d)/\ln K$ of document $d$ (a mean of $m$ corresponds to $K^m$ effective topics).
The normalized document entropy equals $1-(\ln D - H(P(d \mid k)))/\ln D$, a rescaling of the divergence of $P(d\mid k)$ from the uniform distribution over documents by which \citet{alsumait-2009-significance} identify background topics; a value near $1$ identifies such a topic, present in every document, which we call \emph{diffuse}. The rank-1 document fraction follows the count of documents in which a topic is the single largest, one of the diagnostics of \citet{boyd-graber-2014-care}, and a large share of topics that are the main topic of almost no document, which we call \emph{dead}, is related to what \citet{srivastava-2017-avitm} describe as component collapsing, a local optimum in which the inferred topics become identical. We use these names because both diagnostics are read as properties of $\bm\theta$ alone, and because \emph{collapse} is also used for a document placing its mass on one topic.
\paragraph{Results}
\emph{vSLDA is the only compared model whose document-topic distributions keep both per-topic diagnostics low: no diffuse topic at any $K$ and at most $11.8$\% dead topics.}
The cut-offs are $H^{\mathrm{doc}}_k>0.95$ for a diffuse topic and $\rho_k<0.01$ for a dead one (a topic with no mass has $\rho_k=0$ and counts as dead).
Table~\ref{tab:entropy_summary} reports the diffuse and dead percentages and the mean document-topic entropy by dataset and number of topics, and Figure~\ref{fig:entropy_boxplot} shows the per-topic and per-document distributions at $K=20$.
\begin{table}[!t]
\centering
\footnotesize
\setlength{\tabcolsep}{4pt}
\caption{Structure of the document-topic distributions by dataset and number of topics. Panel~(a) is the percentage of the $K$ topics whose normalized document entropy $H(P(d \mid k))/\ln D$ exceeds $0.95$ (diffuse topics), panel~(b) the percentage whose rank-1 document fraction is below $0.01$ (dead topics) and panel~(c) the mean normalized entropy $H(\bm\theta_d)/\ln K$ of the document-topic distributions (a mean of $m$ corresponds to $K^m$ effective topics per document), all computed on the test documents from the document-topic distribution $\bm\theta_d$ of each model. Values are means~$\pm$~standard deviations over the categories of the dataset and the 5 independent runs of each (30 values per cell for 20 Newsgroups and 20 for NYT); the sentence encoder is MiniLM. In panels~(a) and (b), marked $\downarrow$, a lower value is preferable and \textbf{bold} indicates the lowest value within each column; panel~(c) has no preferred direction and is left unmarked. The cut-offs are conventions of this study (Table~\ref{tab:entropy_thresholds} for other cut-offs).}
\label{tab:entropy_summary}
\adjustbox{max width=\textwidth}{%
\begin{tabular}{@{}l *{3}{c}@{\hspace{7pt}} *{3}{c}@{}}
\toprule
\multirow[b]{2}{*}{Model} & \multicolumn{3}{c}{20news} & \multicolumn{3}{c}{NYT} \\
\cmidrule(lr){2-4} \cmidrule(lr){5-7}
 & $K=10$ & $K=20$ & $K=30$ & $K=10$ & $K=20$ & $K=30$ \\
\midrule
\multicolumn{7}{@{}l}{\textbf{(a) Diffuse topics: \% of the $K$ topics with $H(P(d \mid k))/\ln D > 0.95$ ($\downarrow$)}} \\
\addlinespace[1pt]
LDA {\scriptsize(Blei, 2003)} & \msp{\phantom{\textmd{0}}}{\phantom{\textmd{00}}8.0}{4.8} & \msp{\phantom{\textmd{0}}}{\phantom{\textmd{00}}2.5}{2.5} & \msp{\phantom{\textmd{0}}}{\phantom{\textmd{00}}1.2}{1.6} & \msp{\phantom{\textmd{0}}}{\phantom{\textmd{00}}7.0}{7.1} & \msp{\phantom{\textmd{0}}}{\phantom{\textmd{00}}3.5}{3.6} & \msp{\phantom{\textmd{0}}}{\phantom{\textmd{00}}2.3}{2.4} \\
SentLDA {\scriptsize(Balikas, 2016)} & \textbf{\msp{\phantom{\textmd{0}}}{\phantom{\textmd{00}}0.0}{0.0}} & \textbf{\msp{\phantom{\textmd{0}}}{\phantom{\textmd{00}}0.0}{0.0}} & \textbf{\msp{\phantom{\textmd{0}}}{\phantom{\textmd{00}}0.0}{0.0}} & \textbf{\msp{\phantom{\textmd{0}}}{\phantom{\textmd{00}}0.0}{0.0}} & \textbf{\msp{\phantom{\textmd{0}}}{\phantom{\textmd{00}}0.0}{0.0}} & \textbf{\msp{\phantom{\textmd{0}}}{\phantom{\textmd{00}}0.0}{0.0}} \\
SAM {\scriptsize(Reisinger, 2010)} & \ms{\phantom{\textmd{00}}6.3}{10.5} & \msp{\phantom{\textmd{0}}}{\phantom{\textmd{00}}0.5}{1.5} & \msp{\phantom{\textmd{0}}}{\phantom{\textmd{00}}8.9}{9.6} & \ms{\phantom{\textmd{0}}17.5}{30.5} & \ms{\phantom{\textmd{0}}24.2}{10.5} & \ms{\phantom{\textmd{0}}45.7}{27.1} \\
GLDA {\scriptsize(Das, 2015)} & \ms{\phantom{\textmd{0}}58.0}{23.3} & \msp{\phantom{\textmd{0}}}{\phantom{\textmd{0}}10.7}{1.7} & \msp{\phantom{\textmd{0}}}{\phantom{\textmd{00}}6.4}{0.8} & \msp{\phantom{\textmd{0}}}{100.0}{0.0} & \msp{\phantom{\textmd{0}}}{100.0}{0.0} & \msp{\phantom{\textmd{0}}}{\phantom{\textmd{0}}99.8}{0.7} \\
vLDA {\scriptsize(Li, 2016)} & \ms{\phantom{\textmd{0}}76.0}{11.1} & \msp{\phantom{\textmd{0}}}{\phantom{\textmd{0}}40.7}{9.4} & \msp{\phantom{\textmd{0}}}{\phantom{\textmd{0}}23.6}{6.9} & \msp{\phantom{\textmd{0}}}{\phantom{\textmd{0}}98.0}{5.1} & \msp{\phantom{\textmd{0}}}{\phantom{\textmd{0}}89.8}{6.0} & \msp{\phantom{\textmd{0}}}{\phantom{\textmd{0}}83.2}{7.8} \\
ETM {\scriptsize(Dieng, 2020)} & \ms{\phantom{\textmd{0}}78.0}{16.0} & \ms{\phantom{\textmd{0}}82.5}{10.9} & \msp{\phantom{\textmd{0}}}{\phantom{\textmd{0}}84.8}{9.0} & \msp{\phantom{\textmd{0}}}{\phantom{\textmd{0}}85.5}{8.6} & \ms{\phantom{\textmd{0}}81.7}{14.8} & \ms{\phantom{\textmd{0}}85.2}{14.4} \\
ConTM {\scriptsize(Bianchi, 2021)} & \ms{\phantom{\textmd{0}}74.0}{14.0} & \ms{\phantom{\textmd{0}}65.2}{11.5} & \ms{\phantom{\textmd{0}}56.7}{12.9} & \ms{\phantom{\textmd{0}}12.5}{15.5} & \msp{\phantom{\textmd{0}}}{\phantom{\textmd{00}}4.8}{9.0} & \msp{\phantom{\textmd{0}}}{\phantom{\textmd{00}}4.8}{7.3} \\
GSLDA {\scriptsize(Cha, 2024)} & \msp{\phantom{\textmd{0}}}{\phantom{\textmd{00}}5.0}{5.0} & \msp{\phantom{\textmd{0}}}{\phantom{\textmd{00}}2.8}{2.5} & \msp{\phantom{\textmd{0}}}{\phantom{\textmd{00}}2.0}{1.6} & \msp{\phantom{\textmd{0}}}{\phantom{\textmd{00}}8.0}{7.5} & \msp{\phantom{\textmd{0}}}{\phantom{\textmd{00}}4.3}{2.9} & \msp{\phantom{\textmd{0}}}{\phantom{\textmd{00}}2.7}{1.7} \\
\textsc{vSLDA} {\scriptsize(proposed)} & \textbf{\msp{\phantom{\textmd{0}}}{\phantom{\textmd{00}}0.0}{0.0}} & \textbf{\msp{\phantom{\textmd{0}}}{\phantom{\textmd{00}}0.0}{0.0}} & \textbf{\msp{\phantom{\textmd{0}}}{\phantom{\textmd{00}}0.0}{0.0}} & \textbf{\msp{\phantom{\textmd{0}}}{\phantom{\textmd{00}}0.0}{0.0}} & \textbf{\msp{\phantom{\textmd{0}}}{\phantom{\textmd{00}}0.0}{0.0}} & \textbf{\msp{\phantom{\textmd{0}}}{\phantom{\textmd{00}}0.0}{0.0}} \\
\midrule
\multicolumn{7}{@{}l}{\textbf{(b) Dead topics: \% of the $K$ topics with rank-1 document fraction $< 0.01$ ($\downarrow$)}} \\
\addlinespace[1pt]
LDA {\scriptsize(Blei, 2003)} & \ms{49.7}{13.3} & \msp{\phantom{\textmd{0}}}{65.7}{7.7} & \msp{\phantom{\textmd{0}}}{71.3}{7.1} & \ms{43.0}{23.3} & \ms{62.0}{16.5} & \ms{68.2}{15.0} \\
SentLDA {\scriptsize(Balikas, 2016)} & \msp{\phantom{\textmd{0}}}{\phantom{\textmd{0}}7.7}{6.2} & \msp{\phantom{\textmd{0}}}{25.5}{9.0} & \msp{\phantom{\textmd{0}}}{41.3}{5.1} & \msp{\phantom{\textmd{0}}}{\phantom{\textmd{0}}4.5}{8.0} & \ms{17.5}{16.8} & \ms{39.0}{17.7} \\
SAM {\scriptsize(Reisinger, 2010)} & \textbf{\msp{\phantom{\textmd{0}}}{\phantom{\textmd{0}}0.0}{0.0}} & \msp{\phantom{\textmd{0}}}{10.2}{5.8} & \msp{\phantom{\textmd{0}}}{39.4}{7.4} & \textbf{\msp{\phantom{\textmd{0}}}{\phantom{\textmd{0}}0.0}{0.0}} & \msp{\phantom{\textmd{0}}}{13.5}{8.5} & \ms{46.5}{22.9} \\
GLDA {\scriptsize(Das, 2015)} & \msp{\phantom{\textmd{0}}}{80.0}{0.0} & \msp{\phantom{\textmd{0}}}{90.0}{0.0} & \msp{\phantom{\textmd{0}}}{93.3}{0.0} & \msp{\phantom{\textmd{0}}}{88.5}{3.6} & \msp{\phantom{\textmd{0}}}{95.0}{0.0} & \msp{\phantom{\textmd{0}}}{96.7}{0.0} \\
vLDA {\scriptsize(Li, 2016)} & \ms{60.7}{12.4} & \ms{67.3}{11.2} & \msp{\phantom{\textmd{0}}}{70.1}{8.7} & \msp{\phantom{\textmd{0}}}{83.0}{7.8} & \ms{83.2}{10.2} & \msp{\phantom{\textmd{0}}}{79.3}{7.0} \\
ETM {\scriptsize(Dieng, 2020)} & \msp{\phantom{\textmd{0}}}{83.7}{8.4} & \msp{\phantom{\textmd{0}}}{93.7}{2.2} & \msp{\phantom{\textmd{0}}}{96.1}{1.2} & \ms{82.5}{13.4} & \msp{\phantom{\textmd{0}}}{91.5}{6.3} & \msp{\phantom{\textmd{0}}}{95.0}{3.1} \\
ConTM {\scriptsize(Bianchi, 2021)} & \msp{\phantom{\textmd{0}}}{\phantom{\textmd{0}}0.3}{1.8} & \msp{\phantom{\textmd{0}}}{\phantom{\textmd{0}}5.8}{4.1} & \msp{\phantom{\textmd{0}}}{\phantom{\textmd{0}}9.4}{4.1} & \textbf{\msp{\phantom{\textmd{0}}}{\phantom{\textmd{0}}0.0}{0.0}} & \msp{\phantom{\textmd{0}}}{\phantom{\textmd{0}}3.3}{3.6} & \textbf{\msp{\phantom{\textmd{0}}}{\phantom{\textmd{0}}9.7}{6.0}} \\
GSLDA {\scriptsize(Cha, 2024)} & \ms{60.0}{15.5} & \msp{\phantom{\textmd{0}}}{80.2}{8.7} & \msp{\phantom{\textmd{0}}}{88.4}{5.4} & \ms{52.0}{30.4} & \ms{72.3}{27.0} & \ms{82.7}{19.8} \\
\textsc{vSLDA} {\scriptsize(proposed)} & \msp{\phantom{\textmd{0}}}{\phantom{\textmd{0}}0.3}{1.8} & \textbf{\msp{\phantom{\textmd{0}}}{\phantom{\textmd{0}}4.3}{4.0}} & \textbf{\msp{\phantom{\textmd{0}}}{\phantom{\textmd{0}}6.8}{5.1}} & \textbf{\msp{\phantom{\textmd{0}}}{\phantom{\textmd{0}}0.0}{0.0}} & \textbf{\msp{\phantom{\textmd{0}}}{\phantom{\textmd{0}}2.2}{3.7}} & \msp{\phantom{\textmd{0}}}{11.8}{5.6} \\
\midrule
\multicolumn{7}{@{}l}{\textbf{(c) Mean normalized document-topic entropy $H(\bm\theta_d)/\ln K$ (no preferred direction)}} \\
\addlinespace[1pt]
LDA {\scriptsize(Blei, 2003)} & \ms{0.36}{0.06} & \ms{0.33}{0.04} & \ms{0.31}{0.04} & \ms{0.38}{0.06} & \ms{0.35}{0.05} & \ms{0.32}{0.06} \\
SentLDA {\scriptsize(Balikas, 2016)} & \ms{0.33}{0.04} & \ms{0.31}{0.03} & \ms{0.30}{0.03} & \ms{0.37}{0.06} & \ms{0.35}{0.05} & \ms{0.33}{0.03} \\
SAM {\scriptsize(Reisinger, 2010)} & \ms{0.78}{0.01} & \ms{0.79}{0.01} & \ms{0.78}{0.01} & \ms{0.78}{0.04} & \ms{0.80}{0.03} & \ms{0.83}{0.03} \\
GLDA {\scriptsize(Das, 2015)} & \ms{0.68}{0.01} & \ms{0.64}{0.01} & \ms{0.62}{0.02} & \ms{0.74}{0.04} & \ms{0.73}{0.03} & \ms{0.71}{0.04} \\
vLDA {\scriptsize(Li, 2016)} & \ms{0.80}{0.04} & \ms{0.79}{0.03} & \ms{0.77}{0.02} & \ms{0.83}{0.05} & \ms{0.86}{0.03} & \ms{0.87}{0.02} \\
ETM {\scriptsize(Dieng, 2020)} & \ms{0.60}{0.06} & \ms{0.59}{0.07} & \ms{0.58}{0.08} & \ms{0.42}{0.20} & \ms{0.38}{0.19} & \ms{0.36}{0.18} \\
ConTM {\scriptsize(Bianchi, 2021)} & \ms{0.87}{0.01} & \ms{0.89}{0.01} & \ms{0.90}{0.01} & \ms{0.82}{0.03} & \ms{0.84}{0.03} & \ms{0.85}{0.03} \\
GSLDA {\scriptsize(Cha, 2024)} & \ms{0.25}{0.07} & \ms{0.19}{0.06} & \ms{0.15}{0.05} & \ms{0.32}{0.13} & \ms{0.22}{0.14} & \ms{0.16}{0.15} \\
\textsc{vSLDA} {\scriptsize(proposed)} & \ms{0.38}{0.03} & \ms{0.35}{0.03} & \ms{0.34}{0.02} & \ms{0.49}{0.03} & \ms{0.49}{0.03} & \ms{0.48}{0.03} \\
\bottomrule
\end{tabular}}
\end{table}

On diffuse topics (panel~a of Table~\ref{tab:entropy_summary}), vSLDA and SentLDA differ from the other models. Neither has a diffuse topic in any of its 150 runs, whereas between 83 and 100\% of the topics of GLDA and vLDA on NYT are diffuse.
Dead topics (panel~b of the same table) separate vSLDA from SentLDA. At $K=30$, they make up $6.8$\% of the topics of vSLDA on 20 Newsgroups and $11.8$\% on NYT, against $41.3$ and $39.0$\% for SentLDA, so the absence of diffuse topics in SentLDA comes with a large share of unused topics. Of the other models, ConTM stays below $10$\% throughout, SAM reaches $39.4$ and $46.5$\% at $K=30$, and LDA, GLDA, vLDA, ETM and GSLDA leave between 43 and 97\% of their topics dead.
ETM is the only model whose topics are mostly diffuse and mostly dead in every condition: between 78 and 86\% of its topics are diffuse and between 83 and 96\% dead, the largest shares of both when pooled over the 150 runs ($82.7$ and $90.6$\%; Table~\ref{tab:entropy_thresholds}), and $41.1$\% exceed even the cut-off $0.99$. In every condition, therefore, most topics of ETM are spread almost uniformly over the test documents and yet are the main topic of almost none of them, and its coherence advantage on 20 Newsgroups (Section~\ref{subsec:coh_results}) is measured over such topics.
At $K=20$ (panel~c of the same table), vSLDA mixes $2.9$ effective topics per document on 20 Newsgroups and $4.4$ on NYT, a value close to those of SentLDA ($2.6$, $2.9$) and LDA ($2.7$, $2.8$). ConTM and SAM spread each document over $10.6$ to $14.2$ of the 20 topics, and GSLDA lies at the other end with $1.8$ and $1.9$, its distributions having largely collapsed onto single topics (Section~\ref{subsec:why_gslda}).
vSLDA keeps no diffuse and few dead topics under the MPNet and BGE encoders as well (\appref{app:encoders}).

Table~\ref{tab:entropy_thresholds} recomputes both percentages at three further cut-offs each, pooled over the 150 runs of the main experiments. At every cut-off, vSLDA and SentLDA have the lowest diffuse percentages and vSLDA and ConTM the lowest dead percentages. The largest dead cut-off, $0.05$, equals the equal-use value $1/K$ at $K=20$, and there every model lies between $48.7$ and $93.9$\%.
\begin{table}[!t]
\centering
\footnotesize
\setlength{\tabcolsep}{6pt}
\caption{Cut-off sensitivity of panels~(a) and (b) of Table~\ref{tab:entropy_summary}: the percentage of the $K$ topics of a run that are diffuse (normalized document entropy $H(P(d \mid k))/\ln D$ above the cut-off) and dead (rank-1 document fraction below the cut-off), computed per run and averaged with equal weight over the 150 runs of the main experiments (the 10 categories of the two datasets $\times$ $K \in \{10, 20, 30\}$ $\times$ 5 runs; MiniLM). The columns marked (main) are the cut-offs of Table~\ref{tab:entropy_summary}. A lower value is preferable ($\downarrow$) and \textbf{bold} indicates the lowest value within each column.}
\label{tab:entropy_thresholds}
\begin{tabular}{@{}l *{4}{c}@{\hspace{7pt}} *{4}{c}@{}}
\toprule
\multirow[b]{3}{*}{Model} & \multicolumn{4}{c}{Diffuse topics (\%, $\downarrow$)} & \multicolumn{4}{c}{Dead topics (\%, $\downarrow$)} \\
\cmidrule(lr){2-5} \cmidrule(lr){6-9}
 & \multicolumn{4}{c}{$H(P(d \mid k))/\ln D >$} & \multicolumn{4}{c}{rank-1 document fraction $<$} \\
 & $0.85$ & $0.90$ & $0.95$ (main) & $0.99$ & $0.001$ & $0.01$ (main) & $0.02$ & $0.05$ \\
\midrule
LDA {\scriptsize(Blei, 2003)} & 18.4 & 10.6 & 4.1 & 0.1 & 44.6 & 60.4 & 64.8 & 73.7 \\
SentLDA {\scriptsize(Balikas, 2016)} & \textbf{6.1} & 1.9 & \textbf{0.0} & \textbf{0.0} & 5.0 & 23.0 & 35.5 & 57.9 \\
SAM {\scriptsize(Reisinger, 2010)} & 99.4 & 82.7 & 14.8 & 5.6 & 7.6 & 17.9 & 27.7 & 54.7 \\
GLDA {\scriptsize(Das, 2015)} & 99.3 & 84.7 & 55.0 & 13.8 & 76.2 & 90.0 & 91.1 & 93.9 \\
vLDA {\scriptsize(Li, 2016)} & 94.2 & 85.8 & 64.2 & 16.8 & 48.5 & 72.4 & 78.4 & 86.5 \\
ETM {\scriptsize(Dieng, 2020)} & 99.8 & 97.0 & 82.7 & 41.1 & 76.1 & 90.6 & 92.5 & 93.5 \\
ConTM {\scriptsize(Bianchi, 2021)} & 99.9 & 97.3 & 42.1 & \textbf{0.0} & 0.3 & 4.8 & 14.0 & 51.2 \\
GSLDA {\scriptsize(Cha, 2024)} & 17.4 & 10.5 & 4.0 & 1.3 & 55.0 & 73.3 & 75.8 & 78.9 \\
\textsc{vSLDA} {\scriptsize(proposed)} & 8.1 & \textbf{1.8} & \textbf{0.0} & \textbf{0.0} & \textbf{0.2} & \textbf{4.2} & \textbf{12.3} & \textbf{48.7} \\
\bottomrule
\end{tabular}
\end{table}

\paragraph{Interpretation of the contrasts}
Of the compared models, only vSLDA keeps both per-topic diagnostics low: it produces no diffuse topic (a background topic in the sense of \citet{alsumait-2009-significance}) at any $K$ and leaves few topics dead. SentLDA and ConTM each match one half of this pattern: SentLDA has no diffuse topic but its share of dead topics grows steeply with $K$, and ConTM keeps few topics dead but the majority of its topics on 20 Newsgroups are diffuse.
One caveat applies: a $\bm\theta_d$ estimated from the sentence assignments of a document ($6$ to $26$ on average, depending on the experimental unit) rests on far fewer observations than one estimated from word tokens and is sharper for that reason alone, so part of the contrast with the word-level models may reflect the number of observations.
Between vSLDA and SentLDA, the share of dead topics grows steeply with $K$ only for SentLDA, which suggests that the vMF likelihood over sentence embeddings, unlike the categorical likelihood over the words of a sentence, continues to find sentences that an additional topic explains better as topics are added.
The accuracy of vSLDA rises at every step of the grid of Section~\ref{subsec:topic_sweep}, where that of LDA, SentLDA, SAM and ETM peaks and then declines, which is consistent with this reading.

\section{Observations per Dimension, Prior Scale and Covariance Structure of GSLDA}
\label{app:n_over_m}
This appendix defines the quantity $n/M$ of Section~\ref{subsec:why_gslda}, the number of observations per dimension from which each full-covariance Gaussian topic of GSLDA is estimated, and reports the prior-scale sweep and the covariance-structure comparison of that section.
\paragraph{Observations per dimension}
For each training run, $n$ is the median, over the topics with at least one observation, of the number of observations assigned to a topic at the last Gibbs sweep of training, the observations being word tokens for GLDA and sentences for GSLDA. $M$ is the dimension of the embeddings in which the topic covariances are estimated, $300$ for the word vectors of GLDA and $384$ for the MiniLM sentence embeddings of GSLDA, so that $n/M<1$ means that the sample covariance of a typical topic is singular. The counts are those of the training runs themselves.

\begin{figure}[!t]
\centering
\includegraphics{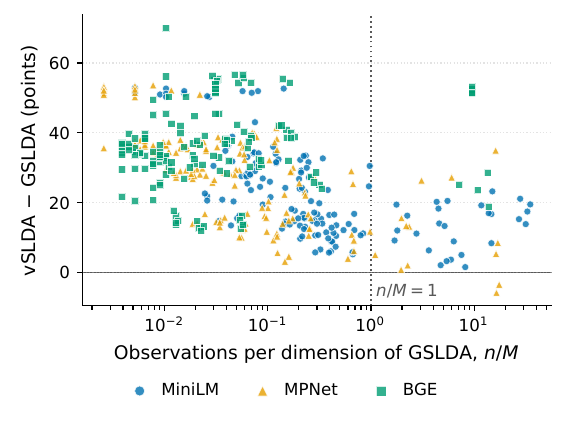}
\caption{\textbf{Accuracy gap between vSLDA and GSLDA against the observations per dimension of GSLDA.} Each point is one GSLDA training run (both datasets, every coarse category, $K\in\{10,20,30\}$, 5 runs, three sentence encoders; $450$ points, $150$ per encoder). The vertical axis is the classification accuracy of vSLDA minus that of GSLDA in the same condition (points), and the horizontal axis, on a logarithmic scale, is $n/M$, the median number of sentences assigned to a non-empty topic at the last Gibbs sweep divided by the embedding dimension $M$ ($384$ for MiniLM, $768$ for MPNet and BGE). The dotted vertical line marks $n/M=1$, below which the sample covariance of a typical topic is singular.}
\label{fig:n_over_m}
\end{figure}

Figure~\ref{fig:n_over_m} plots, for every GSLDA run, the classification accuracy of vSLDA minus that of GSLDA in the same condition against $n/M$, for the MiniLM runs of Table~\ref{tab:cls_logreg_minilm} and the MPNet and BGE runs of \appref{app:encoders} ($M=768$). The gap widens as $n/M$ falls, with a Spearman correlation of $\rho=-0.55$ between $n/M$ and the gap pooled over the three encoders, the value quoted in Section~\ref{subsec:why_gslda}.
Over the $450$ runs, the least-squares slope of the gap on $\log_{10}(n/M)$ is $-7.6$ points per decade, and the gap averages $34.3$ points where $n/M<0.1$, $14.2$ points where $0.3\le n/M<1$ and $10.8$ points where $1\le n/M<3$. Within one encoder the Spearman correlation is $-0.66$ under MiniLM and $-0.70$ under MPNet, but $0.14$ under BGE, whose runs lie mostly to the left of $n/M=0.1$ and so cover little of the range. The pooled value therefore partly reflects offsets between encoders.

The deficit is not confined to the low-$n/M$ regime. On NYT/\textit{sports} at $K=10$, the condition with the most observations per dimension among the 30 of Table~\ref{tab:cls_logreg_minilm} ($n/M=30.8$; median topic size $11{,}810$ sentences), no topic is empty and the largest holds $16.5\%$ of the sentences against $16.1\%$ under the isotropic refit, so the assignments show none of the concentration in a few topics that marks the collapse, yet GSLDA trails vSLDA by $17.5$ points. These runs do not identify why the full covariance still trails at $n/M\approx30$. \appref{app:size_bias} suggests that its predictive remains short of its large-sample limit at such sizes.
Table~\ref{tab:whole_corpus_gslda} repeats the comparison on each corpus as a whole, the unit of Figure~\ref{fig:topic_sweep}, where the classifier separates all fine categories of the corpus, with GSLDA fitted at $K\in\{10,20,30\}$ under the settings of the main experiments (Figure~\ref{fig:topic_sweep} extends the comparison to $K=300$). In these six conditions $n/M$ lies between $3.3$ and $41.1$, no run has an empty topic and the largest topic holds between $13.0\%$ and $16.9\%$ of the sentences. GSLDA trails vSLDA in all six, from $1.8$ points on 20 Newsgroups at $K=10$, less than the run-to-run standard deviation of either model there, to $18.9$ on NYT at $K=30$, and on both corpora the gap is largest at $K=30$.
\begin{table}[!t]
\centering
\footnotesize
\setlength{\tabcolsep}{5pt}
\caption{GSLDA on each corpus as a whole (MiniLM). Accuracy is that of fine-category classification by logistic regression on the document-topic distributions (\%, mean $\pm$ standard deviation over 5 runs), and $\Delta$ the mean accuracy of GSLDA minus that of \textsc{vSLDA}. $n/M$ is the median number of sentences assigned to a non-empty topic at the last Gibbs sweep of training divided by the embedding dimension $M=384$, Largest the percentage of the sentences assigned to the largest topic and Empty the largest number of topics without a sentence in any run. The unit and the \textsc{vSLDA} runs are those of Figure~\ref{fig:topic_sweep}.}
\label{tab:whole_corpus_gslda}
\begin{tabular}{@{}l r c c r @{\hspace{7pt}} r r r@{}}
\toprule
 & & \multicolumn{3}{c}{Accuracy (\%)} & \multicolumn{3}{c}{Topic assignments of GSLDA} \\
\cmidrule(lr){3-5} \cmidrule(lr){6-8}
Dataset & $K$ & GSLDA & \textsc{vSLDA} & $\Delta$ & $n/M$ & Largest (\%) & Empty \\
\midrule
20 Newsgroups & 10 & $35.8 \pm 2.6$ & $37.6 \pm 2.6$ & $-1.8$ & 26.4 & 16.9 & 0 \\
 & 20 & $41.6 \pm 1.0$ & $48.6 \pm 1.0$ & $-7.0$ & 13.4 & 13.8 & 0 \\
 & 30 & $42.3 \pm 2.0$ & $51.9 \pm 1.4$ & $-9.6$ & 3.3 & 13.7 & 0 \\
\midrule
NYT & 10 & $53.0 \pm 3.2$ & $66.6 \pm 3.2$ & $-13.6$ & 41.1 & 16.1 & 0 \\
 & 20 & $64.3 \pm 2.4$ & $77.2 \pm 1.4$ & $-13.0$ & 19.9 & 13.6 & 0 \\
 & 30 & $63.4 \pm 3.4$ & $82.3 \pm 0.7$ & $-18.9$ & 7.5 & 13.0 & 0 \\
\bottomrule
\end{tabular}
\end{table}

\paragraph{Prior-scale sweep}
An alternative explanation of the gap is that the scale of the prior on the topic covariances, which controls how strongly the estimated covariances are pulled toward the prior, was set unfavorably.
To examine this, we swept the prior scale $\lambda$, which sets the scale matrix of the inverse-Wishart prior on the topic covariances to $\lambda\bm{I}$, over $\{0.01, 0.1, 1, 3, 10, 30, 3M\}$ for GSLDA at $K=20$ with MiniLM embeddings, with three runs per condition, and averaged over the coarse-category units of each dataset (Table~\ref{tab:prior_scale}). The grid contains the value reported by \citet{das-2015-gaussian} ($\lambda=3$), the default of the reference implementation ($\lambda=3M$, that is, $1152$ at $M=384$), and the value adopted by \citet{cha-2024-sentencelda} and in our main experiments ($\lambda=0.1$).
Accuracy attains its maximum on both datasets at the $\lambda=0.1$ of the main experiments, an interior point of the grid. At this optimum, GSLDA remains below vSLDA (Section~\ref{subsec:why_gslda}) and $12.8$ and $25.8$ points below the vocabulary-based SentLDA, both reference values being recomputed over the same three runs and therefore differing slightly from Table~\ref{tab:cls_logreg_minilm}.
From $\lambda=10$ upwards, where $19$ of the $20$ topics are empty (Section~\ref{subsec:why_gslda}), training completes but the word-based evaluation cannot be carried out.
The transition to that state is abrupt. A single step from $\lambda=0.1$ to $\lambda=1$ empties most of the topics on both datasets, topic diversity collapses with them, and accuracy falls by $15.1$ and $16.7$ points. On 20 Newsgroups the highest $C_V$ over the grid occurs at that same $\lambda=1$, together with the loss of diversity and the decline in accuracy, so we do not read it as an improvement in topic quality.

These results indicate that the difference between GSLDA and vSLDA is not attributable to the choice of prior scale alone. The covariance-structure comparison below tests the remaining candidate, the covariance structure itself, and \appref{app:size_bias} states why no single scale can fit every topic size. The predictive scale in the directions a topic's sentences do not span, Eq.~\eqref{eq:s_perp}, falls as the topic size grows, so a value of $\lambda$ that suits topics of one size is too wide for smaller and too narrow for larger topics.
\begin{table}[!t]
\centering
\footnotesize
\setlength{\tabcolsep}{4pt}
\caption{Sensitivity of GSLDA to the scale matrix of its covariance prior. Classification accuracy (\%, logistic regression), number of empty topics out of $K=20$, topic diversity and $C_V$ of GSLDA as the prior scale $\lambda$, which sets the scale matrix of the inverse-Wishart prior on the topic covariances to $\lambda\bm{I}$, is varied at $K=20$ with MiniLM embeddings. $\lambda=0.1$ is the setting of the main experiments, $\lambda=3$ the value reported for Gaussian LDA, and $\lambda=3M=1152$ the default of the reference implementation. Values are means over the coarse-category units of each dataset and 3 runs per condition. An empty topic is one to which no sentence is assigned at the last Gibbs sweep of training; from $\lambda=10$ upwards $19$ of the $20$ topics are empty in every run, so the word-based measures cannot be computed (--). The last two rows give the same four quantities for \textsc{vSLDA} and SentLDA over the same runs, whose accuracy, diversity and $C_V$ therefore differ slightly from Tables~\ref{tab:cls_logreg_minilm} and~\ref{tab:topic_coherence_c_v}. A higher value is preferable in every column except Empty, and \textbf{bold} marks the highest accuracy of GSLDA on each dataset.}
\label{tab:prior_scale}
\begin{tabular}{@{}l *{4}{c} @{\hspace{7pt}} *{4}{c}@{}}
\toprule
\multirow[b]{2}{*}{$\lambda$} & \multicolumn{4}{c}{20 Newsgroups} & \multicolumn{4}{c}{NYT} \\
\cmidrule(lr){2-5} \cmidrule(lr){6-9}
 & Acc.\ (\%) & Empty & Div. & $C_V$ & Acc.\ (\%) & Empty & Div. & $C_V$ \\
\midrule
$0.01$ & 49.49 & 0.0 & 0.957 & 0.226 & 52.51 & 1.2 & 0.855 & 0.336 \\
$0.1$ (main) & \textbf{54.95} & 0.0 & 0.969 & 0.249 & \textbf{58.02} & 2.1 & 0.810 & 0.342 \\
$1$ & 39.90 & 18.2 & 0.125 & 0.273 & 41.31 & 17.2 & 0.181 & 0.289 \\
$3$ & 37.34 & 19.0 & 0.050 & 0.258 & 39.04 & 18.8 & 0.075 & 0.269 \\
$10$ & 37.34 & 19.0 & -- & -- & 38.72 & 19.0 & -- & -- \\
$30$ & 37.34 & 19.0 & -- & -- & 38.72 & 19.0 & -- & -- \\
$3M$ ($1152$) & 37.34 & 19.0 & -- & -- & 38.72 & 19.0 & -- & -- \\
\midrule
\textsc{vSLDA} & 73.51 & 0.0 & 0.996 & 0.457 & 86.27 & 0.0 & 0.995 & 0.554 \\
SentLDA & 67.71 & 0.0 & 0.999 & 0.324 & 83.80 & 0.0 & 0.996 & 0.481 \\
\bottomrule
\end{tabular}
\end{table}

\paragraph{Covariance structure}
To isolate the covariance structure itself, GSLDA is refit with a diagonal and with an isotropic (spherical) topic covariance. The inverse-Wishart prior is replaced by scaled inverse-$\chi^2$ priors on the variances, one per dimension under the diagonal structure and a single one under the isotropic structure, with scale $\lambda$ and $\nu_0' = \nu_0 - M + 1 = 1$ and with the prior on the mean unchanged, so that an empty topic keeps the predictive degrees of freedom (one) and the predictive scale $\lambda(\tau_0+1)/\tau_0$ of the full model, $\tau_0=0.1$ being the prior strength of the mean (\appref{app:size_bias}). Under the isotropic structure, the empty-topic predictive distribution is identical to the full one. The posterior predictive of a topic with $n$ sentences is a multivariate $t$ distribution with $\nu_0' + nM$ degrees of freedom under the isotropic structure, one variance pooled over the $nM$ coordinates of its sentences, and a product of univariate $t$ distributions with $\nu_0' + n$ degrees of freedom under the diagonal one, one variance per dimension estimated from $n$ observations.

Table~\ref{tab:covariance} summarizes, for the three structures and for vSLDA, the classification accuracy relative to vSLDA over the 30 conditions of Table~\ref{tab:cls_logreg_minilm}, and the shares of empty topics and of the largest topic. Both refits are ahead of the full covariance in every condition, and the isotropic one is level with vSLDA on average (a mean gap of $0.0$ points) and ahead of it in $10$ of the $30$.
Averaged over the 150 runs of each structure, the largest topic holds $51.0\%$ of the sentences under the full covariance, $16.2\%$ under the diagonal and $11.1\%$ under the isotropic structure, and the empty topics fall from $4.0\%$ of $K$ under the full covariance to none under the isotropic one. At $K=20$ the isotropic refit reaches a mean $C_V$ of $0.516$ against $0.285$ for the full covariance and $0.498$ for vSLDA, above vSLDA on $C_V$ in $9$ of the $10$ units, and over the prior scales $\lambda \in \{0.01, 0.1, 1, 10\}$ its accuracy moves by at most $0.2$ and $0.8$ points on the two datasets against $17.6$ and $19.3$ for the full model, so the sensitivity to the prior scale documented above is itself a property of the full structure.
The diagonal refit instead leaves $30.2\%$ of $K$ empty on average, $6.7$ of $20$ topics at $K=20$ and $15.8$ of $30$ at $K=30$, so the word-based measures, which require the representative words of every topic, cannot be computed for most of its runs, as for the full covariance from $\lambda=10$ upwards.

\begin{table}[!t]
\centering
\footnotesize
\setlength{\tabcolsep}{4pt}
\caption{Classification accuracy relative to \textsc{vSLDA} and topic-assignment diagnostics of the sentence-level Gaussian topic model under the three covariance structures (MiniLM). \emph{Full} is the normal--inverse-Wishart model of Table~\ref{tab:cls_logreg_minilm}; \emph{diagonal} and \emph{isotropic} replace it by scaled-inverse-$\chi^2$ priors on the variances, leaving the prior on the mean, the prior width of an empty topic and every other setting unchanged. The rows under \emph{unnormalized input} repeat the three structures on the pooled encoder output taken before the terminal normalization module of the released encoder. $\Delta$ is the classification accuracy of the structure (\%, logistic regression) minus that of \textsc{vSLDA}, averaged over the 30 conditions of Table~\ref{tab:cls_logreg_minilm} (5 runs each), on which \textsc{vSLDA} averages 77.5\%; Ahead is the number of those conditions in which the structure's mean accuracy exceeds that of \textsc{vSLDA}. Empty is the percentage of the $K$ topics to which no sentence is assigned at the last Gibbs sweep of training, and Largest the percentage of the sentences assigned to the largest topic, both averaged over the same 150 runs. The word-based measures of the refits at $K=20$ are reported in the text.}
\label{tab:covariance}
\begin{tabular}{@{}l c c @{\hspace{7pt}} c c@{}}
\toprule
\multirow[b]{2}{*}{Model} & \multicolumn{2}{c}{Classification accuracy} & \multicolumn{2}{c}{Topic assignments} \\
\cmidrule(lr){2-3} \cmidrule(lr){4-5}
 & $\Delta$ vs.\ \textsc{vSLDA} & Ahead (of 30) & Empty (\%) & Largest (\%) \\
\midrule
GSLDA, full & $-22.4$ & 0 & 4.0 & 51.0 \\
GSLDA, diagonal & $-0.8$ & 9 & 30.2 & 16.2 \\
GSLDA, isotropic & $0.0$ & 10 & 0.0 & 11.1 \\
\midrule
\multicolumn{5}{@{}l}{\textbf{Unnormalized input}} \\
\addlinespace[1pt]
\quad Full & $-30.1$ & 0 & 0.0 & 63.3 \\
\quad Diagonal & $-3.0$ & 0 & 0.6 & 12.2 \\
\quad Isotropic & $-3.7$ & 0 & 0.0 & 10.3 \\
\midrule
\textsc{vSLDA} & -- & -- & 0.0 & 10.5 \\
\bottomrule
\end{tabular}
\end{table}

\paragraph{Normalization}
The first three rows of Table~\ref{tab:covariance} and vSLDA are all fitted to L2-normalized embeddings, so they vary the covariance structure with the directional restriction held fixed. That the isotropic refit reaches the accuracy of vSLDA leaves open whether the directional restriction contributes to that parity or whether an isotropic Gaussian on the unnormalized embeddings would reach it as well, and whether the collapse of the full covariance depends on the normalization. The three rows under \emph{unnormalized input} test this: they vary that restriction rather than the number of covariance parameters, repeating each structure on the pooled encoder output taken before the terminal normalization module of the released encoder with everything else unchanged. The training objective of the encoder is a function of the cosine between embeddings, so the module removes a component that training does not constrain, and the comparison asks what the model gives up when it is removed. Over the same 30 conditions the isotropic refit falls $3.7$ points behind vSLDA and the diagonal one $3.0$, and on average neither leaves more than $0.6\%$ of the topics empty or holds more than $12.2\%$ of the sentences in its largest topic. The full covariance falls $30.1$ points behind, against $22.4$ on the normalized input, and its largest topic holds $63.3\%$ of the sentences against $51.0\%$, although none of its topics is left empty. Removing the normalization therefore costs accuracy under every structure, and the collapse appears under the full covariance on either input but under neither reduced structure.

Per dataset, at $K=20$ with MiniLM and $\lambda=0.1$, over five runs of each condition, the category mean accuracy of the isotropic refit is $72.5$ on the normalized input and $70.2$ without the normalization on 20 Newsgroups, and $87.3$ against $83.0$ on the New York Times. The normalized input is ahead in $26$ of the $30$ paired runs on the first dataset and in $19$ of the $20$ on the second, against a standard deviation of the category mean across the runs of one condition of $0.7$ and $0.6$ points. The parity of the isotropic refit with vSLDA therefore rests on the directional restriction: with the covariance structure held fixed, the norm that the normalization discards lowers the accuracy rather than adding information the model can use.
Across the whole grid the normalized input is ahead in $136$ of the $150$ paired runs under the isotropic structure but in $107$ under the diagonal one, where the margin narrows as $K$ grows and reverses on the New York Times at $K=30$. The coordinate scale of the input enters the diagonal posterior more directly, each of its variances resting on $n$ observations rather than $nM$, so we base the reading above on the isotropic comparison and take the similar gaps of the diagonal and isotropic rows under \emph{unnormalized input} in Table~\ref{tab:covariance} as an average over conditions in which the diagonal structure does not behave uniformly.
Nor does the comparison rest on the prior scale: on 20 Newsgroups the category mean of the isotropic refit moves by $0.1$ points or less between $\lambda=0.1$ and $\lambda=1$ on either input, more than an order of magnitude below the difference between the inputs.
\FloatBarrier

\section{Dependence of the Gaussian Posterior Predictive on Topic Size}
\label{app:size_bias}
This appendix derives the facts Section~\ref{subsec:why_gslda} relies on from the conjugate posterior of the Gaussian, applied to the prior and the topic sizes of the GSLDA reference implementation.
\paragraph{Setting}
Let $\bm w_1,\dots,\bm w_n$ denote the observations assigned to one topic, unit vectors in $\mathbb{R}^M$, and $\bm w^\ast$ a further observation of the same topic. Under $\mathrm{vMF}(\bm\mu,\kappa)$, $\mathbb{E}[\bm w]=A\bm\mu$ with $A=A_M(\kappa)$ and $\mathbb{E}\|\bm w-A\bm\mu\|^2=1-A^2$, so the variance per coordinate averaged over directions is $\sigma^2=(1-A^2)/M$. For the fitted topics of the main experiments ($M=384$, $\kappa\approx250$), $A\approx0.49$ and $\sigma^2\approx2.0\times10^{-3}$. GSLDA places on each topic the prior of the reference implementation, $\bm\Sigma\sim\mathrm{IW}(\nu_0=M,\;\bm\Psi_0=\lambda\bm I)$ and $\bm m\mid\bm\Sigma\sim\mathcal{N}(\bm\mu_0,\bm\Sigma/\tau_0)$ with $\lambda=\tau_0=0.1$, where the prior strength $\tau_0$ of the mean is the pseudo-count of prior observations. It evaluates a sentence under the posterior predictive of each topic~\citep{das-2015-gaussian,cha-2024-sentencelda}, which for $n$ observations is the multivariate $t$ distribution with $\nu_0+n-M+1=n+1$ degrees of freedom, location $\bm\mu_n=(\sum_i\bm w_i+\tau_0\bm\mu_0)/(\tau_0+n)$ and scale matrix
\begin{equation}
  \bm\Sigma_n=\frac{\tau_0+n+1}{(\tau_0+n)(n+1)}\,\bm\Psi_n,
  \qquad
  \bm\Psi_n=\lambda\bm I+\bm S_n+\frac{\tau_0 n}{\tau_0+n}(\bar{\bm w}-\bm\mu_0)(\bar{\bm w}-\bm\mu_0)^\top,
  \label{eq:niw_predictive}
\end{equation}
with $\bar{\bm w}$ the sample mean and $\bm S_n$ the scatter matrix. As a $t$ distribution, the predictive has covariance $(n+1)/(n-1)$ times $\bm\Sigma_n$ for $n>1$ and no finite covariance for $n\le1$.
\paragraph{Predictive scale in directions the data do not span}
Let $V_n=\operatorname{span}\{\bm w_i-\bm\mu_0: i=1,\dots,n\}$ and $V_n^\perp$ its orthogonal complement, of dimension at least $M-n$. $\bm S_n$ has rank at most $n-1$ and the last term of $\bm\Psi_n$ rank one, and both vanish on $V_n^\perp$, so $\bm\Psi_n-\lambda\bm I$ has rank at most $n$ and $V_n^\perp$ lies in its null space. On $V_n^\perp$, $\bm\Psi_n$ is $\lambda$ times the identity and $\bm\Sigma_n$ the scalar
\begin{equation}
  s_\perp(n)=\frac{\lambda\,(\tau_0+n+1)}{(\tau_0+n)(n+1)},
  \label{eq:s_perp}
\end{equation}
a function of the prior scale $\lambda$, the prior strength $\tau_0$ of the mean and $n$ alone. This is the scale Section~\ref{subsec:why_gslda} refers to, which depends on the observations only through $n$. That $s_\perp(n)\approx\lambda/(n+1)$ shrinks with $n$ although the observed residuals $\bm w_i-\bm\mu_0$ have no variation in $V_n^\perp$ is a property of the inverse-Wishart posterior, whose single degrees-of-freedom parameter counts the $n$ observations as evidence of a small variance in every direction. For an empty topic, $s_\perp(0)=1.1$ under the prior of the main experiments ($\lambda=0.1$, $\tau_0=0.1$), a $t$ scale (one degree of freedom, no finite variance) of about $1.05$ per coordinate against per-coordinate standard deviations of $0.047$ (MiniLM) and $0.033$ (MPNet) for the normalized embeddings. In the GSLDA run on 20 Newsgroups/\textit{computer} at $K=20$ (Section~\ref{subsec:why_gslda}) the null space exceeds $M-n$ for every topic with $n<M$, since the embeddings of a topic span fewer than $n$ directions.

$s_\perp(n)=\sigma^2$ holds only at $n^\ast\approx\lambda/\sigma^2-1$, about $50$ for $\lambda=0.1$ at $M=384$ and about $100$ at $M=768$ (equating the predictive variance rather than the scale to $\sigma^2$ moves $n^\ast$ by two). For $n<n^\ast$ the predictive scale on $V_n^\perp$ is above this average variance of the data, for $n>n^\ast$ below it. The non-empty topics of that run hold between $154$ and $5{,}482$ sentences, and on NYT the same setting yields topics of fewer than ten and of more than $10{,}000$ sentences, so a $\lambda$ that suits topics of one size is too wide for smaller and too narrow for larger topics, and the sweep of Table~\ref{tab:prior_scale} moves the mismatch across topic sizes rather than removing it.
\paragraph{Expected log-density deficit}
\begin{figure}[!t]
\centering
\includegraphics{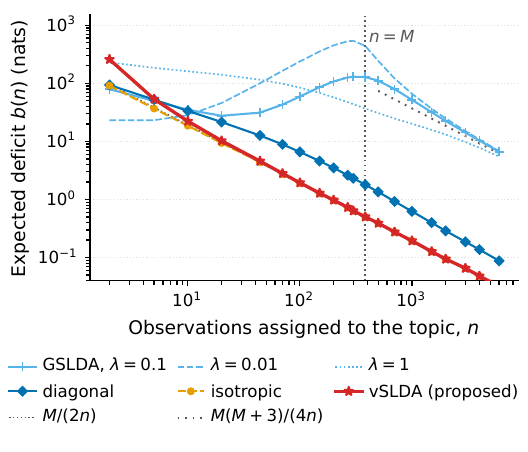}
\caption{\textbf{Expected log-density deficit of a topic against its number of observations.} Each curve is the deficit $b(n)$ of Eq.~\eqref{eq:deficit} for one observation model, for vMF observations at $M=384$ and $\kappa=250$ (both axes logarithmic). The models are the full-covariance predictive of GSLDA at three prior scales $\lambda$, its diagonal and isotropic variants at $\lambda=0.1$ (all with $\tau_0=0.1$), and the vMF likelihood of vSLDA at the estimates of its M-step. Each point is the mean over $50$ independent sets of $n$ observations and $2{,}000$ test draws, and two Monte Carlo standard errors are at most $5$\% of each value. The dotted lines are $M/(2n)$ and $M(M+3)/(4n)$, the large-sample deficits of correctly specified models with the numbers of parameters of the vMF and the full-covariance topic, and the dotted vertical line marks $n=M$. The vMF and isotropic curves follow $M/(2n)$ and overlap, whereas under $\lambda=0.1$ and $\lambda=0.01$ the full-covariance deficit peaks near $n=M$, where both are below one nat.}
\label{fig:size_bias}
\end{figure}
Under the full-covariance predictive, a topic of a few hundred sentences is at an expected disadvantage of about $100$ nats against a large topic with the same distribution (Figure~\ref{fig:size_bias}). The sampler cannot offset a difference of this size, and a change of $\lambda$ moves it to other topic sizes rather than removing it. We measure this loss by the deficit
\begin{equation}
  b(n)=\mathbb{E}\bigl[\log p_\infty(\bm w^\ast)\bigr]-\mathbb{E}\bigl[\log p_n(\bm w^\ast\mid\bm w_1,\dots,\bm w_n)\bigr],
  \label{eq:deficit}
\end{equation}
where $p_n$ is the predictive density of a topic fitted to the observations $\bm w_1,\dots,\bm w_n$, $p_\infty$ its limit as $n\to\infty$, and the expectations are over the observations and $\bm w^\ast$. For $n\gg M$ and a model that contains the distribution of the observations and whose predictive is asymptotically the maximum likelihood fit with $p$ free parameters, $b(n)=p/(2n)+o(1/n)$~\citep{akaike-1974-aic}. This holds for the vMF topic, with $p=M$. The Gaussian topics are fitted to vMF draws and do not contain their distribution, so their constant depends on the misspecification, and $p=M+M(M+1)/2$ gives only the reference value of a correctly specified model of the size of the full-covariance Gaussian topic. For $n$ up to about $M$ the deficit is not a quantity of order $1/n$, and we evaluate it by simulation.

In the simulation, observations are vMF draws at $\kappa=250$ and the prior mean $\bm\mu_0$ is set to the mean $A\bm\mu$ of the topic's distribution, estimated from the test draws, whereas the experiments set it to the corpus mean, which lies farther from the observations of a topic. Each deficit is the mean over $50$ independent training sets of size $n$ and $2{,}000$ test draws relative to the limit of the same model as $n\to\infty$, which is available in closed form (for the Gaussian topics, the Gaussian with the mean and covariance of the vMF distribution). The transcribed full-covariance predictive agrees with the density kernel of the reference implementation on an archived GSLDA state to $10^{-9}$.

The vMF deficit follows $M/(2n)$ ($1.9$ nats at $n=100$, $0.7$ at $n=267$), as does the isotropic Gaussian, and the diagonal one lies a few times higher. The full-covariance deficit instead has a peak near $n\approx M$ under $\lambda=0.1$ and $\lambda=0.01$, largest on the simulated grid at $n=300$ ($130$ and $539$ nats). At $n=267$ it is $126$ nats under $\lambda=0.1$ ($531$ under $\lambda=0.01$, $47$ under $\lambda=1$), against $23$ nats for a topic of $2{,}000$ sentences. It is still $6.6$ nats at $n=6{,}000$, the largest size simulated, and the reference large-sample form $M(M+3)/(4n)$ falls below one nat only from $n\approx3.7\times10^{4}$, about twice the largest median topic size in our data ($15{,}788$ sentences, NYT as a whole at $K=10$), whereas the vMF deficit reaches one nat at $n\approx M/2$ ($1.0$ nat at $n=200$). At $M=768$ and $\kappa=500$ the peak lies near $n\approx M$ at about $260$ nats ($259$ at $n=600$).

A sentence is reassigned by comparing these log densities across topics, and the document-level term $\log(n_{dk}+\alpha_k)$ of the sampler cannot offset an expected difference of about $100$ nats. A larger $\lambda$ lowers the deficit near $n\approx M$, and under $\lambda=1$ the deficit has no interior peak and decreases with $n$ from the empty topic on, but it raises the deficit of an empty topic ($172$ nats at $\lambda=0.1$, $27$ at $\lambda=0.01$, $526$ at $\lambda=1$), so that small topics cannot attract their first sentences, the state observed from $\lambda=1$ upwards in Table~\ref{tab:prior_scale}.
\paragraph{Consequence for the choice of observation model}
The computations above point to the number of variance parameters estimated per topic, and the number of observations behind each of them, rather than to the vMF distribution itself. At the topic sizes of Section~\ref{subsec:why_gslda}, a model with $M(M+1)/2$ covariance parameters is fitted where its predictive is set by the prior in most directions. Among the Gaussian covariance structures, the diagonal and isotropic models both have $O(M)$ parameters in total, but the diagonal model estimates each of its $M$ variances from $n$ observations, whereas the isotropic model pools one variance over the $nM$ coordinate residuals. The vMF topic has $M$ parameters, $M-1$ for the mean direction and one for $\kappa$, and no variance parameter beyond $\kappa$. The vMF and covariance estimates also differ in estimation error. The resultant $\bm r$ of $n$ unit vectors, of which both vMF estimates are functions, concentrates at a rate independent of $M$, $\Pr(\|\bm r/n-A\bm\mu\|\ge t)\le2\exp(-nt^2/8)$~\citep{pinelis-1994-martingales}, whereas a sample covariance of $n$ observations has rank at most $n$ and is singular for $n<M$. On unit vectors, the density of an isotropic Gaussian with fixed mean $\bm m$ and variance $s^2$ is proportional to a vMF density: for $\|\bm w\|=1$,
\begin{equation}
  -\frac{\|\bm w-\bm m\|^2}{2s^2}=-\frac{1+\|\bm m\|^2}{2s^2}+\Bigl(\frac{\bm m}{s^2}\Bigr)^{\!\top}\bm w,
  \label{eq:iso_is_vmf}
\end{equation}
so the density restricted to the sphere and renormalized there is $\mathrm{vMF}(\bm m/\|\bm m\|,\|\bm m\|/s^2)$. The identity holds for fixed parameters. The isotropic refit of Table~\ref{tab:covariance} integrates its mean and variance out and evaluates a Student-$t$ predictive, which restricted to the sphere is not a vMF density. The ordering full $\ll$ diagonal $<$ isotropic $\approx$ vMF observed when the covariance structure of GSLDA is varied with everything else fixed (Table~\ref{tab:covariance}) is therefore the one these counts suggest, but its last step, the parity of the isotropic refit with vSLDA, is an empirical finding.
\FloatBarrier

\section{Comparison of GSLDA with the Released SentenceLDA Implementation}
\label{app:reference_code}
This appendix records how the accuracy of GSLDA in this paper relates to the accuracy published for SentenceLDA~\citep{cha-2024-sentencelda}. The study specifies, per topic, a Gaussian with a full covariance matrix under an inverse-Wishart prior with the scale matrix $0.1\bm I$ (its Table~2 and Appendix~A), and infers it by the collapsed Gibbs sampler with Cholesky decomposition of \citet{das-2015-gaussian}; our GSLDA implements this specification. To compare the two, we ran the implementation from the repository the study names\footnote{\url{https://github.com/cth127/sentencelda}, retrieved on 8 September 2026 (commit 2b27335).} under SciPy 1.16.3 on one experimental unit, as released and with the one change described below. The study does not, however, record the version used for its own experiments, so we cannot verify that the code we retrieved is the code behind the published values. The two implementations agree in the per-category training, the prior, the embeddings, the fold-in procedure and the classifier. They differ in the number of training sweeps (20 against our 10) and the iteration limit of the logistic regression, neither of which moves a value on this unit beyond the run-to-run spread (\appref{app:hyperparams}).

The sampler scores a sentence $\bm w$ under the Student-$t$ posterior predictive of topic $k$, whose scale matrix $\bm\Sigma_k=\bm L_k\bm L_k^\top$ is that of Eq.~\eqref{eq:niw_predictive} for the sentences currently assigned to the topic. The quadratic form $q_k(\bm w)=(\bm w-\bm\mu_k)^\top\bm\Sigma_k^{-1}(\bm w-\bm\mu_k)$ is obtained as $\|\bm y\|^2$, where $\bm L_k\bm y=\bm w-\bm\mu_k$ is solved by forward substitution with the lower-triangular factor $\bm L_k$, as in the derivation of \citet{das-2015-gaussian} (their Section~4.1).
In the training sampler and in the fold-in code alike, the released implementation obtains $\bm y$ from the library's triangular solver under its default setting, which treats the matrix as upper triangular.\footnote{\texttt{scipy.linalg.solve\_triangular} with its default \texttt{lower=False}. The same call appears in the plain Cholesky trainer of the reference Gaussian LDA package, from which the implementation descends. The factor is nonetheless held lower triangular, since its rank-one updates pass $\bm L_k^\top$ to a routine that updates an upper-triangular factor. Only the routine that reports the average log-likelihood during training passes \texttt{lower=True}, and it does not enter the sampling of the topic assignments. The solver call is present from the repository's first commit (7 February 2024).} The solver thus ignores the entries of $\bm L_k$ below the diagonal and, since those above it are zero, solves a diagonal system: it returns $y_i=(\bm w-\bm\mu_k)_i/L_{k,ii}$, and the quadratic form evaluated is
\begin{equation}
  \tilde q_k(\bm w) = \sum_{i=1}^{M}\Big(\frac{(\bm w-\bm\mu_k)_i}{L_{k,ii}}\Big)^2
  = (\bm w-\bm\mu_k)^\top\bm D_k^{-1}(\bm w-\bm\mu_k),
  \qquad
  \bm D_k=\operatorname{diag}\big(L_{k,11}^2,\dots,L_{k,MM}^2\big),
  \label{eq:ref_diag}
\end{equation}
while the determinant term, $\tfrac12\log|\bm\Sigma_k|=\sum_i\log L_{k,ii}$, is unchanged. Because $|\bm D_k|=\prod_i L_{k,ii}^2=|\bm\Sigma_k|$, the two terms are consistent with each other: the density that scores a sentence is the Student-$t$ predictive with, in place of $\bm\Sigma_k$, the diagonal scale matrix $\bm D_k$, whose entries are the Cholesky pivots of the full covariance estimate the sampler maintains. Our implementation carries out the forward substitution in full.

Table~\ref{tab:reference_code} reports the runs. The released implementation was run with all-mpnet-base-v2 and with the data split and sentence segmentation of its repository; our GSLDA and vSLDA use the split of Section~\ref{subsec:datasets} and the same encoder. Run as released, the implementation keeps every topic populated and closely reproduces the accuracy published for the unit ($88.4 \pm 0.6$ in its Table~4). With the forward substitution carried out in full and nothing else changed, it concentrates most of the sentences on one topic and falls to the accuracy level of our GSLDA. Examined separately, none of L2 normalization, the encoder, the number of training sweeps, the prior scale (Table~\ref{tab:prior_scale}) and the released data split brings the full-covariance model to the published values. We did not run the released implementation on the other conditions of Table~\ref{tab:cls_logreg_minilm}. On those conditions, a diagonal-covariance model estimated directly rather than through the Cholesky pivots is covered by the diagonal refit of Table~\ref{tab:covariance}, which we ran over all 30.

\begin{table}[!t]
\centering
\footnotesize
\setlength{\tabcolsep}{4pt}
\caption{Classification accuracy and topic-assignment diagnostics of the released SentenceLDA implementation, GSLDA and \textsc{vSLDA} on 20 Newsgroups/\textit{sports} at $K=20$ (all-mpnet-base-v2). The first two rows are the implementation released with \citet{cha-2024-sentencelda}, run as released, where the posterior predictive has the diagonal scale matrix of Eq.~\eqref{eq:ref_diag}, and run with the forward substitution carried out in full and nothing else changed. The last two rows are the GSLDA of this paper, which evaluates the full quadratic form, and \textsc{vSLDA}, which has no scale matrix. Accuracy is the classification accuracy (\%, logistic regression), \emph{Largest} the percentage of the training sentences assigned to the largest topic and \emph{Empty} the number of topics to which no sentence is assigned. Values are the mean $\pm$ standard deviation over 5 runs.}
\label{tab:reference_code}
\begin{tabular}{@{}l l c c c@{}}
\toprule
Model & Scale matrix & Accuracy & Largest (\%) & Empty \\
\midrule
Released implementation, as released & diagonal, Eq.~\eqref{eq:ref_diag} & $89.1 \pm 0.9$ & $6.2 \pm 0.1$ & $0.0 \pm 0.0$ \\
Released implementation, full forward substitution & full & $56.1 \pm 1.2$ & $80.2 \pm 0.1$ & $0.0 \pm 0.0$ \\
\midrule
GSLDA (this paper) & full & $53.1 \pm 0.8$ & $92.8 \pm 0.5$ & $0.0 \pm 0.0$ \\
\textsc{vSLDA} & -- & $89.4 \pm 0.7$ & $9.1 \pm 0.4$ & $0.0 \pm 0.0$ \\
\bottomrule
\end{tabular}
\end{table}

We therefore read the published values as those of a sampler whose predictive uses the diagonal scale matrix $\bm D_k$ of Eq.~\eqref{eq:ref_diag} rather than of the full-covariance model that the study specifies and that GSLDA implements in this paper, but only as an inference: the runs cover one unit, and the version behind the published values is not recorded. The inference concerns which likelihood the published values reflect. It does not concern the conclusion of the study, that sentence-level topic models yield more discriminative document representations than word-level ones, which our own experiments support under a likelihood of a different family (Section~\ref{subsec:cls_results}). The reading is consistent with Table~\ref{tab:covariance}, where the diagonal and isotropic refits are close to vSLDA on average and the full-covariance model collapses.

\FloatBarrier

\section{Topic-Pair Statistics and Full Topic Listings}
\label{app:topic_pairs}
This appendix first defines the quantities of Table~\ref{tab:topic_pairs} and reports the pair statistics over every run of the main experiments. It then lists every topic of the two runs examined in Section~\ref{subsec:kappa_hierarchy}, defines the matching of topics across models, and states the rule by which the groups of Table~\ref{tab:family_topics} are selected from them and the correlations of Table~\ref{tab:kappa_entropy}.
\paragraph{Topic centroids and pair quantities}
The analysis covers every run of vSLDA, SentLDA and GSLDA in the main experiments (both datasets, every coarse category, $K\in\{10,20,30\}$, five runs; MiniLM) and applies the same procedure to each. Let $\bm w_{di}$ denote the L2-normalized MiniLM embedding of sentence $i$ of training document $d$, and $r_{dik}$ the sentence-topic posterior of Section~\ref{subsec:repwords}, re-estimated for the run by the collapsed fold-in with frozen parameters. The posteriors are therefore defined identically for the three models, although SentLDA does not read the embeddings during training. Topic $k$, of posterior mass $n_k=\sum_{d,i} r_{dik}$, is represented by the centroid direction of its sentences and by its mean resultant length,
\begin{equation}
  \hat{\bm\mu}_k = \frac{\sum_{d,i} r_{dik}\,\bm w_{di}}{\bigl\|\sum_{d,i} r_{dik}\,\bm w_{di}\bigr\|},
  \qquad
  \bar r_k = \frac{\bigl\|\sum_{d,i} r_{dik}\,\bm w_{di}\bigr\|}{\sum_{d,i} r_{dik}},
  \label{eq:pair-centroid}
\end{equation}
and $\bar r_k$ gives a concentration $\hat\kappa_k$ through the approximation of Eq.~\eqref{eq:kappa-approx}. For vSLDA, $\hat\kappa_k$ agrees with the fitted $\kappa_k$ to within $5$\% in the 20 Newsgroups/\textit{computer} run listed below and to within $8$\% in the NYT/\textit{arts} run. Topics that are the posterior mode of no sentence (which occurs for GSLDA on NYT and in two SentLDA runs on NYT/\textit{arts} at $K=30$) are excluded from every pair statistic.
For a pair of topics $(j,k)$ of one run, three quantities are recorded: the centroid cosine $s_{jk}=\hat{\bm\mu}_j^{\top}\hat{\bm\mu}_k$; the assignment confusion $c_{jk}$, the mean posterior mass on topic $k$ of the sentences whose posterior mode is topic $j$, reported symmetrized as $(c_{jk}+c_{kj})/2$; and the Jensen--Shannon divergence (natural logarithm) between the fine-category distributions of the posterior mass of the two topics, each sentence inheriting the fine category of its document. The closest pairs of a run are the top $5$\% of its $K(K-1)/2$ pairs by $s_{jk}$, rounded up ($3$, $10$ and $22$ pairs at $K=10$, $20$ and $30$ for a run with no excluded topic).
\begin{table}[!t]
\centering
\footnotesize
\setlength{\tabcolsep}{4pt}
\caption{Geometry of the topic pairs of the sentence-level models in the MiniLM sentence-embedding space, each topic represented by the posterior-weighted centroid direction of its training sentences, defined in~\eqref{eq:pair-centroid}. (a)~Mean centroid cosine over all $K(K-1)/2$ pairs; (b)--(d)~mean centroid cosine, mean symmetrized assignment confusion and mean fine-category Jensen--Shannon divergence (nats) of the closest 5\% of pairs. The assignment confusion of a pair $(j,k)$ is the mean posterior mass on topic $k$ of the sentences whose posterior mode is topic $j$, averaged over the two orders of the pair; the Jensen--Shannon divergence is taken with the natural logarithm between the fine-category distributions of the posterior mass of the two topics and is at most $\ln 2$. Values are means~$\pm$~standard deviations over the categories of the dataset and the 5 independent runs of each (30 values per cell for 20 Newsgroups and 20 for NYT).}
\label{tab:topic_pairs}
\adjustbox{max width=\textwidth}{%
\begin{tabular}{@{}l *{3}{c}@{\hspace{7pt}} *{3}{c}@{}}
\toprule
\multirow[b]{2}{*}{Model} & \multicolumn{3}{c}{20news} & \multicolumn{3}{c}{NYT} \\
\cmidrule(lr){2-4} \cmidrule(lr){5-7}
 & $K=10$ & $K=20$ & $K=30$ & $K=10$ & $K=20$ & $K=30$ \\
\midrule
\multicolumn{7}{@{}l}{\textbf{(a) Mean centroid cosine over all $K(K-1)/2$ topic pairs}} \\
\addlinespace[1pt]
SentLDA {\scriptsize(Balikas, 2016)} & \ms{0.592}{0.086} & \ms{0.572}{0.083} & \ms{0.564}{0.077} & \ms{0.638}{0.047} & \ms{0.589}{0.057} & \ms{0.562}{0.063} \\
GSLDA {\scriptsize(Cha, 2024)} & \ms{0.630}{0.067} & \ms{0.620}{0.087} & \ms{0.588}{0.120} & \ms{0.514}{0.132} & \ms{0.411}{0.155} & \ms{0.375}{0.150} \\
\textsc{vSLDA} {\scriptsize(proposed)} & \ms{0.449}{0.068} & \ms{0.401}{0.063} & \ms{0.376}{0.061} & \ms{0.531}{0.046} & \ms{0.481}{0.046} & \ms{0.454}{0.048} \\
\midrule
\multicolumn{7}{@{}l}{\textbf{(b) Mean centroid cosine of the closest 5\% of pairs}} \\
\addlinespace[1pt]
SentLDA {\scriptsize(Balikas, 2016)} & \ms{0.820}{0.057} & \ms{0.841}{0.037} & \ms{0.840}{0.039} & \ms{0.837}{0.035} & \ms{0.839}{0.037} & \ms{0.825}{0.042} \\
GSLDA {\scriptsize(Cha, 2024)} & \ms{0.830}{0.037} & \ms{0.801}{0.061} & \ms{0.775}{0.078} & \ms{0.780}{0.053} & \ms{0.713}{0.085} & \ms{0.671}{0.100} \\
\textsc{vSLDA} {\scriptsize(proposed)} & \ms{0.704}{0.062} & \ms{0.716}{0.051} & \ms{0.710}{0.048} & \ms{0.732}{0.043} & \ms{0.755}{0.032} & \ms{0.752}{0.027} \\
\midrule
\multicolumn{7}{@{}l}{\textbf{(c) Mean symmetrized assignment confusion of the closest 5\% of pairs (\%)}} \\
\addlinespace[1pt]
SentLDA {\scriptsize(Balikas, 2016)} & \ms{0.88}{0.44} & \ms{0.50}{0.17} & \ms{0.36}{0.10} & \ms{0.76}{0.52} & \ms{0.48}{0.45} & \ms{0.35}{0.32} \\
GSLDA {\scriptsize(Cha, 2024)} & \ms{0.00}{0.00} & \ms{0.00}{0.00} & \ms{0.00}{0.00} & \ms{0.01}{0.02} & \ms{0.01}{0.01} & \ms{0.01}{0.01} \\
\textsc{vSLDA} {\scriptsize(proposed)} & \ms{0.83}{0.20} & \ms{0.63}{0.11} & \ms{0.49}{0.08} & \ms{0.78}{0.20} & \ms{0.69}{0.15} & \ms{0.54}{0.11} \\
\midrule
\multicolumn{7}{@{}l}{\textbf{(d) Mean fine-label Jensen--Shannon divergence of the closest 5\% of pairs (nats)}} \\
\addlinespace[1pt]
SentLDA {\scriptsize(Balikas, 2016)} & \ms{0.037}{0.032} & \ms{0.045}{0.027} & \ms{0.039}{0.023} & \ms{0.106}{0.091} & \ms{0.074}{0.033} & \ms{0.083}{0.034} \\
GSLDA {\scriptsize(Cha, 2024)} & \ms{0.021}{0.014} & \ms{0.017}{0.008} & \ms{0.021}{0.007} & \ms{0.062}{0.047} & \ms{0.124}{0.044} & \ms{0.127}{0.061} \\
\textsc{vSLDA} {\scriptsize(proposed)} & \ms{0.047}{0.029} & \ms{0.040}{0.012} & \ms{0.042}{0.016} & \ms{0.117}{0.057} & \ms{0.087}{0.049} & \ms{0.087}{0.037} \\
\bottomrule
\end{tabular}}
\end{table}

\paragraph{Pair statistics over the main experiments}
First, in the embedding space in which vSLDA, but not SentLDA, fits its topics, the topic centroids of vSLDA are farther apart than those of SentLDA. The mean cosine between topic centroids is lower at every topic count (at $K=20$, $0.40$ against $0.57$ on 20 Newsgroups and $0.48$ against $0.59$ on NYT), and its closest $5$\% of pairs are farther apart ($0.72$ and $0.75$ against $0.84$ and $0.84$ at $K=20$ on 20 Newsgroups and NYT). Averaged over the categories, GSLDA lies above both on 20 Newsgroups ($0.62$ at $K=20$) and below both on NYT ($0.41$), where its mean cosine is below that of vSLDA at every topic count. The NYT average is lowered by \textit{arts} and \textit{business}, the two categories in which runs of GSLDA leave topics without sentences (Section~\ref{subsec:why_gslda}); in \textit{politics} and \textit{sports} its mean cosine is above that of vSLDA at every topic count. The low values thus coincide with the runs that leave topics without sentences and are not read as separated topics, and the closest-pair statistics of GSLDA are not compared. Because the comparison is made in the space vSLDA is fitted in, Section~\ref{subsec:kappa_hierarchy} does not rest on it. Second, the closest pairs are separated in assignment for every model. The symmetrized assignment confusion of the closest $5$\% of pairs is below $1$\% in every cell of panel~c of Table~\ref{tab:topic_pairs}, and the fine-category distributions of the two topics are close (panel~d, below $0.13$ nats of a maximum of $\ln 2$ in every cell).
\begin{table}[!t]
\centering
\small
\setlength{\tabcolsep}{8pt}
\caption{Rank correlation, within a run, between the concentration of a topic and the breadth of its fine-category distribution. For every \textsc{vSLDA} run of the main experiments (MiniLM; 6 units of 20 Newsgroups and 4 of NYT, 5 runs each), the Spearman correlation across the $K$ topics between the fitted $\kappa_k$ and the normalized fine-category entropy $H$ of the topic's posterior sentence mass, and the same correlation with the ranks of $\log n_k$ (the topic size) partialled out of both. Cells give the mean over the units and runs of the row with the range in brackets, and the number of runs in which the correlation is negative out of all runs of the row. Topics to which no sentence is assigned are excluded.}
\label{tab:kappa_entropy}
\begin{tabular}{@{}lrcccc@{}}
\toprule
Dataset & $K$ & $\rho(\kappa_k, H)$ & $\rho<0$ & $\rho(\kappa_k, H \mid \log n_k)$ & $\rho<0$ \\
\midrule
20news & 10 & $-0.69$ [$-0.95$, $-0.21$] & 30/30 & $-0.60$ [$-0.93$, $0.12$] & 28/30 \\
 & 20 & $-0.65$ [$-0.91$, $-0.27$] & 30/30 & $-0.63$ [$-0.88$, $-0.33$] & 30/30 \\
 & 30 & $-0.66$ [$-0.88$, $-0.32$] & 30/30 & $-0.65$ [$-0.87$, $-0.33$] & 30/30 \\
NYT & 10 & $-0.65$ [$-0.90$, $-0.16$] & 20/20 & $-0.57$ [$-0.88$, $0.06$] & 18/20 \\
 & 20 & $-0.61$ [$-0.86$, $-0.16$] & 20/20 & $-0.59$ [$-0.85$, $-0.16$] & 20/20 \\
 & 30 & $-0.60$ [$-0.82$, $-0.24$] & 20/20 & $-0.58$ [$-0.84$, $-0.15$] & 20/20 \\
\bottomrule
\end{tabular}
\end{table}

\paragraph{All topics of the two runs}
Tables~\ref{tab:topics_20news_computer_vslda}--\ref{tab:topics_nyt_arts_sentlda} list every topic of vSLDA and SentLDA in the two runs, so that the groups of Table~\ref{tab:family_topics} can be read in the context of the whole topic set. Each row gives the fitted concentration $\kappa_k$ (vSLDA only, since SentLDA fits no concentration), the posterior mass $n_k$ of the training sentences assigned to it, the normalized entropy of the fine-category distribution of that mass, the fine category holding the largest share of it and its top five representative words under the ranking of Section~\ref{subsec:repwords}. GSLDA is omitted because its mass is concentrated to a degree that makes a topic listing uninformative. On NYT/\textit{arts}, two of its topics hold $99$\% of the training sentences and half of the $20$ topics are empty, and on 20 Newsgroups/\textit{computer} three topics hold $58$\%. Its topic structure is described in Section~\ref{subsec:why_gslda}.
In the 20 Newsgroups/\textit{computer} run, the closest pairs of vSLDA include topics of low $\kappa$ whose sentences spread evenly over the fine categories (the sixth-closest pair joins two topics with $\kappa=159.2$ and $132.6$ and normalized fine-category entropy $0.99$ and $0.99$).
\paragraph{Qualitative reading of the listings}
The observations below on the four tables are qualitative. They are read from the representative words of one run per dataset and are not tested. Subjects with a distinct vocabulary appear as a topic in both models, for example requests for help on 20 Newsgroups/\textit{computer} (topic 1 of vSLDA and topic 12 of SentLDA) and ballet on NYT/\textit{arts} (topics 14 and 10).
Where the two models differ, the topics of SentLDA often follow regularities of wording. On NYT/\textit{arts} (Table~\ref{tab:topics_nyt_arts_sentlda}), topic 16 collects the words of home-video release listings (\textit{blu}, \textit{ray}, \textit{combo}, \textit{pg}), topic 19 those of venue listings (\textit{org}, \textit{avenue}, \textit{com}), and topics 2 and 9 names of opera roles and of composers, while topic 6 holds hip-hop and rock in one topic. On 20 Newsgroups/\textit{computer} (Table~\ref{tab:topics_20news_computer_sentlda}), topic 13 collects the words of file paths (\textit{usr}, \textit{lib}, \textit{bin}) and topic 18 those of file archives (\textit{mirror}, \textit{simtel}, \textit{zip}, \textit{pub}), and image files are divided into the vocabulary of color quantization (topic 1), the names of formats (topic 5) and one further topic (topic 15), which vSLDA holds as one topic (topic 13). SentLDA also holds subjects that have no topic of their own in the listing of vSLDA, jazz (topic 8) and the finances of arts institutions (topic 4) on NYT/\textit{arts}.
The topics of vSLDA more often follow the subject of the sentence. On NYT/\textit{arts} (Table~\ref{tab:topics_nyt_arts_vslda}), music is divided into instruments, rap, songs and albums, festivals, opera, vocal repertoire, and the distribution of recordings (topics 4, 17, 19, 15, 11, 16 and 6), and movies into awards, films and actors (topics 0, 2 and 7). On 20 Newsgroups/\textit{computer} (Table~\ref{tab:topics_20news_computer_vslda}), the topics of monitors, of processors and memory, of reports of faults, of prices and of ports and drivers (topics 3, 6, 10, 14 and 15) each have $H\geq 0.89$ and a dominant fine category that holds at most $42$\% of the mass. These subjects cross the division into newsgroups, so that a high $H$ does not by itself indicate a topic without a subject.
The embedding also groups sentences by features other than their subject. Topic 9 of vSLDA on NYT/\textit{arts} has the representative words \textit{her}, \textit{she} and \textit{actress}, and follows the gender of the person referred to and not a subject. Topic 18 has first- and second-person pronouns, as does topic 5 of SentLDA.
The vSLDA topics whose representative words share no evident subject (topics 18 and 19 on 20 Newsgroups/\textit{computer}, topics 10, 12 and 13 on NYT/\textit{arts}) have a low $\kappa_k$ and a high $H$ ($\kappa_k\leq 205.7$, $H\geq 0.90$). For a topic without a vocabulary of its own, the ranking of Section~\ref{subsec:repwords} places first the words near the minimum document frequency of the evaluation vocabulary (\appref{app:hyperparams}), and these word lists are not read as subjects. SentLDA has topics of the same kind (topic 8 of Table~\ref{tab:topics_20news_computer_sentlda} and topic 7 of Table~\ref{tab:topics_nyt_arts_sentlda}).
\paragraph{Matching of topics across models}
Let $A$ and $B$ be two models trained on the same training sentences (the same category, $K$ and run). $N_{ab}=\sum_{d,i} r^{A}_{dia}\,r^{B}_{dib}$, the expected co-assignment of $a$ and $b$, is the expected number of sentences that $A$ assigns to its topic $a$ and $B$ to its topic $b$, which with hard assignments would be the cross-tabulation of the two partitions. Each topic $a$ of $A$ is matched to the topic of $B$ that receives the largest share of its sentences, $\arg\max_b N_{ab}$. The matching is many-to-one, so several topics of $A$ may be matched to one topic of $B$; a topic of $B$ that receives several topics of $A$ represents by one topic what $A$ divides, without necessarily holding the majority of their sentences, because the matching reads only the plurality target of each topic. The expected co-assignment is computed between vSLDA and SentLDA for the two conditions of Section~\ref{subsec:kappa_hierarchy} (20 Newsgroups/\textit{computer} and NYT/\textit{arts}, $K=20$, five runs each). Table~\ref{tab:family_topics} shows the groups it defines in run 0 of each condition.
\paragraph{Selection of the groups}
Table~\ref{tab:family_topics} (Section~\ref{subsec:kappa_hierarchy}) shows, for each of the two runs, the subject that SentLDA represents by one topic and vSLDA divides into the most topics, receiving topics without a dominant subject being set aside. Formally, the selection has three steps. First, the vSLDA topics are grouped by the SentLDA topic they are matched to. Second, the groups whose receiving topic has a dominant fine category holding less than half of its posterior mass are excluded. Third, of the remaining groups the one with the most vSLDA topics is shown, a tie in the number of topics being broken by the co-assignment the receiving topic collects from the group. In both runs the excluded receiving topics include the one that receives the most topics, a topic whose sentences spread over the fine categories; on NYT/\textit{arts}, a SentLDA topic with $H=0.93$ receives seven vSLDA topics. In the opposite direction, the largest group under the same rule in the 20 Newsgroups/\textit{computer} run is that of the three SentLDA topics matched to the vSLDA topic of image files (topics 1, 5 and 15 of Table~\ref{tab:topics_20news_computer_sentlda} and topic 13 of Table~\ref{tab:topics_20news_computer_vslda}). The only other group of more than one topic that remains under the same rule in that run is the one quoted in Section~\ref{subsec:kappa_hierarchy}, the two SentLDA topics (topics 10 and 13) matched to topic 7 of vSLDA.
\paragraph{Correlation between $\kappa_k$ and $H$}
Table~\ref{tab:kappa_entropy} gives, per dataset and topic count, the Spearman correlation across the topics of a vSLDA run between the fitted $\kappa_k$ and $H$, as its mean and range over the units and runs and the number of runs in which it is negative, and the same correlation with the ranks of $\log n_k$ partialled out of both. The first-order finite-sample bias of \appref{app:estimation} raises the expectation of $\kappa_k$ by a relative amount of about $2.6/n_k$, below $1$\% for each vSLDA topic of panel~b of Table~\ref{tab:family_topics} (posterior mass $507$ to $606$). The approximation says nothing about the sampling variability of $\kappa_k$, and both $\kappa_k$ and $H$ may depend on the topic size; the table therefore reports the correlation over all runs rather than the ordering within one run, and also with the ranks of the topic size partialled out. For vSLDA, the correlation is negative in each of the $150$ runs, with a mean between $-0.69$ and $-0.60$ per dataset and topic count, and with the topic size partialled out it is negative in $146$ of the $150$ runs, with a mean between $-0.65$ and $-0.57$; the four runs in which it is not negative are at $K=10$. A negative correlation means that the more concentrated topics of a run tend to have narrower fine-category distributions, although not without exception, since a topic whose sentences spread over the fine categories can also receive a large $\kappa_k$.
\clearpage
\begin{table}[!t]
\centering
\small
\setlength{\tabcolsep}{3pt}
\caption{Every topic of \textsc{vSLDA} in the 20 Newsgroups/\textit{computer} run ($K=20$, MiniLM, run 0), with the fitted concentration parameter $\kappa_k$, its posterior sentence mass $n_k$, its normalized fine-category entropy $H$, its dominant fine category with the share and its top 5 representative words; the columns are defined in the text of this appendix.}
\label{tab:topics_20news_computer_vslda}
\begin{tabularx}{0.9\textwidth}{@{}rrrrl>{\raggedright\arraybackslash\fontsize{9.5}{11.5}\selectfont}X@{}}
\toprule
\footnotesize Topic & \footnotesize $\kappa_k$ & \footnotesize $n_k$ & \footnotesize $H$ & \footnotesize Fine category & \footnotesize Representative words \\
\midrule
0 & 251.7 & 1254 & 0.87 & {\footnotesize os.ms-windows.misc (44\%)} & nt, msw, microsoft, window, apps \\
1 & 288.8 & 683 & 0.99 & {\footnotesize graphics (23\%)} & appreciate, thank, help, greatly, advance \\
2 & 138.7 & 1413 & 0.85 & {\footnotesize graphics (47\%)} & mahan, tgv, patrick, washer, printf \\
3 & 289.3 & 866 & 0.95 & {\footnotesize graphics (33\%)} & monitor, vga, resolution, hz, multisync \\
4 & 263.0 & 1385 & 0.43 & {\footnotesize graphics (80\%)} & graphic, visualization, ray, animation, plot \\
5 & 281.7 & 524 & 0.39 & {\footnotesize sys.mac.hardware (85\%)} & apple, mac, macintosh, appletalk, macweek \\
6 & 237.6 & 1408 & 0.89 & {\footnotesize sys.ibm.pc.hardware (38\%)} & mhz, cpu, speed, ram, cache \\
7 & 267.3 & 1459 & 0.17 & {\footnotesize windows.x (94\%)} & xterm, xdm, xmu, argv, libxmu \\
8 & 213.5 & 1937 & 0.84 & {\footnotesize graphics (47\%)} & ftp, edu, mail, lipman, anonymous \\
9 & 207.1 & 784 & 0.94 & {\footnotesize sys.ibm.pc.hardware (36\%)} & pin, hartmann, lance, mcc, ft \\
10 & 215.1 & 1658 & 0.92 & {\footnotesize os.ms-windows.misc (33\%)} & reinstall, problem, symptom, reboot, error \\
11 & 293.4 & 883 & 0.59 & {\footnotesize sys.ibm.pc.hardware (64\%)} & drive, disk, hard, floppy, mfm \\
12 & 196.7 & 1256 & 0.90 & {\footnotesize graphics (32\%)} & imakefile, shar, lha, file, font \\
13 & 406.7 & 1101 & 0.24 & {\footnotesize graphics (92\%)} & jpeg, gif, image, lossless, gifs \\
14 & 188.7 & 1229 & 0.92 & {\footnotesize sys.ibm.pc.hardware (34\%)} & price, discount, cost, retail, pound \\
15 & 238.5 & 1332 & 0.92 & {\footnotesize sys.ibm.pc.hardware (42\%)} & driver, port, irq, scsi, modem \\
16 & 140.2 & 1417 & 0.91 & {\footnotesize windows.x (41\%)} & button, expose, ctrl, clockwise, roller \\
17 & 229.8 & 1109 & 0.86 & {\footnotesize windows.x (34\%)} & paradox, tcl, engine, borland, vogle \\
18 & 159.2 & 1256 & 0.99 & {\footnotesize graphics (29\%)} & accumulation, trivially, confusion, vote, deliberate \\
19 & 132.6 & 1554 & 0.99 & {\footnotesize graphics (25\%)} & clothe, dftsrv, forevermore, food, devil \\
\bottomrule
\end{tabularx}
\end{table}

\begin{table}[!t]
\centering
\small
\setlength{\tabcolsep}{3pt}
\caption{Every topic of SentLDA in the 20 Newsgroups/\textit{computer} run ($K=20$, MiniLM, run 0), with its posterior sentence mass $n_k$, its normalized fine-category entropy $H$, its dominant fine category with the share and its top 5 representative words; the columns are defined in the text of this appendix.}
\label{tab:topics_20news_computer_sentlda}
\begin{tabularx}{0.9\textwidth}{@{}rrrl>{\raggedright\arraybackslash\fontsize{9.5}{11.5}\selectfont}X@{}}
\toprule
\footnotesize Topic & \footnotesize $n_k$ & \footnotesize $H$ & \footnotesize Fine category & \footnotesize Representative words \\
\midrule
0 & 1329 & 0.89 & {\footnotesize graphics (40\%)} & mail, subscribe, send, newsgroups, subscription \\
1 & 898 & 0.46 & {\footnotesize graphics (79\%)} & color, quantization, palette, quantize, lossless \\
2 & 2629 & 0.65 & {\footnotesize windows.x (63\%)} & cursor, icon, event, ctrl, expose \\
3 & 862 & 0.79 & {\footnotesize os.ms-windows.misc (45\%)} & irq, slip, xxxx, modem, dtr \\
4 & 1462 & 0.48 & {\footnotesize graphics (78\%)} & phigs, analysis, plot, conference, dimensional \\
5 & 874 & 0.55 & {\footnotesize graphics (74\%)} & format, hsi, jfif, pict, ppm \\
6 & 650 & 0.83 & {\footnotesize os.ms-windows.misc (43\%)} & donation, vb, shortly, moron, adpro \\
7 & 563 & 0.94 & {\footnotesize os.ms-windows.misc (29\%)} & lipman, maryland, penev, operational, bethesda \\
8 & 2685 & 0.94 & {\footnotesize graphics (34\%)} & hacker, mahan, tgv, confusion, decision \\
9 & 1914 & 0.92 & {\footnotesize sys.ibm.pc.hardware (38\%)} & dx, mhz, vesa, card, ati \\
10 & 1344 & 0.44 & {\footnotesize windows.x (76\%)} & gopher, contrib, sun, osf, motif \\
11 & 2834 & 0.79 & {\footnotesize sys.ibm.pc.hardware (40\%)} & repair, reinstall, fan, heat, boot \\
12 & 1162 & 0.99 & {\footnotesize sys.ibm.pc.hardware (23\%)} & appreciate, thank, help, greatly, advance \\
13 & 516 & 0.21 & {\footnotesize windows.x (93\%)} & usr, lib, rx, bin, openwinhome \\
14 & 1568 & 0.71 & {\footnotesize sys.mac.hardware (62\%)} & lc, fps, pb, performa, stereo \\
15 & 819 & 0.61 & {\footnotesize graphics (69\%)} & tt, atm, gifconverter, polygon, sphere \\
16 & 908 & 0.38 & {\footnotesize sys.ibm.pc.hardware (85\%)} & controller, scsi, aspi, slave, drive \\
17 & 276 & 0.69 & {\footnotesize windows.x (52\%)} & printf, vpic, scr, keycode, dpy \\
18 & 898 & 0.52 & {\footnotesize graphics (78\%)} & mirror, simtel, zip, pub, msdos \\
19 & 315 & 0.85 & {\footnotesize graphics (51\%)} & manchester, lilley, rt, hartmann, lance \\
\bottomrule
\end{tabularx}
\end{table}

\begin{table}[!t]
\centering
\small
\setlength{\tabcolsep}{3pt}
\caption{Every topic of \textsc{vSLDA} in the NYT/\textit{arts} run ($K=20$, MiniLM, run 0), with the fitted concentration parameter $\kappa_k$, its posterior sentence mass $n_k$, its normalized fine-category entropy $H$, its dominant fine category with the share and its top 5 representative words; the columns are defined in the text of this appendix.}
\label{tab:topics_nyt_arts_vslda}
\begin{tabularx}{0.9\textwidth}{@{}rrrrl>{\raggedright\arraybackslash\fontsize{9.5}{11.5}\selectfont}X@{}}
\toprule
\footnotesize Topic & \footnotesize $\kappa_k$ & \footnotesize $n_k$ & \footnotesize $H$ & \footnotesize Fine category & \footnotesize Representative words \\
\midrule
0 & 258.7 & 365 & 0.53 & {\footnotesize movies (77\%)} & oscar, award, nominee, nomination, academy \\
1 & 164.6 & 802 & 0.90 & {\footnotesize music (39\%)} & cambridge, honoree, mandela, golf, roosevelt \\
2 & 262.1 & 949 & 0.31 & {\footnotesize movies (90\%)} & film, movie, cinema, cinemascope, ridley \\
3 & 239.1 & 851 & 0.40 & {\footnotesize television (86\%)} & episode, netflix, show, abc, network \\
4 & 266.8 & 550 & 0.35 & {\footnotesize music (88\%)} & clarinet, instrument, accordion, piano, trumpet \\
5 & 232.8 & 567 & 0.58 & {\footnotesize movies (72\%)} & killing, indonesia, trayvon, jfk, aftermath \\
6 & 225.6 & 386 & 0.56 & {\footnotesize music (75\%)} & copyright, vinyl, download, database, rhapsody \\
7 & 239.7 & 910 & 0.66 & {\footnotesize movies (62\%)} & actor, mcconaughey, somali, aykroyd, reef \\
8 & 225.9 & 686 & 0.91 & {\footnotesize movies (45\%)} & steinbeck, andersen, architecture, painting, visualize \\
9 & 232.0 & 1172 & 0.94 & {\footnotesize movies (38\%)} & her, she, actress, sullen, ing \\
10 & 205.7 & 954 & 0.90 & {\footnotesize movies (34\%)} & summon, volatile, empathy, weakness, gov \\
11 & 289.7 & 586 & 0.13 & {\footnotesize music (96\%)} & opera, orchestra, verdi, philharmonic, britten \\
12 & 175.6 & 602 & 0.93 & {\footnotesize music (40\%)} & spider, fracture, leg, thigh, creature \\
13 & 134.0 & 1187 & 0.92 & {\footnotesize music (38\%)} & accomplishment, substantially, excerpt, salary, brink \\
14 & 330.1 & 668 & 0.28 & {\footnotesize dance (89\%)} & ballet, dance, dancer, choreographer, ballerina \\
15 & 243.9 & 606 & 0.68 & {\footnotesize music (69\%)} & festival, coachella, northside, bonnaroo, randalls \\
16 & 251.5 & 507 & 0.51 & {\footnotesize music (79\%)} & libretto, aria, penseroso, coloratura, freshness \\
17 & 261.7 & 627 & 0.45 & {\footnotesize music (83\%)} & rapper, rap, drake, sweatshirt, kanye \\
18 & 181.3 & 808 & 0.96 & {\footnotesize music (32\%)} & you, yourself, grandma, my, lick \\
19 & 255.9 & 891 & 0.29 & {\footnotesize music (91\%)} & song, album, band, medley, rock \\
\bottomrule
\end{tabularx}
\end{table}

\begin{table}[!t]
\centering
\small
\setlength{\tabcolsep}{3pt}
\caption{Every topic of SentLDA in the NYT/\textit{arts} run ($K=20$, MiniLM, run 0), with its posterior sentence mass $n_k$, its normalized fine-category entropy $H$, its dominant fine category with the share and its top 5 representative words; the columns are defined in the text of this appendix.}
\label{tab:topics_nyt_arts_sentlda}
\begin{tabularx}{0.9\textwidth}{@{}rrrl>{\raggedright\arraybackslash\fontsize{9.5}{11.5}\selectfont}X@{}}
\toprule
\footnotesize Topic & \footnotesize $n_k$ & \footnotesize $H$ & \footnotesize Fine category & \footnotesize Representative words \\
\midrule
0 & 1175 & 0.57 & {\footnotesize television (67\%)} & malick, episode, fx, clyde, pope \\
1 & 971 & 0.83 & {\footnotesize music (45\%)} & copyright, stream, netflix, beatles, bootleg \\
2 & 143 & 0.51 & {\footnotesize music (79\%)} & ferrando, mezzo, lukas, falstaff, soprano \\
3 & 24 & 0.55 & {\footnotesize movies (67\%)} & philippe, banter, exploitation, puckish, blanchett \\
4 & 577 & 0.54 & {\footnotesize music (77\%)} & euro, subsidy, gelb, bankruptcy, endowment \\
5 & 3372 & 0.93 & {\footnotesize music (32\%)} & you, lorne, me, didn, my \\
6 & 1214 & 0.24 & {\footnotesize music (93\%)} & hip, hop, album, band, rocky \\
7 & 146 & 0.92 & {\footnotesize movies (40\%)} & tuxedo, mischievous, fireplace, demonic, unpaid \\
8 & 511 & 0.24 & {\footnotesize music (93\%)} & jazz, saxophonist, armstrong, drum, synthesizer \\
9 & 225 & 0.36 & {\footnotesize music (86\%)} & nico, hindemith, sonata, weber, musicologist \\
10 & 1156 & 0.43 & {\footnotesize dance (79\%)} & ballet, dance, dancer, balanchine, choreographer \\
11 & 171 & 0.55 & {\footnotesize movies (76\%)} & commando, fantastical, gatsby, levi, bootlegger \\
12 & 1152 & 0.14 & {\footnotesize music (96\%)} & orchestra, philharmonic, carnegie, symphony, gilbert \\
13 & 921 & 0.48 & {\footnotesize movies (78\%)} & oscar, affleck, nomination, spielberg, django \\
14 & 157 & 0.74 & {\footnotesize movies (60\%)} & palma, mantua, addiction, strangle, assange \\
15 & 700 & 0.70 & {\footnotesize movies (67\%)} & rib, immigration, iranian, kiarostami, literature \\
16 & 451 & 0.24 & {\footnotesize movies (92\%)} & blu, ray, combo, pg, hangover \\
17 & 656 & 0.82 & {\footnotesize music (42\%)} & sweatshirt, thicke, earl, slender, scowl \\
18 & 866 & 0.54 & {\footnotesize movies (72\%)} & torture, osama, kong, cohn, congo \\
19 & 84 & 0.79 & {\footnotesize music (59\%)} & org, avenue, com, ifccenter, boerum \\
\bottomrule
\end{tabularx}
\end{table}

\FloatBarrier

\end{document}